\documentclass{article}

\PassOptionsToPackage{numbers, compress}{natbib}

\usepackage[final]{neurips_data_2021}
\usepackage{diagbox}
\usepackage{bm}

\usepackage[utf8]{inputenc} 
\usepackage[T1]{fontenc}    
\usepackage{hyperref}       
\usepackage{url}            
\usepackage{booktabs}       
\usepackage{amsfonts}       
\usepackage{nicefrac}       
\usepackage{microtype}      
\usepackage{epsfig}
\usepackage{float}
\usepackage{multicol}
\usepackage{multirow}
\usepackage{amssymb}
\usepackage{wrapfig}
\usepackage{graphicx}

\usepackage{color}
\definecolor{Tablecolor}{RGB}{255,163,250}
\definecolor{Suppcolor}{RGB}{247,7,154}
\usepackage[table,xcdraw,dvipsnames]{xcolor}
\usepackage{hyperref}

\usepackage{amsmath}
\usepackage{xspace}
\def\onedot{.\xspace}
\def\eg{\emph{e.g}\onedot} 

\def\ie{\emph{i.e}\onedot}

\def\supp{{\color{Suppcolor}supplementary}}

\makeatletter 
  \newcommand\figcaption{\def\@captype{figure}\caption} 
  \newcommand\tabcaption{\def\@captype{table}\caption} 
\makeatother

\title{AP-10K: A Benchmark for Animal Pose Estimation in the Wild}

\author{%
  Hang Yu$^{1*}$,
  Yufei Xu$^2$\thanks{Equal contribution. The work was done during the first authors' internship at JD Explore Academy.}~~, 
  Jing Zhang$^2$\thanks{Corresponding author}~~,
  Wei Zhao$^{1\dag}$,
  Ziyu Guan$^1$,
  Dacheng Tao$^{{3,2}}$
  \AND
  \vspace{-5 mm}
  \\
  \textsuperscript{1}Xidian University, China,
  \textsuperscript{2}The University of Sydney, Australia,
  \textsuperscript{3}JD Explore Academy, China \\
  \vspace{-25 mm}
}

\begin{document}

\maketitle

\begin{abstract}

Accurate animal pose estimation is an essential step towards understanding animal behavior, and can potentially benefit many downstream applications, such as wildlife conservation. Previous works of animal pose estimation only focus on specific animals while ignoring the diversity of animal species, limiting their generalization ability. In this paper, we propose AP-10K, the first large-scale benchmark for mammal animal pose estimation, to facilitate research in animal pose estimation. AP-10K consists of 10,015 images collected and filtered from 23 animal families and 54 species following the taxonomic rank and high-quality keypoint annotations labeled and checked manually. Based on AP-10K, we benchmark representative pose estimation models on the following three tracks: (1) supervised learning for animal pose estimation, (2) cross-domain transfer learning from human pose estimation to animal pose estimation, and (3) intra- and inter-family domain generalization for unseen animals. The experimental results provide sound empirical evidence on the superiority of learning from diverse animals species in terms of both accuracy and generalization ability. It opens new directions for facilitating future research in animal pose estimation. AP-10k is publicly available at  https://github.com/AlexTheBad/AP10K\footnote{The code will also be integrated into mmpose.}.
\end{abstract}

\section{Introduction}
\label{sec:intro}

Pose estimation, referring to identifying the category and location of a series of body keypoints, is a fundamental computer vision task, which is also useful in many practical applications, such as activity recognition\cite{ni2017learning,baradel2018glimpse}, behavior understanding~\cite{arac2019deepbehavior}, human-object interaction\cite{zhang2020empowering,mazhar2018towards}. While many research efforts have been made for human pose estimation~\cite{zhang2021towards,WangSCJDZLMTWLX19} in both large-scale benchmarks as well as advanced algorithms, fewer research works are focusing on animal pose estimation. One of the main reasons is that it is very challenging to collect and annotate a large-scale benchmark dataset for animal pose estimation, like MPII dataset \cite{andriluka14cvpr} and COCO dataset \cite{lin2014microsoft} for human pose estimation, especially considering the fact that there are many species of animals and it requires some domain knowledge to perform annotation. Nevertheless, animal pose estimation is of great significance in both research purposes and practical applications such as zoology and wildlife conservation~\cite{harding2004animal,davies2015keep}. Therefore, it is necessary to pave the way for research in this field by establishing a large-scale benchmark covering many different animal species.

Previous works for animal pose estimation focus on specific animal species, \eg, horse~\cite{mathis2021pretraining}, zebra~\cite{graving2019deepposekit}, macaque~\cite{labuguen2020macaquepose}, fly~\cite{pereira2019fast}, and tiger~\cite{li2020atrw}. While they make contributions in advancing the research for animal pose estimation regarding both datasets and algorithms, they suffer from the poor generalization ability to unseen animal species, limiting their practical significance. Some other works~\cite{Cao_2019_ICCV,graving2019deepposekit} try to mitigate this issue by taking several animal species into consideration and construct new datasets covering multiple animal species. However, the number of animal species is still very limited, \eg, only five species in~\cite{Cao_2019_ICCV}. In addition, these datasets are not organized following the taxonomic rank, and there is no animal family and species structure. Consequently, the following important research questions remain unclear, \ie, 1) how about the performance of different representative human pose models on the animal pose estimation task? 2) will the representation ability of a deep model benefit from training on a large-scale dataset with diverse species? 3) how about the impact of pretraining, \eg, on the ImageNet dataset \cite{deng2009imagenet} or human pose estimation dataset \cite{lin2014microsoft}, in the context of the large-scale of dataset with diverse species? 4) how about the intra- and inter-family generalization ability of a model trained using data from specific species or family?

In this paper, we make an attempt to answer these questions by collecting the first large-scale benchmark AP-10K for general mammal pose estimation. It consists of 10,015 images collected and filtered from 23 animal families and 54 species following the taxonomic rank, where the keypoints of all animal instances on each image are manually labeled and carefully double-checked. Specifically, various animal images are collected from existing publicly available datasets, and a cleaning process is carried out to remove the replicated images. Then, they are organized following the taxonomic rank. Finally, thirteen annotators are recruited to carefully annotate the bounding boxes for all animal instances in each image and their body keypoints, as well as the background category of each image, following the COCO~\cite{lin2014microsoft} annotation style. In addition to the labeled 10,015 images, AP-10K also contains about 50k animal images organized following the taxonomic rank but without keypoint annotations, which can be used for animal pose estimation at the settings of semi-supervised learning~\cite{dai2015semi} and self-supervised learning~\cite{he2020momentum,chen2020simple,chen2020improved}.

\begin{figure}
    \centering
    \includegraphics[width=0.85\linewidth]{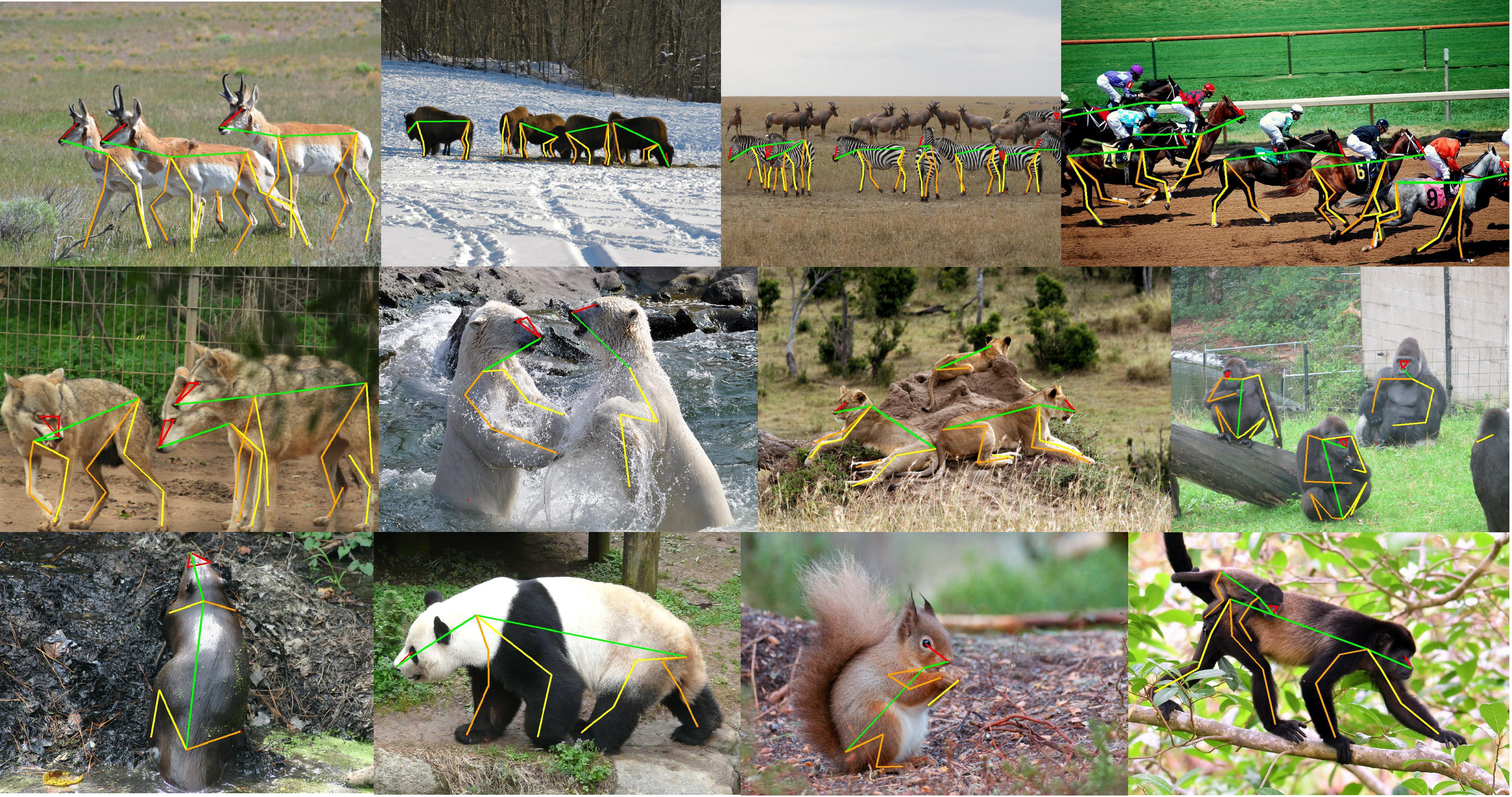}
    \vspace{-3 mm}
    \caption{A glance at diverse animal species in AP-10K. Figures are best viewed in color.
    }
    \vspace{-8 mm}
    \label{fig:Overview}
\end{figure}

The diversity in the animal species and the family-species organization of AP-10K provides a possibility to study the aforementioned research questions. To this end, we benchmark representative pose estimation models~\cite{cheng2020higherhrnet,he2016deep,xiao2018simple} on the following three tracks, \ie, (1) supervised learning for animal pose estimation (SL track), (2) cross-domain transfer learning between human pose estimation and animal pose estimation (CD-TL track), and (3) intra- and inter-family domain generalization for unseen animals (DG track). In the SL track, we study the representation ability of different models at the settings of random weight initialization or initialization with ImageNet pretrained weights. In the CD-TL track, we investigate whether pretraining on a human pose estimation dataset like COCO benefits the animal pose estimation on AP-10K. In the DG track, we first study the intra-family domain generalization, where the model is trained using some species belonging to a specific family and tested on other species from the same family. Then, we also study the inter-family domain generalization, where the model is trained using all species from a certain family and tested on species from a different family. The detailed experiment settings and results are presented in Section~\ref{sec:experiment}. In summary, the experimental results provide sound empirical evidence on the superiority of learning from diverse animals species in terms of both accuracy and generalization ability. 

The main contribution of the paper is twofold. (1) We establish the first large-scale benchmark AP-10K for mammal animal pose estimation, which contains 10,015 images from 23 families and 54 species following the taxonomic rank, along with high-quality keypoint annotations. (2) Based on AP-10K, we study several challenging and open research questions in this area by benchmarking representative pose estimation models on the SL track, the CD-TL track, and the DG track. The results show sound empirical evidence on the superiority of learning from diverse animals species. 
\section{Related work}
\label{sec:related}

\subsection{Human pose estimation}

Human pose estimation is an active research area in computer vision. In general, human pose estimation methods can be categorized as bottom-up methods~\cite{newell2017associative} and top-down ones~\cite{xiao2018simple,wang2020deep,liu2020improving}. While the former ones are usually fast, the latter ones always achieve the top entries in representative benchmarks due to their high accuracy. Well-known large-scale human pose estimation benchmarks are COCO~\cite{lin2014microsoft} and MPII~\cite{andriluka14cvpr}, while some new datasets have been established recently, regarding either crowd scenes like CrowdPose~\cite{li2018crowdpose} or occlusion scenes like OCHuman~\cite{zhang2019pose2seg}. These datasets have significantly advanced the research in this area. Models~\cite{xiao2018simple,zhang2021towards,WangSCJDZLMTWLX19} trained using these datasets have been proven effective for human pose estimation, owing to the rich diversity of posture, illumination, scale, occlusion, and the number of person instances per image. 

\subsection{Animal pose estimation}
Animal pose estimation has attracted increasing attention in the research community. Typical human pose estimation models can be applied to animals since there is no difference in modeling, \ie, except for the number of defined keypoints. It is the labeled dataset that matters in the era of deep learning. In the past few years, some datasets have been established to facilitate research in animal pose estimation~\cite{mathis2021pretraining,graving2019deepposekit,li2020atrw}. However, most of them focus on specific animal species, \eg, horse~\cite{mathis2021pretraining}, zebra~\cite{graving2019deepposekit}, macaque~\cite{labuguen2020macaquepose}, fly~\cite{pereira2019fast}, and tiger~\cite{li2020atrw}, and are limited in diversity of postures, textures, and habitats. Recently, the Animal Pose Dataset is introduced in \citep{Cao_2019_ICCV}, which has 5 animal species. In addition to its limited number of species, it is not organized following the taxonomic rank, \ie, there is no structure of animal families and species, making it impossible to study the intra- and inter-family generalization problem and other ones mentioned above. In contrast to human pose datasets, which have only one species, the very familiar human, collecting and annotating animal keypoints is more challenging since there are many different species of animals belonging to different families, and specific biological knowledge is required to distinguish the keypoints of different animals. While challenging and costly, it is very necessary to establish a large-scale benchmark to facilitate research in this field.

\subsection{Transfer learning}

Transfer learning refers to transferring the learned knowledge from a source task to a target task, which is related to source one. Usually, the source domain has a large scale of labeled data while there is only a limited number of labeled data in the target domain. The prevalent transfer learning methods follow the ``pretraining and finetuning'' route. It has been proven effective in many computer vision tasks, including image classification~\cite{kolesnikov2020big}, object detection~\cite{he2017mask}, semantic segmentation~\cite{chen2017rethinking}, as well as human pose estimation~\cite{wang2020deep}. While the ImageNet dataset is widely used as the source domain for transfer learning in different computer vision tasks, including animal pose estimation, some works also investigate the effectiveness of transferring knowledge from the human pose estimation task to the animal pose estimation task~\cite{Cao_2019_ICCV}. However, it is unexplored and unclear whether transferring from ImageNet or existing human pose datasets is still effective or not, when a large-scale animal pose dataset is available. A related research is \cite{He_2019_ICCV}, which shows that a longer training schedule on a large-scale dataset for a downstream task (\ie, object detection) can catch up the performance gain from pretraining on ImageNet. In this paper, we study this problem based on our proposed AP-10K dataset and show that pretraining on ImageNet is not necessary for animal pose estimation when a longer training schedule is used.

\section{Dataset}
\label{sec:dataset}

\subsection{Data collection and organization} 

\noindent \textbf{Data collection} 

As we have discussed, the limited number of animal species in existing animal pose datasets makes it difficult to comprehensively evaluate the performance of animal pose estimation models, considering the diversity in real-world animal species. To facilitate research in this area, there is a need to collect a large-scale animal pose estimation dataset with many different animal species. To this end, we propose the AP-10K dataset. First, we resort to existing publicly available datasets focusing on animals~\cite{xian2018zero,africanWildLife,wcats,animal5,animalDCP,animals10,IUCN,endangeredanimals}. Although these datasets are designed for the animal classification task or animal detection task, they provide abundant animal images from different fine-grained species. Then, we collect them as a whole and remove the repeated images or mislabeled images, including a coarse stage by aHash~\cite{ahash_algorithm} and a refinement stage by manual double-check. In this way, we obtain all the candidate animal images with high-quality species category labels for our AP-10K dataset, \ie, 59,658 images in total.

\noindent \textbf{Data organization} We re-organize and re-label the images following the taxonomic rank, \ie, family and species. Instead of using the exact biology definition of family-genus-species, we do not distinguish genus and species for simplicity. Such a classification takes advantage of a biological prior about evolution, where animals belonging to the same family/species have a similar texture, appearance, and pose distributions. Besides, when evaluating the model generalization to unseen animals, ignoring biological relationships among seen and unseen animals will lead to highly biased results. The hierarchical categorization in our AP-10K can provide a possibility to evaluate intra and inter-family model generalization in a fair setting.

\begin{figure}[htbp] 

  \begin{minipage}[b]{0.5\linewidth} 
    \centering 
    \scriptsize
    \begin{tabular}{c|l|c|c}
    \hline
    Keypoint & \multicolumn{1}{c|}{Definition} & Keypoint & Definition \\
    \hline
    1     & Left Eye & 10    & \multicolumn{1}{l}{Right Elbow} \\
    2     & Right Eye & 11    & \multicolumn{1}{l}{Right Front Paw} \\
    3     & Nose  & 12    & \multicolumn{1}{l}{Left Hip} \\
    4     & Neck  & 13    & \multicolumn{1}{l}{Left Knee} \\
    5     & Root of Tail & 14    & \multicolumn{1}{l}{Left Back Paw} \\
    6     & Left Shoulder & 15    & \multicolumn{1}{l}{Right Hip} \\
    7     & Left Elbow & 16    & \multicolumn{1}{l}{Right Knee} \\
    8     & Left Front Paw & 17    & \multicolumn{1}{l}{Right Back Paw} \\
    9     & Right Shoulder &       &  \\
    \hline
    \end{tabular}%
    \tabcaption{Definition of animal keypoints.} 
    \label{tab:KeypointDefinition} 
  \end{minipage}%
  \begin{minipage}[b]{0.5\linewidth} 
    \centering
    \includegraphics[width=0.5\linewidth]{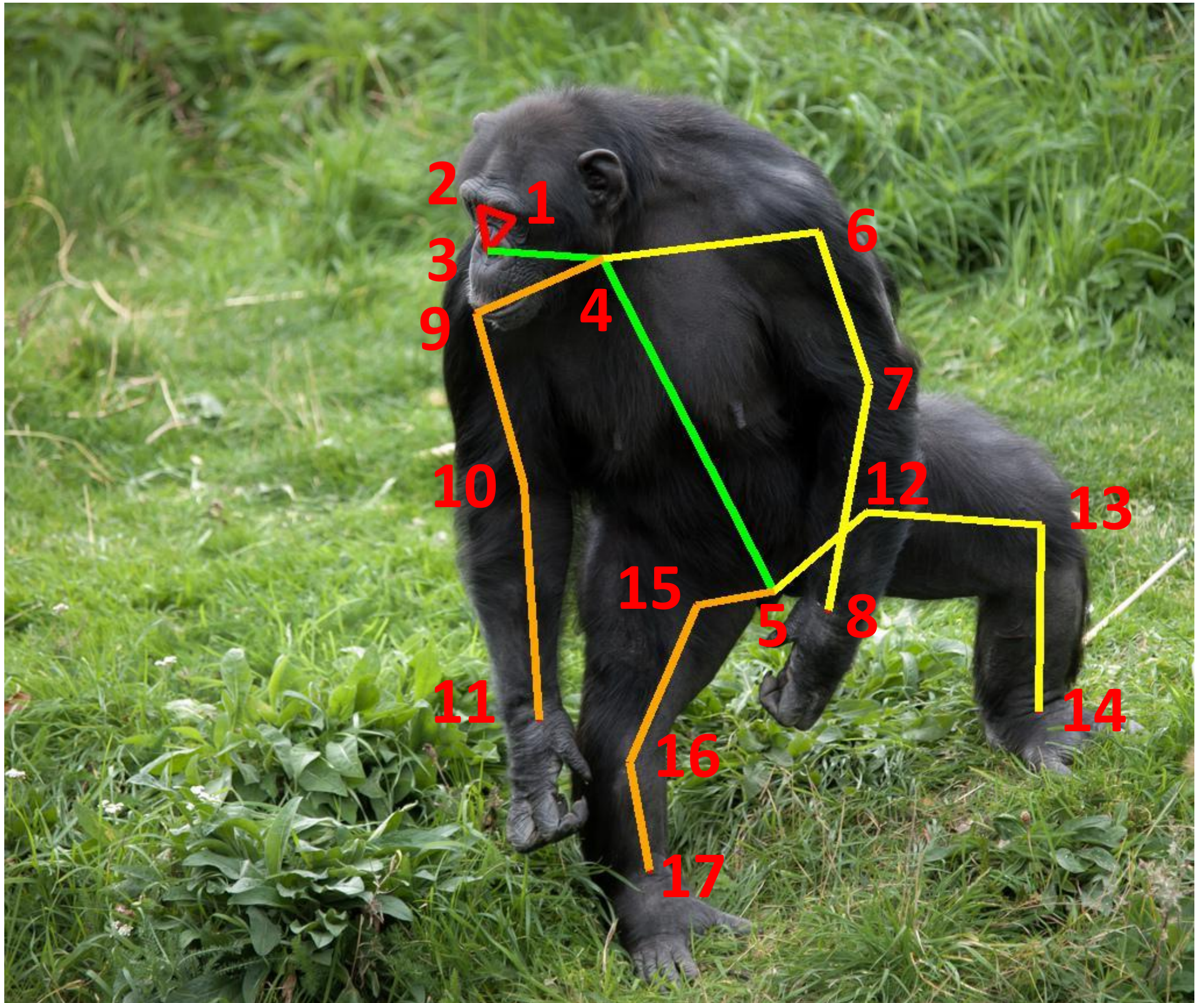}
    \caption{The keypoints of a Chimpanzee.}
    \label{fig:KeypointDefinition} 
  \end{minipage} 
  \vspace{-10 mm}
\end{figure}

\subsection{Data annotation}

To obtain high-quality animal pose annotations for each image, we recruited 13 well-trained annotators and asked them to annotate all the keypoints of each animal they can distinguished. Three rounds of cross-checking and correction are then carried out to improve the annotation quality. Finally, it took about three months to complete the whole annotation process, where 10,015 images were labeled. Similar to the keypoints defined for representing human pose, 17 keypoints are defined to represent animal pose, including two eyes, one nose, one neck, one tail, two shoulders, two elbows, two knees, two hips, and four paws, as listed in Table~\ref{tab:KeypointDefinition}. A visual example of the keypoint annotations of a Chimpanzee is shown in Figure~\ref{fig:KeypointDefinition}. Besides, We also labeled 8 background type for all posed images, \ie, grass or savanna, forest or shrub, mud or rock, snowfield, zoo or human habitation, swamp or riverside, desert or gobi and mugshot. The annotations are saved in line with the COCO format to facilitate further research in animal pose estimation by reusing common training and evaluation tools developed in the human pose estimation community. The AP-10K is split into three disjoint subsets, \ie, train, validation, and test sets, at the ratio of 7:1:2 per animal species. It is also noteworthy that in addition to the 10,015 images with keypoint annotations, we also include the remaining 49,643 images with their family and species labels in our AP-10K to facilitate future research, \eg, semi-supervised learning and self-supervised learning for animal pose estimation.

\subsection{Statistics of the AP-10K dataset}

\begin{table}[htbp]
  \centering
  \footnotesize
  \caption{Comparison of different animal pose datasets.}
   \setlength{\tabcolsep}{0.008\linewidth}{\begin{tabular}{c|ccccccc}
    \hline
    dataset & species & family & labeled image & unlabeled image & keypoint & instance  \\
    \hline
    Animal-Pose Dataset~\cite{Cao_2019_ICCV} & 5 & N/A & 4,666 & 0   & 20    & 6,117  \\
    Horses-10~\cite{mathis2021pretraining} & 1 & N/A & 8,110  & 0  & 22    & 8,110 \\
    ATRW~\cite{li2020atrw}  & 1 &  N/A & 8,076 & 0 & 15    & 9,496 \\
    \textbf{AP-10K} & \textbf{54} & \textbf{23} & \textbf{10,015}  & \textbf{50k}  &  \textbf{17} & \textbf{13,028} \\
    \hline
    \end{tabular}}%
  \label{tab:comparation of datasets}%
\end{table}%

\begin{figure}[htbp] 
  \begin{minipage}[htbp]{0.5\linewidth} 
    \centering 
    \includegraphics[width=0.72\linewidth]{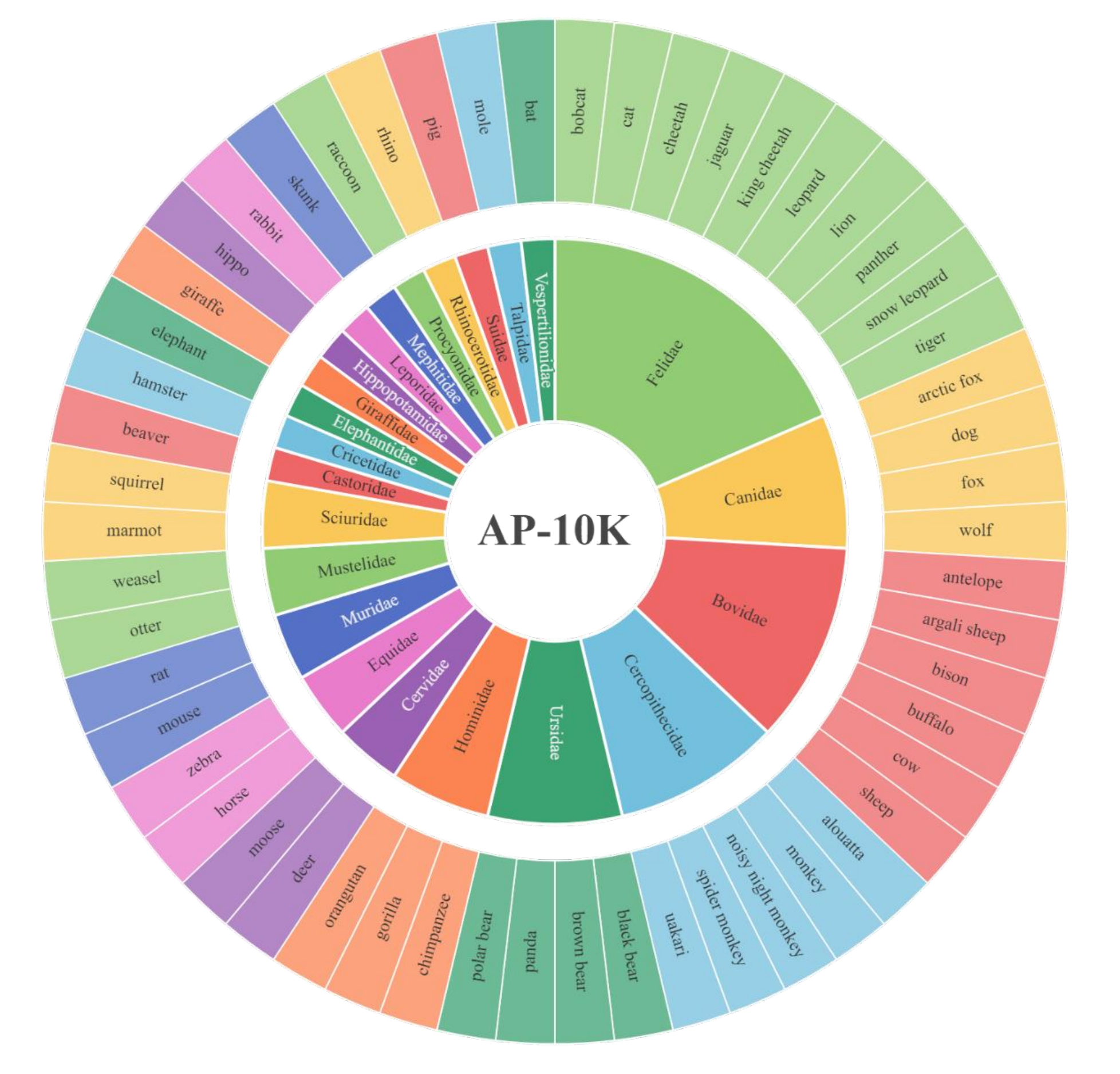}
    \end{minipage}%
  \begin{minipage}[htbp]{0.5\linewidth} 
    \centering
    \includegraphics[width=0.9\linewidth]{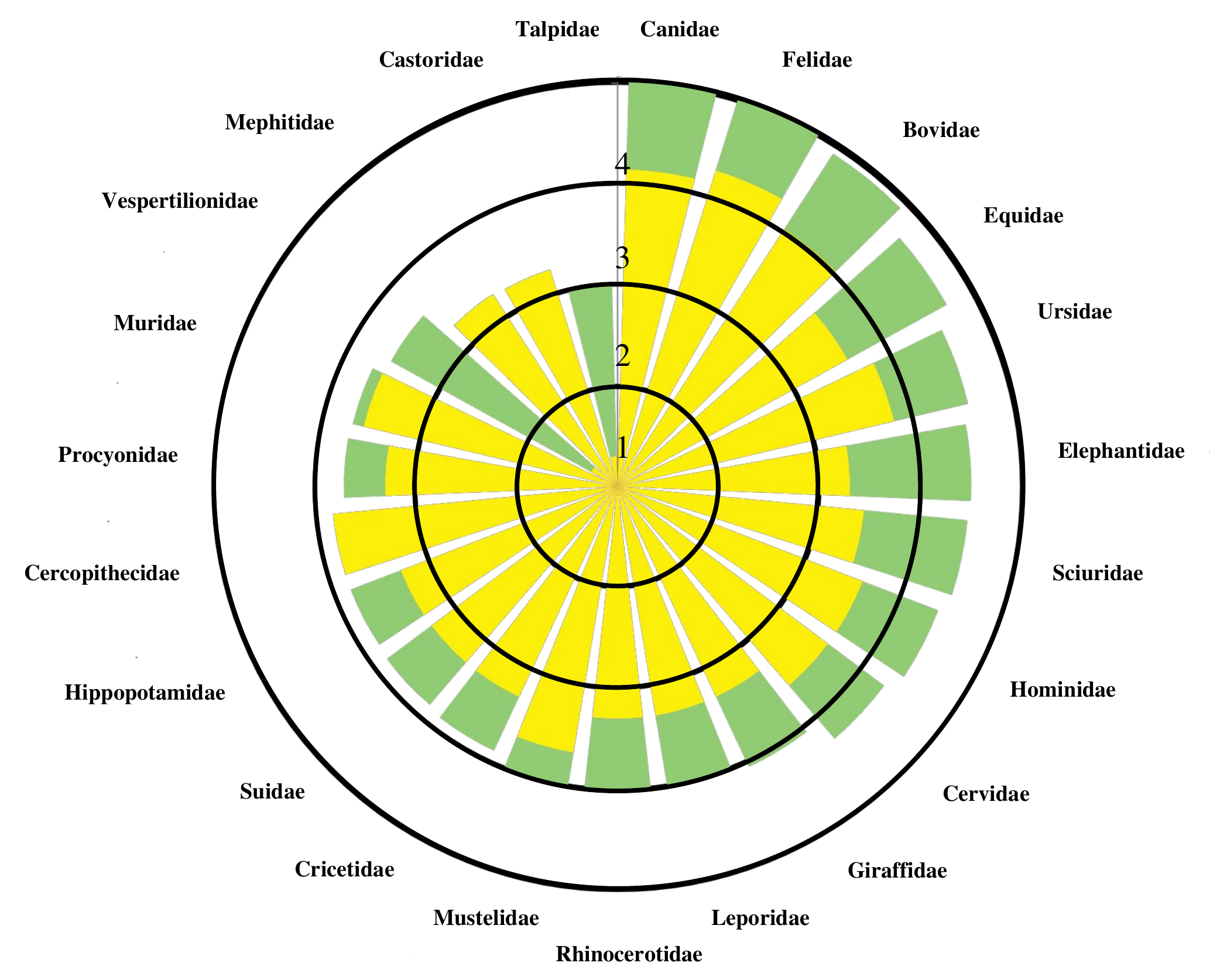}
  \end{minipage}
  \vspace{-3 mm}
  \caption{(a) The organization of family and species in AP-10K. (b) The distributions (in the logarithmic scale) of the number of labeled (yellow) and whole (green) images in each family.}
  \vspace{-5 mm}
  \label{fig:posed_images_scale_per_family_20}
\end{figure}

\noindent \textbf{Overview of AP-10K} As shown in Table~\ref{tab:comparation of datasets}, AP-10K dataset covers 23 families and 54 species of animals, which is much richer in the diversity of animal species than previous datasets. Besides, AP-10K contains much more labeled images and instances than the other datasets, \ie, 10,015 images and 13,028 instances in total, resulting in a more complex animal pose distribution. Besides, we also provide extra 50k images without keypoint annotations but family and species labels in our AP-10K. Thanks to the taxonomic rank-based organization of AP-10K, these unlabeled data can be exploited for further research, \eg, semi-supervised learning and self-supervised learning for animal pose estimation, which is not applicable in other datasets.

\noindent \textbf{Long-tail property} As shown in Figure~\ref{fig:posed_images_scale_per_family_20}, the number of images in each family of AP-10K has a long-tail distribution, which reflects the true distribution of animals in the wild due to the commonness or rarity of the animals in some extent. For example, there are 1,913 labeled images (9,277 unlabeled images) for common families like Felidae and only 200 labeled images (312 unlabeled images) for rare families like Procyonidae. Similar properties can be found in CityScape~\cite{cordts2016cityscapes} and LVIS~\cite{gupta2019lvis} datasets. On the other hand, the number of species per family varies considerably. For example, the most common family in AP-10K, \ie, Felidae, contains 10 more species than the rarest family, \ie, Castoridae, which contains only 1 species. The property of long-tail distribution makes AP-10K be a challenging benchmark for mammal animal pose estimation. Specifically, it can also be used to study few-shot learning for animal pose estimation.

\section{Experiment}
\label{sec:experiment}

\subsection{Implementation details}
We benchmark several representative pose estimation frameworks~\cite{newell2016stacked,xiao2018simple,wang2020deep} with different backbone networks~\cite{He_2019_ICCV,wang2020deep} on the proposed AP-10K dataset based on the MMPose~\cite{mmpose2020} codebase. A single NVIDIA Tesla V100 GPU with 16GB memory is used during both the training and testing for all tracks, \ie, the SL Track, CD-TL Track, and DG Track. The detailed settings for each track are presented in the following part. We adopt the mean average precision (mAP) as the primary evaluation metric in all tracks, following the human and animal pose estimation literature~\cite{wang2020deep,WangSCJDZLMTWLX19,xiao2018simple,Cao_2019_ICCV}.

\subsection{Supervised learning track}
\label{subsec:SL_track}

\begin{table}[htbp]
  \centering
  \small
  \caption{The evaluation results (mAP) of different models on the validation set of the SL Track.}
    \setlength{\tabcolsep}{0.008\linewidth}{\begin{tabular}{c|cccccc}
    \hline
     & HRNet-w32~\cite{wang2020deep} & HRNet-w48~\cite{wang2020deep} & ResNet50~\cite{he2016deep} & ResNet101~\cite{he2016deep} & Hourglass~\cite{newell2016stacked}\\
    \hline
    w/o pretraining & $0.703_{\pm 0.002}$ & $0.713_{\pm 0.002}$ & $0.646_{\pm 0.001}$ & $0.667_{\pm 0.002}$ & $0.686_{\pm 0.006}$\\
    w/ pretraining  &  $0.738_{\pm 0.006}$ & $0.744_{\pm 0.004}$ & $0.699_{\pm 0.004}$ & $0.698_{\pm 0.002}$ & $0.729_{\pm 0.001}$\\
    \hline
    \end{tabular}}%
  \label{tab:SLResults}%
\end{table}%

\begin{table}[htbp]
  \centering
  \caption{The evaluation results (mAP) of HRNet-w32 \cite{wang2020deep} on the validation set of the SL track.}
    \begin{tabular}{c|ccc}
    \hline
    Training epochs & 210 & 420 & 630 \\
    \hline
    Training from scratch & $0.703_{\pm 0.002}$ & $0.721_{\pm 0.005}$ & $0.725_{\pm 0.003}$ \\
    Pretraining on ImageNet & $0.738_{\pm 0.005}$ & $0.738_{\pm 0.005}$ & $0.735_{\pm 0.002}$ \\
    \hline
    \end{tabular}%
  \label{tab:HREpochs}%
\end{table}%

\noindent \textbf{Settings} The SL track aims to evaluate the representation ability of different human pose models, including Hourglass~\cite{newell2016stacked}, SimpleBaseline~\cite{xiao2018simple} with ResNet50~\cite{he2016deep} and ResNet101~\cite{he2016deep} as backbones, and HRNet \cite{wang2020deep} for the animal pose estimation task. Hourglass utilizes multi-stage structure with skip connections for human pose estimation. SimpleBaseline adopts an encoder-decoder structure to regress the keypoints while HRNet keeps both high- and low-level features for more accurate pose estimation. We use Adam~\cite{kingma2014adam} as the optimizer and train these representative models for 210 epochs with a batch size of 64. The initial learning rate is set to 5e-4, and we adopt the step-wise learning rate schedule which reduces the learning rate at the 170 and 200 epoch, respectively. The training image of each instance is cropped and resized to $\text{256} \times \text{256}$, with random rotation, flip, and scale jitter as data augmentation. The AP-10K dataset is randomly split into the disjoint train, validation, and test sets three times following the same ratio 7:1:2, where we perform the experiment three times accordingly. We consider two popular training settings, \ie, with and without ImageNet~\cite{deng2009imagenet} pretraining, respectively. \textbf{\textit{Pretraining on ImageNet}}: We follow the common practice in human pose estimation literature \cite{xiao2018simple,wang2020deep} by initializing the backbone network with the pretrained weights on ImageNet and then finetuning the whole network on the AP-10K training set. \textbf{\textit{Training from scratch on AP-10K}}: We randomly initialize the network weights and directly train them on the AP-10K training set. We also adopt longer training schedules, \ie, 2$\times$ for 420 epochs and 3$\times$ for 630 epochs at these two settings.
By comparing the performance gap between these two settings, we can investigate the impact of pretraining on ImageNet, especially when a longer training schedule is used for training from scratch.

\noindent \textbf{Results on the SL Track} The results are summarized in Table~\ref{tab:SLResults}. We have the following observations. Firstly, with the increase of network complexity, the performance of both SimpleBaseline \cite{xiao2018simple} and HRNet \cite{wang2020deep} methods is generally improved, especially at the setting of training from scratch. It is reasonable since the representation ability of each model mainly depends on the representation ability of the backbone network. Secondly, advanced network architecture like HRNet outperforms the encoder-decoder architecture in \cite{xiao2018simple} and hourglass architecture in \cite{newell2016stacked} due to its high-resolution feature representation ability. Thirdly, pretraining on the ImageNet leads to better performance for all the methods, which is attributed to the strong generalization ability of the pretrained network. By making a trade-off between model complexity and performance, we choose HRNet-w32 as the default model in the following experiments.

When adopting a longer training schedule, the performance gap between pretraining on ImageNet and training from scratch has been reduced as shown in Table~\ref{tab:HREpochs}, which is similar to the observation in \cite{He_2019_ICCV} for generic object detection. In other word, pretraining helps accelerate the convergence speed since it can provide a good starting point of the network weights in the parameter space. However, when a longer training schedule is used, the network can also learn a good feature representation from the abundant training data. It again confirms the value of the proposed AP-10K. 

\subsection{Cross-domain transfer learning track}

\begin{table}[htbp]
  \centering
  \caption{The evaluation results (mAP) of HRNet-w32 \cite{wang2020deep} on the validation set of the CD-TL track.}
    \begin{tabular}{c|ccccc}
    \hline
    epoch   & AP    & AP$_{.5}$  & AP$_{.75}$ & AP$_{\text{M}}$ & AP$_{\text{L}}$ \\
    \hline
    20 & $0.606_{\pm 0.004}$ & $0.906_{\pm 0.005}$ & $0.635_{\pm 0.006}$ & $0.501_{\pm 0.037}$ & $0.610_{\pm 0.003}$\\
    30 & $0.642_{\pm 0.002}$ & $0.921_{\pm 0.010}$ & $0.680_{\pm 0.002}$ & $0.521_{\pm 0.044}$ & $0.645_{\pm 0.002}$\\
    40 & $0.667_{\pm 0.003}$ & $0.934_{\pm 0.004}$ & $0.714_{\pm 0.007}$ & $0.547_{\pm 0.059}$ & $0.671_{\pm 0.003}$\\
    210 & $0.753_{\pm 0.005}$ & $0.962_{\pm 0.002}$ & $0.827_{\pm 0.003}$ & $0.616_{\pm 0.031}$ & $0.756_{\pm 0.004}$\\
    \hline
    \end{tabular}%
  \label{tab:ResultsCD}%
\end{table}%

\noindent \textbf{Settings} 
Since four-foot animals usually share some similar keypoints as humans, \eg, two eyes, one nose, and one neck, the feature representation learned from a human pose estimation task may also be beneficial for animal pose estimation, especially on a small-scale dataset as shown in \cite{Cao_2019_ICCV}. However, it is unclear the impact of such cross-domain transfer learning on a large-scale animal pose dataset with diverse species, \eg, our AP-10K. To answer this question, we use the pretrained HRNet-w32~\cite{wang2020deep} on the COCO human pose dataset~\cite{lin2014microsoft} to initialize the network weights and then finetune the model for different epochs, \ie, 20, 30, 40, and 210 epochs, respectively. We follow the same setting in the SL track except that the learning rate is fixed as 1e-4.

\noindent \textbf{Results on the CD-TL Track} The results are summarized in Table~\ref{tab:ResultsCD}, $AP_{.5/75}$ means using $0.5$ and $0.75$ as the threshold when computing the average precision and $AP_{M/L}$ represents the average precision on medium and large scale objects, respectively, as in COCO~\cite{lin2014microsoft}. It can be concluded that the models transferred from human pose estimation do not perform well on the animal pose estimation task when the finetuning schedule is short, \eg, 0.606 mAP (20 epochs), 0.642 mAP (30 epochs), and 0.667 mAP (40 epochs) compared with 0.738 mAP in Table~\ref{tab:SLResults}. We suspect the performance gap is attributed to the domain gap between these two domains, \eg, the instances in human pose estimation always wear clothes while animals are always covered by different types of fur. Nevertheless, training for a longer schedule (\eg, 210 epochs) can help to close the gap and even improve the performance further, compared with pretraining on ImageNet, \eg, 0.753 v.s. 0.738. implying that the domain gap between classification and animal pose estimation is larger than that between the two pose estimation tasks.

\subsection{Intra- and inter-family domain generalization track}

\subsubsection{Intra-family domain generalization}
\label{subsubsec:Intra-family}

\noindent \textbf{Settings} 
We choose the three largest families, \ie, Bovidae, Canidae, and Felidae, to conduct the Intra-family DG experiments. For each family, we randomly select one species as the test set and use all other species as the train set. We follow the same setting as the pretraining on ImageNet in the SL track except that we train the models for more epochs to ensure that the same amount of training images are seen as in the SL Track.

\noindent \textbf{Results on the Intra-family DG Track} The results are summarized in Table~\ref{tab:intraCow}, Table~\ref{tab:intraDog}, and Table~\ref{tab:intraCat}. The diagonal scores denote the results when testing on unseen species at different settings (as shown in the first column). Besides, we calculate the average evaluation result on each seen species at different settings and show them at the bottom of each column, respectively. We have the following observations. Firstly, when testing on unseen species, the model can still obtain a good performance although it is inferior to that when testing on seen species. It is reasonable since the species within the same family share some common features while having unique characteristics too. It is noteworthy that the evaluation result on Dog is worse than others since there are more images of dogs than fox and wolf. Besides, dogs varies widely in appearance owning to human's cultivation for family pets. Similar phenomena can also be observed in Table~\ref{tab:intraCat}, \eg, Cat. Secondly, the performance on each seen species is not as good as the corresponding performance when using all the species in AP-10K for training, \eg, 0.663 mAP v.s. 0.761 mAP (as shown in Table~\ref{tab:per_species_test_pretrained_54} in the \supp.) for Sheep. It implies that using the amount of training data but from more diverse species helps the model learning better feature representation, \ie, confirming the value of our AP-10K again.

\begin{table}[htbp]
  \centering
  \small
  \caption{Intra-family DG results (mAP) of HRNet-w32~\cite{wang2020deep} on the test set of the Bovidae family. Bov.=Bovidae, Ant.=Antelope, A.S.=Argali Sheep, Bis.=Bison, Buf.=Buffalo, She.=Sheep.}
    \setlength{\tabcolsep}{0.001\linewidth}{
    \begin{tabular}{c|lllllll}
    \hline
	    \diagbox{Test}{Train}  & \multicolumn{1}{c}{{\textbf{\textit{Bov.}}/Ant.}}  & \multicolumn{1}{c}{\textbf{\textit{Bov.}}/A.S.}  & \multicolumn{1}{c}{\textbf{\textit{Bov.}}/Bis.}  & \multicolumn{1}{c}{\textbf{\textit{Bov.}}/Buf.}  & \multicolumn{1}{c}{\textbf{\textit{Bov.}}/Cow}  & \multicolumn{1}{c}{\textbf{\textit{Bov.}}/She.}  & \multicolumn{1}{c}{\textbf{Average}} \\
    \hline
    Antelope & \bm{$0.607_{\pm ±0.010}$}	&	$0.742_{\pm 0.013}$	&	$0.775_{\pm 0.004}$	&	$0.842_{\pm 0.005}$	&	$0.729_{\pm 0.002}$	&	$0.853_{\pm 0.002}$ &  $0.788_{\pm 0.051}$ \\
	A.S. & $0.836_{\pm 0.016}$	&	\bm{$0.655_{\pm 0.015}$}	&	$0.805_{\pm 0.022}$	&	$0.840_{\pm 0.007}$	&	$0.725_{\pm 0.027}$	&	$0.697_{\pm 0.008}$ & $0.781_{\pm 0.059}$ \\
	Bison & $0.731_{\pm 0.017}$	&	$0.646_{\pm 0.009}$	&	\bm{$0.530_{\pm 0.006}$}	&	$0.605_{\pm 0.006}$	&	$0.616_{\pm 0.009}$	&	$0.693_{\pm 0.014}$ & $0.658_{\pm 0.047}$ \\
	Buffalo & $0.783_{\pm 0.010}$	&	$0.748_{\pm 0.031}$	&	$0.726_{\pm 0.017}$	&	\bm{$0.658_{\pm 0.004}$}	&	$0.794_{\pm 0.022}$	&	$0.750_{\pm 0.008}$ & $0.760_{\pm 0.025}$ \\
	Cow & $0.597_{\pm 0.011}$	&	$0.691_{\pm 0.004}$	&	$0.740_{\pm 0.007}$	&	$0.732_{\pm 0.009}$	&	\bm{$0.586_{\pm 0.006}$}	&	$0.683_{\pm 0.002}$ & $0.689_{\pm 0.051}$ \\
	Sheep & $0.707_{\pm 0.012}$	&	$0.607_{\pm 0.006}$	&	$0.681_{\pm 0.004}$	&	$0.676_{\pm 0.007}$	&	$0.645_{\pm 0.005}$	&	\bm{$0.520_{\pm 0.001}$} & $0.663_{\pm 0.034}$ \\ 
    \hline
    \end{tabular}}%
  \label{tab:intraCow}%
\end{table}%

\begin{table}[htbp]
  \centering
  \caption{Intra-family DG results (mAP) of HRNet-w32~\cite{wang2020deep} on the test set of the Canidae family.}
    \begin{tabular}{c|cccc}
    \hline
    \diagbox{Test}{Train} & \textbf{\textit{Can.}}/Dog   &  \textbf{\textit{Can.}}/Fox  & \textbf{\textit{Can.}}/Wolf & \textbf{Average} \\
    \hline
	Dog & $0.224_{\pm 0.011}$ & $0.699_{\pm 0.009}$ & $0.699_{\pm 0.003}$ & $0.699_{\pm 0.000}$ \\
	Fox & $0.614_{\pm 0.013}$ & $0.627_{\pm 0.005}$ & $0.732_{\pm 0.013}$ & $0.673_{\pm 0.059}$ \\
	Wolf & $0.663_{\pm 0.024}$ & $0.694_{\pm 0.013}$ & $0.633_{\pm 0.006}$ & $0.679_{\pm 0.016}$ \\
    \hline
    \end{tabular}%
  \label{tab:intraDog}%
\end{table}%

\begin{table}[htbp]
  \centering
  \small
  \caption{Intra-family DG results (mAP) of HRNet-w32~\cite{wang2020deep} on the test set of the Felidae family. Fel.=Felidae, Bob.=Bobcat, Che.=Cheetah, Jag.=Jaguar, K.C.=King Cheetah, Leo.=Leopard, Lio.=Lion, Pan.=Panther, S.L.=Snow Leopard, Tig.=Tiger.}
  \setlength{\tabcolsep}{0.001\linewidth}{
    \begin{tabular}{c|ccccccccccc}
    \hline
    \diagbox{Test}{Train} & \textbf{\textit{Fel.}}/Bob.  & \textbf{\textit{Fel.}}/Cat   & \textbf{\textit{Fel.}}/Che.  & \textbf{\textit{Fel.}}/Jag.  & \textbf{\textit{Fel.}}/K.C.  & \textbf{\textit{Fel.}}/Leo.  & \textbf{\textit{Fel.}}/Lio.  & \textbf{\textit{Fel.}}/Pan.  & \textbf{\textit{Fel.}}/S.L.  & \textbf{\textit{Fel.}}/Tig. & \textbf{Average}\\
    \hline
	Bob. & $\underset{\bm{\pm 0.005}}{\bm{0.631}}$	& $\underset{\pm 0.016}{0.714}$	& $\underset{\pm 0.004}{0.664}$	& $\underset{\pm 0.013}{0.674}$	& $\underset{\pm 0.013}{0.673}$	& $\underset{\pm 0.006}{0.663}$	& $\underset{\pm 0.016}{0.691}$	& $\underset{\pm 0.004}{0.623}$	& $\underset{\pm 0.005}{0.669}$	& $\underset{\pm 0.008}{0.713}$ & $\underset{\pm 0.026}{0.676}$ \\
	Cat  & $\underset{\pm 0.002}{0.638}$	& $\underset{\bm{\pm 0.004}}{\bm{0.332}}$	& $\underset{\pm 0.018}{0.625}$	& $\underset{\pm 0.010}{0.552}$	& $\underset{\pm 0.007}{0.629}$	& $\underset{\pm 0.009}{0.641}$	& $\underset{\pm 0.004}{0.601}$	& $\underset{\pm 0.010}{0.609}$	& $\underset{\pm 0.014}{0.582}$	& $\underset{\pm 0.007}{0.608}$ & $\underset{\pm 0.027}{0.609}$ \\
	Che. & $\underset{\pm 0.002}{0.715}$	& $\underset{\pm 0.012}{0.716}$	& $\underset{\bm{\pm 0.003}}{\bm{0.660}}$	& $\underset{\pm 0.013}{0.762}$	& $\underset{\pm 0.014}{0.731}$	& $\underset{\pm 0.010}{0.747}$	& $\underset{\pm 0.021}{0.734}$	& $\underset{\pm 0.008}{0.790}$	& $\underset{\pm 0.008}{0.713}$	& $\underset{\pm 0.008}{0.662}$ & $\underset{\pm 0.034}{0.730}$ \\
	Jag. & $\underset{\pm 0.005}{0.757}$	& $\underset{\pm 0.017}{0.770}$	& $\underset{\pm 0.006}{0.754}$	& $\underset{\bm{\pm 0.008}}{\bm{0.704}}$	& $\underset{\pm 0.004}{0.750}$	& $\underset{\pm 0.012}{0.759}$	& $\underset{\pm 0.013}{0.798}$	& $\underset{\pm 0.008}{0.724}$	& $\underset{\pm 0.011}{0.756}$	& $\underset{\pm 0.005}{0.734}$ & $\underset{\pm 0.020}{0.756}$ \\
	K.C. & $\underset{\pm 0.008}{0.961}$	& $\underset{\pm 0.035}{0.804}$	& $\underset{\pm 0.042}{0.692}$	& $\underset{\pm 0.028}{0.771}$	& $\underset{\bm{\pm 0.010}}{\bm{0.779}}$	& $\underset{\pm 0.008}{0.958}$	& $\underset{\pm 0.017}{0.713}$	& $\underset{\pm 0.026}{0.924}$	& $\underset{\pm 0.033}{0.864}$	& $\underset{\pm 0.016}{0.838}$ & $\underset{\pm 0.094}{0.836}$ \\
	Leo. & $\underset{\pm 0.005}{0.730}$	& $\underset{\pm 0.007}{0.697}$	& $\underset{\pm 0.014}{0.766}$	& $\underset{\pm 0.006}{0.741}$	& $\underset{\pm 0.005}{0.682}$	& $\underset{\bm{\pm 0.009}}{\bm{0.686}}$	& $\underset{\pm 0.012}{0.700}$	& $\underset{\pm 0.012}{0.705}$	& $\underset{\pm 0.010}{0.775}$	& $\underset{\pm 0.004}{0.744}$ & $\underset{\pm 0.031}{0.727}$ \\
	Lio. & $\underset{\pm 0.016}{0.623}$	& $\underset{\pm 0.023}{0.582}$	& $\underset{\pm 0.012}{0.639}$	& $\underset{\pm 0.010}{0.694}$	& $\underset{\pm 0.002}{0.688}$	& $\underset{\pm 0.018}{0.690}$	& $\underset{\bm{\pm 0.002}}{\bm{0.528}}$	& $\underset{\pm 0.007}{0.638}$	& $\underset{\pm 0.011}{0.630}$	& $\underset{\pm 0.024}{0.625}$ & $\underset{\pm 0.036}{0.645}$ \\
	Pan. & $\underset{\pm 0.020}{0.705}$	& $\underset{\pm 0.011}{0.722}$	& $\underset{\pm 0.020}{0.718}$	& $\underset{\pm 0.023}{0.720}$	& $\underset{\pm 0.013}{0.727}$	& $\underset{\pm 0.014}{0.785}$	& $\underset{\pm 0.026}{0.763}$	& $\underset{\bm{\pm 0.014}}{\bm{0.511}}$	& $\underset{\pm 0.004}{0.719}$	& $\underset{\pm 0.018}{0.684}$ & $\underset{\pm 0.028}{0.727}$ \\
	S.L. & $\underset{\pm 0.011}{0.792}$	& $\underset{\pm 0.008}{0.776}$	& $\underset{\pm 0.018}{0.810}$	& $\underset{\pm 0.019}{0.779}$	& $\underset{\pm 0.024}{0.790}$	& $\underset{\pm 0.004}{0.818}$	& $\underset{\pm 0.009}{0.821}$	& $\underset{\pm 0.015}{0.760}$	& $\underset{\bm{\pm 0.010}}{\bm{0.724}}$	& $\underset{\pm 0.012}{0.855}$ & $\underset{\pm 0.027}{0.800}$ \\
	Tig. & $\underset{\pm 0.008}{0.754}$	& $\underset{\pm 0.018}{0.741}$	& $\underset{\pm 0.012}{0.751}$	& $\underset{\pm 0.015}{0.715}$	& $\underset{\pm 0.021}{0.768}$	& $\underset{\pm 0.015}{0.753}$	& $\underset{\pm 0.005}{0.797}$	& $\underset{\pm 0.023}{0.848}$	& $\underset{\pm 0.011}{0.744}$	& $\underset{\bm{\pm 0.007}}{\bm{0.675}}$ & $\underset{\pm 0.036}{0.763}$ \\
    \hline
    \end{tabular}}%
  \label{tab:intraCat}%
\end{table}%

\subsubsection{Inter-family domain generalization}
\label{subsubsec:Inter-family}

\noindent \textbf{Settings} In this track, we select one family, \ie, Bovidae (Bov.), for training while using several other families, \ie, Cervidae (Cerv.), Equidae (Equ.), and Hominidae (Hom.), for testing. We follow the same training setting as the pretraining on ImageNet in the SL track.

\noindent \textbf{Results on the inter-family DG Track} The results are summarized in Table~\ref{tab:Inter}. We calculate the average score of all species in each family at the top row in each part (\ie, train and test). As can be seen, the model trained on the Bovidae family not only performs well on the species belonging to it but also on species belonging to the Cervidae family. In addition, its performance drops a little on the Equidae family while generalizing poorly on the Hominidae family. It is reasonable as the Bovidae family and the Cervidae family belong to the same order, \ie, Artiodactyla, and the biological proximity implies the similarity of pose distribution in these two families. For the Equidae family belonging to the same Euungulata clade but in adifferent order as Bovidae, their posture distributions may have a small difference, which explains the performance drop. For the Hominidae family belonging to a different clade, \ie, Primates clade, the learned knowledge about body keypoints from the Bovidae family is not suitable anymore. As shown in the last column in Table~\ref{tab:Inter}, where the model is trained on the Cercopithecidae family, it can generalize to Hominidae well since they belong to the same order, \ie, in line with the observation from the first column.

\begin{table}[htbp]
  \centering
  \small
  \caption{Inter-family DG results (mAP) of HRNet-w32~\cite{wang2020deep} on the different families' test set. Bov. = Bovidae, Ant. = Antelope, A.S. = Argali Sheep, Bis. = Bison, Buf. = Buffalo, She. = Sheep, Cer. = Cervidae, Der. = Deer, Moo. = Moose, Equ. = Equidae, Hor. = Horse, Zeb. = Zebra, Hom. = Hominidae, Chi. = Chihuahua, Gor. = Gorilla, Cerc. = Cercopithecidae, Alo. = Alouatta, Mon. = Monkey, N.N.M. = Noisy Night Monkey, S.M. = Spider Monkey, Uak. = Uakari.}
    \begin{tabular}{rcl|cl|cl|cl}
    \hline
          & Bov.  & $0.782_{\pm 0.002}$ & Bov.  & $0.782_{\pm 0.002}$ & Bov.  & $0.782_{\pm 0.002}$ & Cerc.  & $0.695_{\pm 0.007}$ \\
    \cmidrule{2-9}
          & Ant.  & $0.856_{\pm 0.001}$ & Ant.  & $0.856_{\pm 0.001}$ & Ant.  & $0.856_{\pm 0.001}$ & Alo.  & $0.697_{\pm 0.020}$ \\
          & A.S.  & $0.887_{\pm 0.006}$ & A.S.  & $0.887_{\pm 0.006}$ & A.S.  & $0.887_{\pm 0.006}$ & Mon.  & $0.725_{\pm 0.013}$ \\
    \multicolumn{1}{l}{train} & Bis.  & $0.643_{\pm 0.005}$ & Bis.  & $0.643_{\pm 0.005}$ & Bis.  & $0.643_{\pm 0.005}$ & N.N.M. & $0.750_{\pm 0.027}$ \\
          & Buf.  & $0.815_{\pm 0.004}$ & Buf.  & $0.815_{\pm 0.004}$ & Buf.  & $0.815_{\pm 0.004}$ & S.M.  & $0.581_{\pm 0.008}$ \\
          & Cow   & $0.737_{\pm 0.004}$ & Cow   & $0.737_{\pm 0.004}$ & Cow   & $0.737_{\pm 0.004}$ & Uak.  & $0.720_{\pm 0.009}$ \\
          & She.  & $0.754_{\pm 0.005}$ & She.  & $0.754_{\pm 0.005}$ & She.  & $0.754_{\pm 0.005}$ &       &  \\
    \hline
          & Cer.  & $0.641_{\pm 0.007}$ & Equ.  & $0.468_{\pm 0.019}$ & Hom.  & $0.015_{\pm 0.001}$ & Hom.  & $0.446_{\pm 0.007}$ \\
    \cmidrule{2-9}
    \multicolumn{1}{l}{test} & Der.  & $0.724_{\pm 0.004}$ & Hor.  & $0.618_{\pm 0.005}$ & Chi.  & $0.005_{\pm 0.000}$ & Chi.  & $0.446_{\pm 0.011}$ \\
          & Moo.  & $0.558_{\pm 0.010}$ & Zeb.  & $0.319_{\pm 0.035}$ & Gor.  & $0.026_{\pm 0.003}$ & Gor.  & $0.445_{\pm 0.011}$ \\
    \hline
    \end{tabular}%
    \vspace{-4 mm}
  \label{tab:Inter}%
\end{table}%

\subsubsection{Inter-family transfer learning and few-shot learning}

\noindent \textbf{Settings} We further evaluate the ability of models about inter-family generalization under a normal transfer learning setting and a few-shot learning setting. We follow the setting in Section~\ref{subsubsec:Inter-family} in these experiments, \ie, we pretrain the model using the Bovidae family and test on the species from the Cervidae, Equidae, and Hominidae families. \textit{\textbf{Transfer learning setting}}: We randomly select 140 images from each species as training set and use the others as test set. The models are then finetuned with each species's training images and evaluated with the test images. We use initial learning rate 1e-5 with linear weight decay schedule and Adam~\cite{kingma2014adam} optimizer for 35 epochs during the training. \textit{\textbf{Few-shot learning setting}}: Further, while keeping the test set unchanged, we select 20 images randomly in the training set of each species to finetune the models at the few-shot learning setting. The models are trained for 50 epochs with fixed learning rate 1e-5 and Adam~\cite{kingma2014adam} optimizer.

\noindent \textbf{Inter-family transfer learning and few-shot learning results} The results are summarized in Table~\ref{tab:fewshot_transfer}. 'Generalization' means directly test the pretrained models on each species' test set, as in Section~\ref{subsubsec:Inter-family}. With only 20 images for finetuning, the models' performance on the other species increase quickly, especially for the Zebra species, which have similar pose with the Bovidae family but different textures. With more images for finetuning, \ie, in the transfer learning setting, the performance further improves. \eg, the models' performance on the Chimpanzee species increase from 0.022 to 0.550. Such observation demonstrates that with more training data do help to improve the performance on new categories.
\begin{table}[htbp]
  \centering
  \small
  \caption{Inter-family's generalization, few-shot (20-shot), and transfer learning results (mAP).}
    \setlength{\tabcolsep}{0.008\linewidth}{
    \begin{tabular}{ccc|ccc}
    \hline
    Species & Setting & Performance & Species & Setting & Performance \\
    \hline
    \multirow{3}[1]{*}{Deer} & Generalization & $0.723_{\pm 0.036}$ & \multirow{3}[1]{*}{Moose} & Generalization & $0.587_{\pm 0.025}$ \\
          & Few-Shot & $0.742_{\pm 0.034}$ &       & Few-Shot & $0.648_{\pm 0.025}$ \\
          & Transfer & $0.751_{\pm 0.024}$ &       & Transfer & $0.726_{\pm 0.011}$ \\
    \hline
    \multirow{3}[0]{*}{Horse} & Generalization & $0.592_{\pm 0.047}$ & \multirow{3}[0]{*}{Zebra} & Generalization & $0.324_{\pm 0.021}$ \\
          & Few-Shot & $0.635_{\pm 0.034}$ &       & Few-Shot & $0.480_{\pm 0.029}$ \\
          & Transfer & $0.718_{\pm 0.023}$ &       & Transfer & $0.708_{\pm 0.024}$ \\
    \hline
    \multirow{3}[1]{*}{Chimpanzee} & Generalization & $0.009_{\pm 0.006}$ & \multirow{3}[1]{*}{Gorilla} & Generalization & $0.017_{\pm 0.006}$ \\
          & Few-Shot & $0.022_{\pm 0.010}$ &       & Few-Shot & $0.144_{\pm 0.121}$ \\
          & Transfer & $0.550_{\pm 0.032}$ &       & Transfer & $0.662_{\pm 0.039}$ \\
    \hline
    \end{tabular}}%
    \vspace{-6 mm}
  \label{tab:fewshot_transfer}%
\end{table}%

\subsubsection{Cross animal pose dataset evaluation}
\begin{table}[htbp]
  \centering
  \small
  \caption{Results of HRNet-w32 for cross animal pose dataset evaluation. Direct Test: evaluating the pretrained model from the source dataset on the target test set directly. Finetune\&Test: finetuning the pretrained model from the source dataset on the target dataset and then testing on the target test set. Train\&Test: training and testing the model on the target dataset.}
    \setlength{\tabcolsep}{0.006\linewidth}{
    \begin{tabular}{l|ccc}
    \hline
          & Direct Test (mAP)  & Finetune\&Test (mAP) & Train\&Test (mAP) \\
    \hline
    Animal-Pose Dataset~\cite{Cao_2019_ICCV} $\rightarrow$ AP-10K  & 0.424 &  0.722 & 0.727 \\
    AP-10K $\rightarrow$ Animal-Pose Dataset~\cite{Cao_2019_ICCV} & 0.913 & 0.935 & 0.932 \\
    \hline
    \end{tabular}}%
  \label{tab:cross}%
\end{table}%

\begin{figure}
    \centering
    \includegraphics[width=1\linewidth]{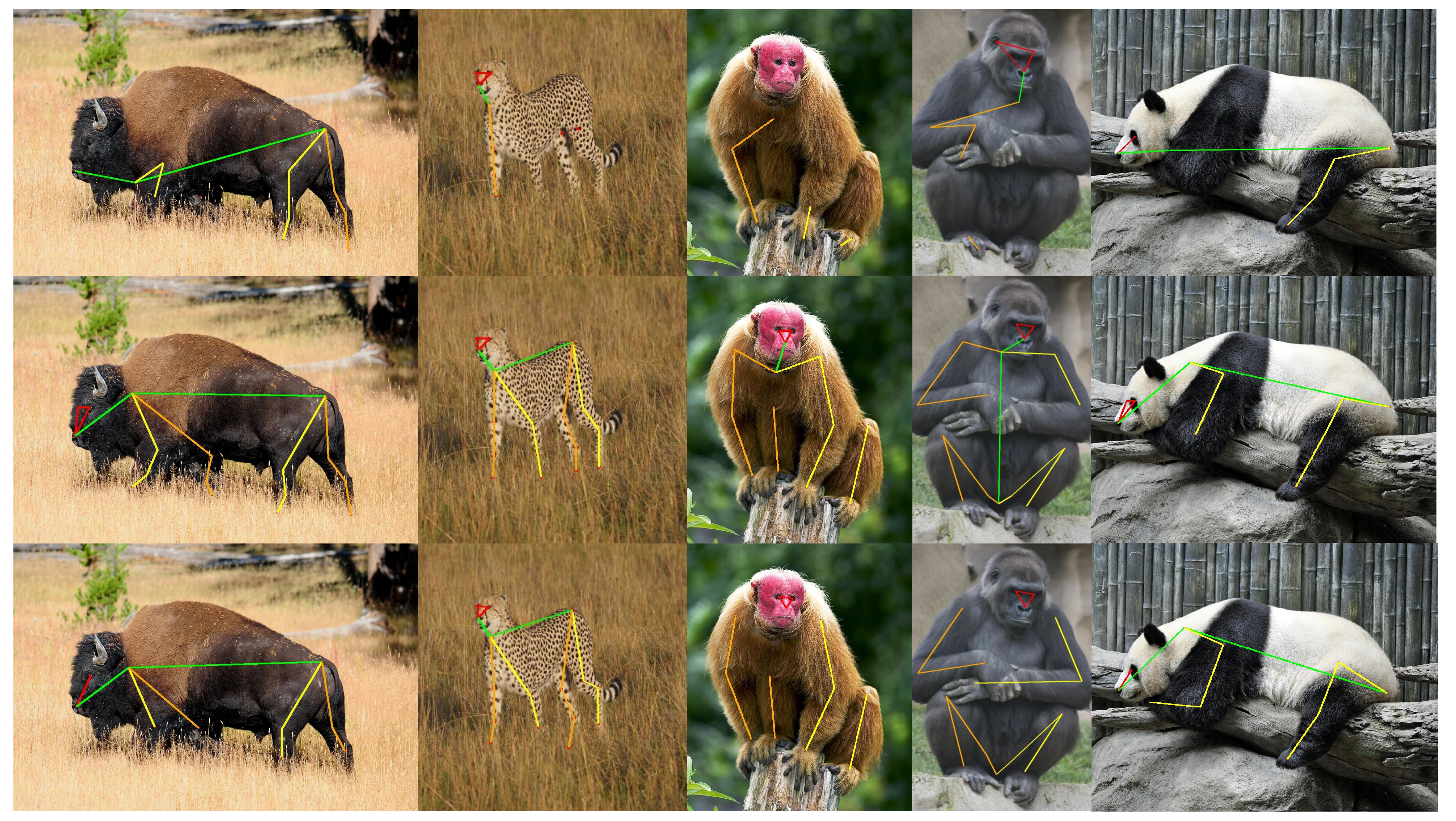}
    \vspace{-3 mm}
    \caption{Some qualitative results of HRNet-w32 trained on the Animal Pose dataset~\cite{Cao_2019_ICCV} (the first row) and our AP-10K dataet (the second row). The ground truth poses are shown in the last row.}
    \label{fig:Compare}
    \vspace{-5 mm}
\end{figure}

In this part, we evaluate the generalization ability of models trained on different animal pose datasets, including the Animal Pose dataset~\cite{Cao_2019_ICCV} and our AP-10K dataset. The results are shown in Table~\ref{tab:cross}. To be fair, we only compute the mAP on the 17 common keypoints shared by these two datasets. As can be seen, the model trained on the AP-10K dataset can generalize well on the Animal Pose dataset~\cite{Cao_2019_ICCV} but not vice versa. This is because the Animal Pose dataset~\cite{Cao_2019_ICCV} only has 5 animal species, which is far from enough to generalize well to other animal species. On the contrary, AP-10K has 60 animal species and can generalize well to animals, including those in the Animal Pose dataset~\cite{Cao_2019_ICCV}. Some visual results are presented in Figure~\ref{fig:Compare}. It is exciting to find that the model trained on our AP-10K can even predict the keypoints that are not annotated by annotators, \eg, the right eye of the bison and panda or the throat of the uakari and gorilla. Such phenomena imply that the model has learned good knowledge of animal body from the diverse species in our AP-10K dataset.

\section{Discussion}
\label{sec:discussion}

Based on the experimental results on the SL track, the CD-TL track, and the DG track, we make a step forward towards understanding the research questions posed in Section~\ref{sec:intro} and obtain empirical evidence to support the value of the proposed AP-10K dataset. The results also suggest that more efforts could be made to improve the efficiency of cross-domain transfer learning, address the domain gap issue between different intra- and inter-family species, as well as deal with the rare species in the long-tail distribution of animals. In addition to the study, AP-10K can also enable further research in different contexts. Firstly, since there are about 50k animal images without keypoint annotations while having family and species category labels, it is promising to study semi-supervised learning and self-supervised learning for animal pose estimation. Secondly, few-shot learning can also be an interesting direction, since images in AP-10K have a long tail distribution. How to deal with the rare species and improve the poses estimation performance remains underexplored. Thirdly, the study of inter-family or inter-species domain generalization deserves more effort, especially for those families or species that are not in the same clade or order. Although AP-10K is the largest dataset in this area, it is 10$\times$ smaller than those for human pose estimation, \eg, COCO \cite{lin2014microsoft}, we plan to increase its volume as well as species diversity in the future.

\section{Conclusion}

In this paper, we establish the first large-scale dataset for mammal pose estimation, \ie, AP-10K. It facilitates the study of new research questions due to its rich diversity in posture, scale, occlusion, and species of animals and organization structure following the taxonomic rank. We benchmark representative pose estimation methods on AP-10K and benefit from the empirical results to understand the representation ability of different models, the impact of pretraining on ImageNet or human pose dataset, the benefit of using diverse animal species for training, as well as intra- and inter-family model generalization. We hope AP-10K can pave the way for the follow-up study in this area. 

\textbf{Broad Impacts} 
AP-10K can potentially benefit the study of animal behavior understanding, zoology, and wildlife conservation. Nevertheless, it covers much fewer species than those in the real world, the generalization ability of trained models on it should be paid careful attention to. 

{
\textbf{Acknowledgement} 
The creation of the dataset is founded by the Innovation Capability Support Program of Shaanxi under the grant of Program No.2021TD-05 and the National Natural Science Foundation of China under the grant of No.62133012, No.61936006. Mr. Yufei Xu and Dr. Jing Zhang are supported by the ARC project FL-170100117.
}
\newpage

{\small
\bibliographystyle{ieee}
\bibliography{egbib}
}


\newpage

\appendix

\setcounter{figure}{0}
\renewcommand{\thefigure}{S\arabic{figure}}

\setcounter{table}{0}
\renewcommand{\thetable}{S\arabic{table}}

\section{Appendix}

\subsection{More quantitative and qualitative results}

\begin{table}[htbp]
  \centering
  \small
  \caption{The complete results of different models on the validation set of the SL track.}
    \begin{tabular}{c|clllll}
    \hline
    Pretraining on ImageNet  & \multicolumn{1}{c}{AP} & \multicolumn{1}{c}{AP$_{.5}$} & \multicolumn{1}{c}{AP$_{.75}$} & \multicolumn{1}{c}{AP$_{\text{M}}$} & \multicolumn{1}{c}{AP$_{\text{L}}$} \\
    \hline
    \multirow{1}[0]{*}{HRNet-w32~\cite{wang2020deep}}  & $0.738_{\pm 0.006}$	& $0.958_{\pm 0.004}$ & $0.808_{\pm 0.007}$ & $0.592_{\pm 0.074}$ & $0.743_{\pm 0.006}$ \\
    \multirow{1}[0]{*}{HRNet-w48~\cite{wang2020deep}}  & $0.744_{\pm 0.004}$	& $0.959_{\pm 0.002}$ &	$0.807_{\pm 0.006}$ & $0.589_{\pm 0.042}$ &	$0.748_{\pm 0.003}$ \\
    \multirow{1}[0]{*}{ResNet50~\cite{he2016deep}} & $0.699_{\pm 0.004}$ & $0.940_{\pm 0.003}$ & $0.760_{\pm 0.003}$ & $0.570_{\pm 0.078}$ & $0.703_{\pm 0.004}$ \\
    \multirow{1}[0]{*}{ResNet101~\cite{he2016deep}}  & $0.698_{\pm 0.002}$ & $0.943_{\pm 0.002}$ &	$0.754_{\pm 0.007}$ & $0.543_{\pm 0.070}$ &	$0.702_{\pm 0.001}$ \\
    \multirow{1}[0]{*}{Hourglass~\cite{newell2016stacked}}  & $0.729_{\pm 0.001}$ & $0.951_{\pm 0.002}$ &	$0.793_{\pm 0.002}$ & $0.606_{\pm 0.042}$ &	$0.733_{\pm 0.001}$ \\
    \hline
    Training from scratch & \multicolumn{1}{c}{AP} & \multicolumn{1}{c}{AP$_{.5}$} & \multicolumn{1}{c}{AP$_{.75}$} & \multicolumn{1}{c}{AP$_{\text{M}}$} & \multicolumn{1}{c}{AP$_{\text{L}}$} \\
    \hline
    \multirow{1}[0]{*}{HRNet-w32~\cite{wang2020deep}}  & $0.703_{\pm 0.002}$ & $0.940_{\pm 0.001}$ &	$0.764_{\pm 0.004}$ & $0.599_{\pm 0.020}$ &	$0.707_{\pm 0.003}$ \\
    \multirow{1}[0]{*}{HRNet-w48~\cite{wang2020deep}}    & $0.711_{\pm 0.003}$ & $0.943_{\pm 0.002}$ &	$0.771_{\pm 0.004}$	& $0.604_{\pm 0.047}$ &	$0.715_{\pm 0.003}$ \\
    \multirow{1}[0]{*}{ResNet50~\cite{he2016deep}}    & $0.646_{\pm 0.001}$ & $0.913_{\pm 0.006}$ & $0.694_{\pm 0.007}$ & $0.522_{\pm 0.053}$ & $0.649_{\pm 0.002}$ \\
    \multirow{1}[0]{*}{ResNet101~\cite{he2016deep}}    & $0.667_{\pm 0.002}$	& $0.924_{\pm 0.003}$ &	$0.715_{\pm 0.003}$	& $0.542_{\pm 0.042}$ &	$0.671_{\pm 0.002}$ \\
    \multirow{1}[0]{*}{Hourglass~\cite{newell2016stacked}}    & $0.686_{\pm 0.006}$	& $0.933_{\pm 0.003}$ &	$0.739_{\pm 0.011}$	& $0.591_{\pm 0.027}$ &	$0.690_{\pm 0.004}$ \\
    \hline
    \end{tabular}%
  \label{tab:fullResults}%
\end{table}%

\begin{table}[htbp]
  \centering
  \caption{Per-species results of HRNet-w32~\cite{wang2020deep} on the test set the SL Track at the setting of pretraining on ImageNet.}
    \begin{tabular}{c|lllll}
    \hline
    species & \multicolumn{1}{c}{AP} & \multicolumn{1}{c}{AP.5} & \multicolumn{1}{c}{AP.75} & \multicolumn{1}{c}{AP(M)} & \multicolumn{1}{c}{AP(L)} \\
	\hline
	Antelope	&	$0.856_{\pm 0.014}$	&	$0.969_{\pm 0.008}$	&	$0.924_{\pm 0.007}$	&	$0.732_{\pm 0.024}$	&	$0.869_{\pm 0.012}$		\\
	Argali Sheep	&	$0.871_{\pm 0.069}$	&	$0.966_{\pm 0.048}$	&	$0.933_{\pm 0.095}$	&	N/A	&	$0.914_{\pm 0.011}$		\\
	Bison	&	$0.723_{\pm 0.021}$	&	$0.941_{\pm 0.028}$	&	$0.804_{\pm 0.017}$	&	$0.487_{\pm 0.070}$	&	$0.788_{\pm 0.015}$		\\
	Buffalo	&	$0.834_{\pm 0.045}$	&	$0.978_{\pm 0.032}$	&	$0.895_{\pm 0.040}$	&	$0.141_{\pm 0.813}$	&	$0.844_{\pm 0.039}$		\\
	Cow	&	$0.757_{\pm 0.019}$	&	$0.979_{\pm 0.015}$	&	$0.824_{\pm 0.042}$	&	$0.551_{\pm 0.035}$	&	$0.765_{\pm 0.020}$		\\
	Sheep	&	$0.761_{\pm 0.011}$	&	$0.959_{\pm 0.015}$	&	$0.833_{\pm 0.010}$	&	$0.598_{\pm 0.154}$	&	$0.769_{\pm 0.008}$		\\
	Dog	&	$0.771_{\pm 0.013}$	&	$0.962_{\pm 0.006}$	&	$0.836_{\pm 0.003}$	&	$0.129_{\pm 0.803}$	&	$0.772_{\pm 0.011}$		\\
	Fox	&	$0.780_{\pm 0.002}$	&	$0.943_{\pm 0.025}$	&	$0.843_{\pm 0.014}$	&	N/A	&	$0.780_{\pm 0.002}$		\\
	Wolf	&	$0.778_{\pm 0.033}$	&	$0.972_{\pm 0.020}$	&	$0.864_{\pm 0.061}$	&	N/A	&	$0.779_{\pm 0.034}$		\\
	Beaver	&	$0.562_{\pm 0.004}$	&	$0.919_{\pm 0.021}$	&	$0.663_{\pm 0.015}$	&	$0.867_{\pm 0.047}$	&	$0.557_{\pm 0.008}$		\\
	Alouatta	&	$0.784_{\pm 0.038}$	&	$1.000_{\pm 0.000}$	&	$0.924_{\pm 0.061}$	&	N/A	&	$0.784_{\pm 0.038}$		\\
	Monkey	&	$0.690_{\pm 0.027}$	&	$0.961_{\pm 0.013}$	&	$0.769_{\pm 0.060}$	&	N/A	&	$0.690_{\pm 0.027}$		\\
	Noisy Night Monkey	&	$0.765_{\pm 0.026}$	&	$0.960_{\pm 0.000}$	&	$0.803_{\pm 0.068}$	&	N/A	&	$0.777_{\pm 0.040}$		\\
	Spider Monkey	&	$0.602_{\pm 0.046}$	&	$0.871_{\pm 0.024}$	&	$0.627_{\pm 0.072}$	&	$0.300_{\pm 0.920}$	&	$0.600_{\pm 0.048}$		\\
	Uakari	&	$0.829_{\pm 0.007}$	&	$1.000_{\pm 0.000}$	&	$0.976_{\pm 0.034}$	&	N/A	&	$0.829_{\pm 0.007}$		\\
	Deer	&	$0.794_{\pm 0.029}$	&	$0.950_{\pm 0.014}$	&	$0.847_{\pm 0.029}$	&	$0.157_{\pm 0.829}$	&	$0.811_{\pm 0.006}$		\\
	Moose	&	$0.776_{\pm 0.014}$	&	$0.973_{\pm 0.005}$	&	$0.843_{\pm 0.054}$	&	$0.838_{\pm 0.044}$	&	$0.773_{\pm 0.017}$		\\
	Hamster	&	$0.571_{\pm 0.064}$	&	$0.864_{\pm 0.042}$	&	$0.563_{\pm 0.079}$	&	$0.241_{\pm 0.175}$	&	$0.603_{\pm 0.051}$		\\
	Elephant	&	$0.704_{\pm 0.040}$	&	$0.918_{\pm 0.010}$	&	$0.774_{\pm 0.051}$	&	N/A	&	$0.703_{\pm 0.041}$		\\
	Horse	&	$0.749_{\pm 0.028}$	&	$0.934_{\pm 0.038}$	&	$0.798_{\pm 0.036}$	&	N/A	&	$0.764_{\pm 0.022}$		\\
	Zebra	&	$0.755_{\pm 0.026}$	&	$0.921_{\pm 0.013}$	&	$0.775_{\pm 0.041}$	&	N/A	&	$0.764_{\pm 0.019}$		\\
	Bobcat	&	$0.751_{\pm 0.055}$	&	$0.976_{\pm 0.034}$	&	$0.842_{\pm 0.057}$	&	N/A	&	$0.753_{\pm 0.057}$		\\
	Cat	&	$0.672_{\pm 0.036}$	&	$0.935_{\pm 0.006}$	&	$0.750_{\pm 0.067}$	&	N/A	&	$0.675_{\pm 0.037}$		\\
	Cheetah	&	$0.735_{\pm 0.029}$	&	$0.936_{\pm 0.010}$	&	$0.800_{\pm 0.026}$	&	N/A	&	$0.769_{\pm 0.032}$		\\
	Jaguar	&	$0.785_{\pm 0.035}$	&	$0.972_{\pm 0.020}$	&	$0.864_{\pm 0.082}$	&	$0.592_{\pm 0.170}$	&	$0.803_{\pm 0.024}$		\\
	King Cheetah	&	$0.848_{\pm 0.122}$	&	$0.944_{\pm 0.079}$	&	$0.944_{\pm 0.079}$	&	N/A	&	$0.848_{\pm 0.122}$		\\
	Leopard	&	$0.731_{\pm 0.009}$	&	$0.956_{\pm 0.034}$	&	$0.796_{\pm 0.047}$	&	N/A	&	$0.731_{\pm 0.009}$		\\
	Lion	&	$0.726_{\pm 0.017}$	&	$0.927_{\pm 0.024}$	&	$0.817_{\pm 0.038}$	&	N/A	&	$0.726_{\pm 0.017}$		\\
	Panther	&	$0.772_{\pm 0.048}$	&	$0.985_{\pm 0.021}$	&	$0.851_{\pm 0.027}$	&	$0.084_{\pm 0.769}$	&	$0.780_{\pm 0.051}$		\\
	Snow Leopard	&	$0.858_{\pm 0.042}$	&	$1.000_{\pm 0.000}$	&	$0.939_{\pm 0.048}$	&	$0.183_{\pm 0.849}$	&	$0.865_{\pm 0.035}$		\\
	Tiger	&	$0.811_{\pm 0.012}$	&	$0.973_{\pm 0.021}$	&	$0.872_{\pm 0.009}$	&	N/A	&	$0.812_{\pm 0.013}$		\\
	Giraffe	&	$0.803_{\pm 0.016}$	&	$0.958_{\pm 0.007}$	&	$0.842_{\pm 0.046}$	&	N/A	&	$0.808_{\pm 0.016}$		\\
	Hippo	&	$0.457_{\pm 0.062}$	&	$0.746_{\pm 0.089}$	&	$0.451_{\pm 0.064}$	&	$0.627_{\pm 0.273}$	&	$0.462_{\pm 0.052}$		\\
	Chimpanzee	&	$0.642_{\pm 0.009}$	&	$0.912_{\pm 0.034}$	&	$0.689_{\pm 0.044}$	&	$0.567_{\pm 0.126}$	&	$0.645_{\pm 0.018}$		\\
	Gorilla	&	$0.753_{\pm 0.010}$	&	$0.950_{\pm 0.029}$	&	$0.881_{\pm 0.038}$	&	N/A	&	$0.753_{\pm 0.010}$		\\
	Rabbit	&	$0.731_{\pm 0.013}$	&	$0.968_{\pm 0.003}$	&	$0.821_{\pm 0.023}$	&	N/A	&	$0.731_{\pm 0.013}$		\\
	Skunk	&	$0.558_{\pm 0.016}$	&	$0.864_{\pm 0.035}$	&	$0.652_{\pm 0.020}$	&	N/A	&	$0.562_{\pm 0.018}$		\\
	Mouse	&	$0.640_{\pm 0.015}$	&	$0.954_{\pm 0.033}$	&	$0.717_{\pm 0.036}$	&	$0.123_{\pm 0.796}$	&	$0.640_{\pm 0.013}$		\\
	Rat	&	$0.650_{\pm 0.031}$	&	$0.929_{\pm 0.015}$	&	$0.729_{\pm 0.052}$	&	$0.200_{\pm 0.849}$	&	$0.652_{\pm 0.030}$		\\
	Otter	&	$0.559_{\pm 0.017}$	&	$0.909_{\pm 0.031}$	&	$0.582_{\pm 0.050}$	&	N/A	&	$0.559_{\pm 0.017}$		\\
	Weasel	&	$0.716_{\pm 0.025}$	&	$0.955_{\pm 0.018}$	&	$0.790_{\pm 0.016}$	&	N/A	&	$0.717_{\pm 0.025}$		\\
	Raccoon	&	$0.629_{\pm 0.032}$	&	$0.954_{\pm 0.005}$	&	$0.675_{\pm 0.061}$	&	$0.401_{\pm 0.143}$	&	$0.651_{\pm 0.023}$		\\
	Rhino	&	$0.819_{\pm 0.021}$	&	$0.980_{\pm 0.000}$	&	$0.892_{\pm 0.048}$	&	N/A	&	$0.819_{\pm 0.021}$		\\
	Marmot	&	$0.784_{\pm 0.032}$	&	$1.000_{\pm 0.000}$	&	$0.860_{\pm 0.053}$	&	N/A	&	$0.784_{\pm 0.032}$		\\
	Squirrel	&	$0.845_{\pm 0.008}$	&	$1.000_{\pm 0.000}$	&	$0.943_{\pm 0.005}$	&	N/A	&	$0.845_{\pm 0.008}$		\\
	Pig	&	$0.642_{\pm 0.033}$	&	$0.964_{\pm 0.026}$	&	$0.677_{\pm 0.066}$	&	N/A	&	$0.640_{\pm 0.033}$		\\
	Black Bear	&	$0.662_{\pm 0.040}$	&	$0.967_{\pm 0.047}$	&	$0.865_{\pm 0.128}$	&	N/A	&	$0.662_{\pm 0.040}$		\\
	Brown Bear	&	$0.617_{\pm 0.029}$	&	$0.940_{\pm 0.011}$	&	$0.660_{\pm 0.051}$	&	$0.067_{\pm 0.759}$	&	$0.620_{\pm 0.026}$		\\
	Panda	&	$0.583_{\pm 0.008}$	&	$0.912_{\pm 0.020}$	&	$0.607_{\pm 0.082}$	&	N/A	&	$0.583_{\pm 0.009}$		\\
	Polar Bear	&	$0.631_{\pm 0.016}$	&	$0.948_{\pm 0.054}$	&	$0.659_{\pm 0.056}$	&	N/A	&	$0.631_{\pm 0.016}$		\\

    \hline
    \end{tabular}%
  \label{tab:per_species_test_pretrained_54}%
\end{table}%

\begin{table}[htbp]
  \centering
  \caption{Per-species results of HRNet-w32~\cite{wang2020deep} on the test set the SL Track at the setting of training from scratch.}
    \begin{tabular}{c|lllll}
    \hline
    species & \multicolumn{1}{c}{AP} & \multicolumn{1}{c}{AP.5} & \multicolumn{1}{c}{AP.75} & \multicolumn{1}{c}{AP(M)} & \multicolumn{1}{c}{AP(L)} \\
	\hline
	Antelope	&	$0.835_{\pm 0.019}$	&	$0.955_{\pm 0.021}$	&	$0.911_{\pm 0.021}$	&	$0.706_{\pm 0.026}$	&	$0.851_{\pm 0.020}$		\\
	Argali Sheep	&	$0.858_{\pm 0.065}$	&	$0.966_{\pm 0.048}$	&	$0.932_{\pm 0.097}$	&	N/A	&	$0.904_{\pm 0.005}$		\\
	Bison	&	$0.705_{\pm 0.026}$	&	$0.917_{\pm 0.036}$	&	$0.773_{\pm 0.020}$	&	$0.484_{\pm 0.115}$	&	$0.769_{\pm 0.014}$		\\
	Buffalo	&	$0.810_{\pm 0.044}$	&	$0.973_{\pm 0.039}$	&	$0.870_{\pm 0.060}$	&	$0.136_{\pm 0.806}$	&	$0.818_{\pm 0.043}$		\\
	Cow	&	$0.733_{\pm 0.018}$	&	$0.962_{\pm 0.017}$	&	$0.810_{\pm 0.009}$	&	$0.571_{\pm 0.091}$	&	$0.742_{\pm 0.018}$		\\
	Sheep	&	$0.741_{\pm 0.011}$	&	$0.950_{\pm 0.020}$	&	$0.797_{\pm 0.013}$	&	$0.593_{\pm 0.122}$	&	$0.748_{\pm 0.014}$		\\
	Dog	&	$0.743_{\pm 0.016}$	&	$0.960_{\pm 0.014}$	&	$0.807_{\pm 0.011}$	&	$0.100_{\pm 0.787}$	&	$0.745_{\pm 0.013}$		\\
	Fox	&	$0.753_{\pm 0.028}$	&	$0.917_{\pm 0.022}$	&	$0.792_{\pm 0.036}$	&	N/A	&	$0.753_{\pm 0.028}$		\\
	Wolf	&	$0.756_{\pm 0.030}$	&	$0.966_{\pm 0.025}$	&	$0.771_{\pm 0.053}$	&	N/A	&	$0.756_{\pm 0.030}$		\\
	Beaver	&	$0.483_{\pm 0.026}$	&	$0.848_{\pm 0.024}$	&	$0.465_{\pm 0.111}$	&	$0.384_{\pm 0.327}$	&	$0.490_{\pm 0.029}$		\\
	Alouatta	&	$0.745_{\pm 0.074}$	&	$0.970_{\pm 0.042}$	&	$0.896_{\pm 0.097}$	&	N/A	&	$0.745_{\pm 0.074}$		\\
	Monkey	&	$0.621_{\pm 0.025}$	&	$0.906_{\pm 0.030}$	&	$0.655_{\pm 0.056}$	&	N/A	&	$0.623_{\pm 0.027}$		\\
	Noisy Night Monkey	&	$0.722_{\pm 0.031}$	&	$0.935_{\pm 0.034}$	&	$0.764_{\pm 0.046}$	&	N/A	&	$0.721_{\pm 0.030}$		\\
	Spider Monkey	&	$0.514_{\pm 0.043}$	&	$0.824_{\pm 0.062}$	&	$0.513_{\pm 0.054}$	&	$0.300_{\pm 0.920}$	&	$0.513_{\pm 0.045}$		\\
	Uakari	&	$0.750_{\pm 0.004}$	&	$1.000_{\pm 0.000}$	&	$0.886_{\pm 0.029}$	&	N/A	&	$0.750_{\pm 0.004}$		\\
	Deer	&	$0.805_{\pm 0.032}$	&	$0.967_{\pm 0.034}$	&	$0.856_{\pm 0.022}$	&	$0.134_{\pm 0.806}$	&	$0.822_{\pm 0.011}$		\\
	Moose	&	$0.741_{\pm 0.007}$	&	$0.939_{\pm ±0.03}$	&	$0.815_{\pm 0.011}$	&	$0.776_{\pm 0.055}$	&	$0.739_{\pm 0.011}$		\\
	Hamster	&	$0.543_{\pm 0.031}$	&	$0.833_{\pm 0.054}$	&	$0.553_{\pm 0.017}$	&	$0.209_{\pm 0.086}$	&	$0.574_{\pm 0.019}$		\\
	Elephant	&	$0.664_{\pm 0.035}$	&	$0.881_{\pm 0.051}$	&	$0.725_{\pm 0.064}$	&	N/A	&	$0.662_{\pm 0.037}$		\\
	Horse	&	$0.759_{\pm 0.033}$	&	$0.928_{\pm 0.043}$	&	$0.802_{\pm 0.025}$	&	$0.023_{\pm 0.731}$	&	$0.765_{\pm 0.033}$		\\
	Zebra	&	$0.727_{\pm 0.025}$	&	$0.886_{\pm 0.014}$	&	$0.741_{\pm 0.027}$	&	N/A	&	$0.736_{\pm 0.018}$		\\
	Bobcat	&	$0.746_{\pm 0.044}$	&	$0.983_{\pm 0.024}$	&	$0.812_{\pm 0.058}$	&	N/A	&	$0.746_{\pm 0.044}$		\\
	Cat	&	$0.638_{\pm 0.030}$	&	$0.926_{\pm 0.023}$	&	$0.693_{\pm 0.053}$	&	N/A	&	$0.639_{\pm 0.031}$		\\
	Cheetah	&	$0.707_{\pm 0.043}$	&	$0.923_{\pm 0.018}$	&	$0.762_{\pm 0.074}$	&	N/A	&	$0.739_{\pm 0.038}$		\\
	Jaguar	&	$0.764_{\pm 0.038}$	&	$0.959_{\pm 0.032}$	&	$0.834_{\pm 0.039}$	&	$0.662_{\pm 0.225}$	&	$0.775_{\pm 0.027}$		\\
	King Cheetah	&	$0.813_{\pm 0.074}$	&	$0.944_{\pm 0.079}$	&	$0.944_{\pm 0.079}$	&	N/A	&	$0.813_{\pm 0.074}$		\\
	Leopard	&	$0.722_{\pm 0.026}$	&	$0.932_{\pm 0.020}$	&	$0.787_{\pm 0.024}$	&	N/A	&	$0.722_{\pm 0.026}$		\\
	Lion	&	$0.689_{\pm 0.012}$	&	$0.916_{\pm 0.018}$	&	$0.738_{\pm 0.017}$	&	N/A	&	$0.689_{\pm 0.012}$		\\
	Panther	&	$0.760_{\pm 0.024}$	&	$0.984_{\pm 0.023}$	&	$0.815_{\pm 0.024}$	&	$0.167_{\pm 0.829}$	&	$0.764_{\pm 0.025}$		\\
	Snow Leopard	&	$0.839_{\pm 0.036}$	&	$0.980_{\pm 0.028}$	&	$0.889_{\pm 0.005}$	&	$0.183_{\pm 0.849}$	&	$0.844_{\pm 0.030}$		\\
	Tiger	&	$0.764_{\pm 0.018}$	&	$0.971_{\pm 0.024}$	&	$0.828_{\pm 0.026}$	&	N/A	&	$0.764_{\pm 0.018}$		\\
	Giraffe	&	$0.784_{\pm 0.009}$	&	$0.942_{\pm 0.020}$	&	$0.819_{\pm 0.016}$	&	N/A	&	$0.789_{\pm 0.013}$		\\
	Hippo	&	$0.402_{\pm 0.026}$	&	$0.685_{\pm 0.030}$	&	$0.410_{\pm 0.016}$	&	$0.705_{\pm 0.206}$	&	$0.396_{\pm 0.023}$		\\
	Chimpanzee	&	$0.619_{\pm 0.031}$	&	$0.900_{\pm 0.049}$	&	$0.707_{\pm 0.026}$	&	$0.537_{\pm 0.140}$	&	$0.619_{\pm 0.042}$		\\
	Gorilla	&	$0.736_{\pm 0.026}$	&	$0.943_{\pm 0.038}$	&	$0.875_{\pm 0.077}$	&	N/A	&	$0.736_{\pm 0.026}$		\\
	Rabbit	&	$0.711_{\pm 0.045}$	&	$0.960_{\pm 0.008}$	&	$0.799_{\pm 0.072}$	&	N/A	&	$0.711_{\pm 0.045}$		\\
	Skunk	&	$0.499_{\pm 0.039}$	&	$0.830_{\pm 0.079}$	&	$0.514_{\pm 0.054}$	&	N/A	&	$0.504_{\pm 0.041}$		\\
	Mouse	&	$0.603_{\pm 0.035}$	&	$0.904_{\pm 0.011}$	&	$0.675_{\pm 0.032}$	&	$0.085_{\pm 0.768}$	&	$0.608_{\pm 0.036}$		\\
	Rat	&	$0.580_{\pm 0.043}$	&	$0.936_{\pm 0.048}$	&	$0.643_{\pm 0.059}$	&	$0.100_{\pm 0.787}$	&	$0.581_{\pm 0.046}$		\\
	Otter	&	$0.537_{\pm 0.017}$	&	$0.868_{\pm 0.021}$	&	$0.538_{\pm 0.018}$	&	N/A	&	$0.538_{\pm 0.017}$		\\
	Weasel	&	$0.649_{\pm 0.041}$	&	$0.945_{\pm 0.026}$	&	$0.691_{\pm 0.075}$	&	N/A	&	$0.649_{\pm 0.041}$		\\
	Raccoon	&	$0.585_{\pm 0.028}$	&	$0.940_{\pm 0.009}$	&	$0.590_{\pm 0.113}$	&	$0.363_{\pm 0.124}$	&	$0.601_{\pm 0.026}$		\\
	Rhino	&	$0.762_{\pm 0.048}$	&	$0.950_{\pm 0.021}$	&	$0.824_{\pm 0.067}$	&	N/A	&	$0.762_{\pm 0.048}$		\\
	Marmot	&	$0.768_{\pm 0.043}$	&	$1.000_{\pm 0.000}$	&	$0.843_{\pm 0.099}$	&	N/A	&	$0.768_{\pm 0.043}$		\\
	Squirrel	&	$0.849_{\pm 0.022}$	&	$1.000_{\pm 0.000}$	&	$0.962_{\pm 0.027}$	&	N/A	&	$0.849_{\pm 0.022}$		\\
	Pig	&	$0.612_{\pm 0.045}$	&	$0.907_{\pm 0.049}$	&	$0.648_{\pm 0.057}$	&	N/A	&	$0.610_{\pm 0.045}$		\\
	Black Bear	&	$0.684_{\pm 0.034}$	&	$0.967_{\pm 0.047}$	&	$0.861_{\pm 0.133}$	&	N/A	&	$0.684_{\pm 0.034}$		\\
	Brown Bear	&	$0.584_{\pm 0.025}$	&	$0.960_{\pm 0.014}$	&	$0.608_{\pm 0.067}$	&	$0.167_{\pm 0.834}$	&	$0.583_{\pm 0.023}$		\\
	Panda	&	$0.514_{\pm 0.056}$	&	$0.908_{\pm 0.034}$	&	$0.557_{\pm 0.099}$	&	N/A	&	$0.513_{\pm 0.056}$		\\
	Polar Bear	&	$0.619_{\pm 0.016}$	&	$0.928_{\pm 0.028}$	&	$0.650_{\pm 0.021}$	&	N/A	&	$0.619_{\pm 0.016}$		\\
    \hline
    \end{tabular}%
  \label{tab:per_species_test_scratch_54}%
\end{table}%

\begin{table}[htbp]
  \centering
  \small
  \caption{Per-species results of HRNet-w32~\cite{wang2020deep} on the \textbf{56 animals} test set the SL Track at the setting of pretraining on ImageNet.}
    \begin{tabular}{c|lllll}
    \hline
    species & \multicolumn{1}{c}{AP} & \multicolumn{1}{c}{AP$_{.5}$} & \multicolumn{1}{c}{AP$_{.75}$} & \multicolumn{1}{c}{AP$_{\text{M}}$} & \multicolumn{1}{c}{AP$_{\text{L}}$} \\
    \hline
    Antelope	&	$0.856_{\pm 0.017}$	&	$0.965_{\pm 0.011}$	&	$0.920_{\pm 0.019}$	&	$0.739_{\pm 0.016}$	&	$0.869_{\pm 0.014}$	\\
	Argali Sheep	&	$0.868_{\pm 0.061}$	&	$0.987_{\pm 0.019}$	&	$0.934_{\pm 0.093}$	&	$\text{N/A}$	&	$0.917_{\pm 0.012}$	\\
	Bison	&	$0.713_{\pm 0.020}$	&	$0.945_{\pm 0.016}$	&	$0.783_{\pm 0.023}$	&	$0.501_{\pm 0.111}$	&	$0.769_{\pm 0.008}$	\\
	Buffalo	&	$0.825_{\pm 0.050}$	&	$0.973_{\pm 0.026}$	&	$0.899_{\pm 0.049}$	&	$0.113_{\pm 0.796}$	&	$0.837_{\pm 0.043}$	\\
	Cow	&	$0.764_{\pm 0.023}$	&	$0.993_{\pm 0.009}$	&	$0.828_{\pm 0.047}$	&	$0.663_{\pm 0.052}$	&	$0.770_{\pm 0.025}$	\\
	Sheep	&	$0.749_{\pm 0.014}$	&	$0.950_{\pm 0.022}$	&	$0.834_{\pm 0.025}$	&	$0.591_{\pm 0.214}$	&	$0.754_{\pm 0.015}$	\\
	Chihuahua	&	$0.760_{\pm 0.021}$	&	$0.982_{\pm 0.025}$	&	$0.851_{\pm 0.042}$	& $\text{N/A}$ &	$0.760_{\pm 0.021}$	\\
	Collie	&	$0.775_{\pm 0.026}$	&	$0.973_{\pm 0.021}$	&	$0.877_{\pm 0.053}$	& $\text{N/A}$ &	$0.775_{\pm 0.026}$	\\
	Dalmatian	&	$0.759_{\pm 0.038}$	&	$0.960_{\pm 0.030}$	&	$0.776_{\pm 0.050}$	& $\text{N/A}$ &	$0.759_{\pm 0.038}$	\\
	Dog	&	$0.801_{\pm 0.020}$	&	$0.980_{\pm 0.014}$	&	$0.857_{\pm 0.040}$	&	$0.082_{\pm 0.765}$	&	$0.809_{\pm 0.013}$	\\
	Fox	&	$0.776_{\pm 0.009}$	&	$0.943_{\pm 0.009}$	&	$0.850_{\pm 0.030}$	& $\text{N/A}$ &	$0.776_{\pm 0.009}$	\\
	German Shepherd	&	$0.777_{\pm 0.045}$	&	$0.958_{\pm 0.011}$	&	$0.864_{\pm 0.063}$	& $\text{N/A}$ &	$0.777_{\pm 0.045}$	\\
	Wolf	&	$0.776_{\pm 0.025}$	&	$0.973_{\pm 0.019}$	&	$0.873_{\pm 0.028}$	& $\text{N/A}$ &	$0.776_{\pm 0.025}$	\\
	Beaver	&	$0.564_{\pm 0.046}$	&	$0.868_{\pm 0.047}$	&	$0.623_{\pm 0.128}$	&	$0.585_{\pm 0.300}$	&	$0.565_{\pm 0.045}$	\\
	Alouatta	&	$0.764_{\pm 0.034}$	&	$1.000_{\pm 0.000}$	&	$0.922_{\pm 0.061}$	& $\text{N/A}$ &	$0.764_{\pm 0.034}$	\\
	Monkey	&	$0.671_{\pm 0.034}$	&	$0.927_{\pm 0.024}$	&	$0.750_{\pm 0.074}$	& $\text{N/A}$ &	$0.671_{\pm 0.034}$	\\
	Noisy Night Monkey	&	$0.755_{\pm 0.049}$	&	$0.973_{\pm 0.019}$	&	$0.847_{\pm 0.081}$	&	$\text{N/A}$	&	$0.759_{\pm 0.055}$	\\
	Spider Monkey	&	$0.573_{\pm 0.044}$	&	$0.871_{\pm 0.036}$	&	$0.620_{\pm 0.059}$	&	$0.267_{\pm 0.899}$	&	$0.571_{\pm 0.045}$	\\
	Uakari	&	$0.809_{\pm 0.026}$	&	$1.000_{\pm 0.000}$	&	$1.000_{\pm 0.000}$	& $\text{N/A}$ &	$0.809_{\pm 0.026}$	\\
	Deer	&	$0.798_{\pm 0.019}$	&	$0.963_{\pm 0.024}$	&	$0.848_{\pm 0.022}$	&	$0.133_{\pm 0.805}$	&	$0.815_{\pm 0.006}$	\\
	Moose	&	$0.764_{\pm 0.015}$	&	$0.982_{\pm 0.013}$	&	$0.823_{\pm 0.025}$	&	$0.845_{\pm 0.103}$	&	$0.758_{\pm 0.014}$	\\
	Hamster	&	$0.595_{\pm 0.036}$	&	$0.875_{\pm 0.044}$	&	$0.628_{\pm 0.078}$	&	$0.288_{\pm 0.133}$	&	$0.626_{\pm 0.040}$	\\
	Elephant	&	$0.710_{\pm 0.030}$	&	$0.913_{\pm 0.048}$	&	$0.800_{\pm 0.049}$	&	$\text{N/A}$	&	$0.709_{\pm 0.031}$	\\
	Horse	&	$0.770_{\pm 0.043}$	&	$0.941_{\pm 0.031}$	&	$0.826_{\pm 0.069}$	&	$0.068_{\pm 0.758}$	&	$0.779_{\pm 0.038}$	\\
	Zebra	&	$0.748_{\pm 0.024}$	&	$0.910_{\pm 0.017}$	&	$0.760_{\pm 0.054}$	&	$\text{N/A}$	&	$0.761_{\pm 0.016}$	\\
	Bobcat	&	$0.765_{\pm 0.043}$	&	$0.983_{\pm 0.024}$	&	$0.845_{\pm 0.022}$	&	$\text{N/A}$	&	$0.768_{\pm 0.046}$	\\
	Cat	&	$0.697_{\pm 0.052}$	&	$0.962_{\pm 0.008}$	&	$0.807_{\pm 0.086}$	&	$\text{N/A}$	&	$0.697_{\pm 0.052}$	\\
	Cheetah	&	$0.756_{\pm 0.047}$	&	$0.942_{\pm 0.011}$	&	$0.821_{\pm 0.093}$	&	$\text{N/A}$	&	$0.785_{\pm 0.036}$	\\
	Jaguar	&	$0.793_{\pm 0.041}$	&	$0.984_{\pm 0.011}$	&	$0.866_{\pm 0.080}$	&	$0.638_{\pm 0.195}$	&	$0.809_{\pm 0.029}$	\\
	King Cheetah	&	$0.854_{\pm 0.085}$	&	$0.944_{\pm 0.079}$	&	$0.944_{\pm 0.079}$	& $\text{N/A}$ &	$0.854_{\pm 0.085}$	\\
	Leopard	&	$0.760_{\pm 0.005}$	&	$0.983_{\pm 0.024}$	&	$0.850_{\pm 0.017}$	& $\text{N/A}$ &	$0.760_{\pm 0.005}$	\\
	Lion	&	$0.716_{\pm 0.022}$	&	$0.922_{\pm 0.008}$	&	$0.787_{\pm 0.016}$	& $\text{N/A}$ &	$0.716_{\pm 0.022}$	\\
	Panther	&	$0.794_{\pm 0.037}$	&	$0.983_{\pm 0.024}$	&	$0.914_{\pm 0.037}$	&	$0.184_{\pm 0.837}$	&	$0.796_{\pm 0.038}$	\\
	Persian Cat	&	$0.592_{\pm 0.036}$	&	$0.930_{\pm 0.016}$	&	$0.640_{\pm 0.079}$	& $\text{N/A}$ &	$0.592_{\pm 0.036}$	\\
	Siamese Cat	&	$0.710_{\pm 0.053}$	&	$0.952_{\pm 0.037}$	&	$0.804_{\pm 0.054}$	& $\text{N/A}$ &	$0.710_{\pm 0.053}$	\\
	Snow Leopard	&	$0.860_{\pm 0.038}$	&	$1.000_{\pm 0.000}$	&	$0.980_{\pm 0.029}$	&	$0.217_{\pm 0.866}$	&	$0.866_{\pm 0.035}$	\\
	Tiger	&	$0.810_{\pm 0.014}$	&	$0.979_{\pm 0.015}$	&	$0.858_{\pm 0.019}$	& $\text{N/A}$ &	$0.810_{\pm 0.014}$	\\
	Giraffe	&	$0.799_{\pm 0.025}$	&	$0.949_{\pm 0.021}$	&	$0.840_{\pm 0.044}$	&	$\text{N/A}$	&	$0.805_{\pm 0.020}$	\\
	Hippo	&	$0.471_{\pm 0.077}$	&	$0.778_{\pm 0.082}$	&	$0.484_{\pm 0.056}$	&	$0.640_{\pm 0.283}$	&	$0.476_{\pm 0.077}$	\\
	Chimpanzee	&	$0.659_{\pm 0.027}$	&	$0.912_{\pm 0.034}$	&	$0.692_{\pm 0.021}$	&	$0.624_{\pm 0.057}$	&	$0.661_{\pm 0.035}$	\\
	Gorilla	&	$0.754_{\pm 0.014}$	&	$0.952_{\pm 0.037}$	&	$0.877_{\pm 0.013}$	& $\text{N/A}$ &	$0.754_{\pm 0.014}$	\\
	Rabbit	&	$0.729_{\pm 0.006}$	&	$0.990_{\pm 0.014}$	&	$0.830_{\pm 0.044}$	& $\text{N/A}$ &	$0.729_{\pm 0.006}$	\\
	Skunk	&	$0.576_{\pm 0.033}$	&	$0.874_{\pm 0.044}$	&	$0.651_{\pm 0.046}$	& $\text{N/A}$ &	$0.580_{\pm 0.035}$	\\
	Mouse	&	$0.641_{\pm 0.017}$	&	$0.954_{\pm 0.033}$	&	$0.716_{\pm 0.039}$	&	$0.100_{\pm 0.778}$	&	$0.644_{\pm 0.018}$	\\
	Rat	&	$0.647_{\pm 0.040}$	&	$0.930_{\pm 0.044}$	&	$0.747_{\pm 0.040}$	&	$0.133_{\pm 0.818}$	&	$0.647_{\pm 0.044}$	\\
	Otter	&	$0.567_{\pm 0.012}$	&	$0.897_{\pm 0.020}$	&	$0.594_{\pm 0.017}$	& $\text{N/A}$ &	$0.567_{\pm 0.012}$	\\
	Weasel	&	$0.694_{\pm 0.033}$	&	$0.935_{\pm 0.017}$	&	$0.781_{\pm 0.040}$	& $\text{N/A}$ &	$0.694_{\pm 0.033}$	\\
	Raccoon	&	$0.649_{\pm 0.041}$	&	$0.972_{\pm 0.021}$	&	$0.759_{\pm 0.035}$	&	$0.433_{\pm 0.128}$	&	$0.673_{\pm 0.035}$	\\
	Rhino	&	$0.800_{\pm 0.011}$	&	$0.980_{\pm 0.000}$	&	$0.864_{\pm 0.031}$	& $\text{N/A}$ &	$0.800_{\pm 0.011}$	\\
	Marmot	&	$0.792_{\pm 0.046}$	&	$0.967_{\pm 0.047}$	&	$0.917_{\pm 0.072}$	& $\text{N/A}$ &	$0.792_{\pm 0.046}$	\\
	Squirrel	&	$0.863_{\pm 0.008}$	&	$1.000_{\pm 0.000}$	&	$0.953_{\pm 0.013}$	& $\text{N/A}$ &	$0.863_{\pm 0.008}$	\\
	Pig	&	$0.645_{\pm 0.040}$	&	$0.928_{\pm 0.050}$	&	$0.706_{\pm 0.039}$	&	$\text{N/A}$	&	$0.644_{\pm 0.040}$	\\
	Black Bear	&	$0.685_{\pm 0.035}$	&	$1.000_{\pm 0.000}$	&	$0.853_{\pm 0.144}$	& $\text{N/A}$ &	$0.685_{\pm 0.035}$	\\
	Brown Bear	&	$0.612_{\pm 0.010}$	&	$0.960_{\pm 0.014}$	&	$0.680_{\pm 0.042}$	&	$0.068_{\pm 0.755}$	&	$0.615_{\pm 0.008}$	\\
	Panda	&	$0.572_{\pm 0.028}$	&	$0.935_{\pm 0.006}$	&	$0.604_{\pm 0.091}$	&	$\text{N/A}$	&	$0.571_{\pm 0.029}$	\\
	Polar Bear	&	$0.623_{\pm 0.001}$	&	$0.931_{\pm 0.053}$	&	$0.654_{\pm 0.015}$	& $\text{N/A}$ &	$0.623_{\pm 0.001}$	\\
    \hline
    \end{tabular}%
  \label{tab:per_species_test_pretrained_60}%
\end{table}%

\begin{table}[htbp]
  \centering
  \small
  \caption{Per-species results of HRNet-w32~\cite{wang2020deep} on the \textbf{56 animals} test set the SL Track at the setting of training from scratch.}
    \begin{tabular}{c|lllll}
    \hline
	species & \multicolumn{1}{c}{AP} & \multicolumn{1}{c}{AP$_{.5}$} & \multicolumn{1}{c}{AP$_{.75}$} & \multicolumn{1}{c}{AP$_{\text{M}}$} & \multicolumn{1}{c}{AP$_{\text{L}}$} \\
    \hline
    Antelope	&	$0.834_{\pm 0.019}$	&	$0.964_{\pm 0.011}$	&	$0.910_{\pm 0.013}$	&	$0.694_{\pm 0.026}$	&	$0.854_{\pm 0.017}$	\\
	Argali Sheep	&	$0.856_{\pm 0.087}$	&	$0.957_{\pm 0.061}$	&	$0.932_{\pm 0.096}$	&	\text{N/A}	&	$0.909_{\pm 0.014}$	\\
	Bison	&	$0.700_{\pm 0.041}$	&	$0.925_{\pm 0.033}$	&	$0.778_{\pm 0.034}$	&	$0.524_{\pm 0.138}$	&	$0.754_{\pm 0.025}$	\\
	Buffalo	&	$0.807_{\pm 0.038}$	&	$0.976_{\pm 0.033}$	&	$0.876_{\pm 0.051}$	&	$0.091_{\pm 0.779}$	&	$0.820_{\pm 0.033}$	\\
	Cow	&	$0.756_{\pm 0.013}$	&	$0.980_{\pm 0.014}$	&	$0.838_{\pm 0.018}$	&	$0.644_{\pm 0.080}$	&	$0.762_{\pm 0.013}$	\\
	Sheep	&	$0.749_{\pm 0.013}$	&	$0.958_{\pm 0.024}$	&	$0.831_{\pm 0.019}$	&	$0.616_{\pm 0.160}$	&	$0.755_{\pm 0.013}$	\\
	Chihuahua	&	$0.716_{\pm 0.029}$	&	$0.949_{\pm 0.002}$	&	$0.776_{\pm 0.071}$	&	\text{N/A}	&	$0.718_{\pm 0.026}$	\\
	Collie	&	$0.763_{\pm 0.010}$	&	$0.983_{\pm 0.012}$	&	$0.840_{\pm 0.035}$	&	\text{N/A}	&	$0.763_{\pm 0.010}$	\\
	Dalmatian	&	$0.723_{\pm 0.041}$	&	$0.916_{\pm 0.025}$	&	$0.731_{\pm 0.055}$	&	\text{N/A}	&	$0.723_{\pm 0.041}$	\\
	Dog	&	$0.782_{\pm 0.016}$	&	$0.964_{\pm 0.014}$	&	$0.850_{\pm 0.013}$	&	$0.149_{\pm 0.825}$	&	$0.791_{\pm 0.019}$	\\
	Fox	&	$0.738_{\pm 0.017}$	&	$0.939_{\pm 0.008}$	&	$0.803_{\pm 0.049}$	&	\text{N/A}	&	$0.738_{\pm 0.017}$	\\
	German Shepherd	&	$0.733_{\pm 0.041}$	&	$0.945_{\pm 0.005}$	&	$0.840_{\pm 0.061}$	&	\text{N/A}	&	$0.734_{\pm 0.040}$	\\
	Wolf	&	$0.751_{\pm 0.040}$	&	$0.971_{\pm 0.020}$	&	$0.794_{\pm 0.070}$	&	\text{N/A}	&	$0.751_{\pm 0.040}$	\\
	Beaver	&	$0.529_{\pm 0.034}$	&	$0.879_{\pm 0.023}$	&	$0.557_{\pm 0.087}$	&	$0.584_{\pm 0.245}$	&	$0.528_{\pm 0.033}$	\\
	Alouatta	&	$0.725_{\pm 0.045}$	&	$1.000_{\pm 0.000}$	&	$0.876_{\pm 0.120}$	&	\text{N/A}	&	$0.725_{\pm 0.045}$	\\
	Monkey	&	$0.629_{\pm 0.052}$	&	$0.930_{\pm 0.023}$	&	$0.686_{\pm 0.082}$	&	\text{N/A}	&	$0.629_{\pm 0.052}$	\\
	Noisy Night Monkey	&	$0.721_{\pm 0.004}$	&	$0.958_{\pm 0.001}$	&	$0.758_{\pm 0.040}$	&	\text{N/A}	&	$0.725_{\pm 0.010}$	\\
	Spider Monkey	&	$0.535_{\pm 0.029}$	&	$0.830_{\pm 0.048}$	&	$0.588_{\pm 0.033}$	&	$0.233_{\pm 0.873}$	&	$0.533_{\pm 0.031}$	\\
	Uakari	&	$0.763_{\pm 0.028}$	&	$1.000_{\pm 0.000}$	&	$0.921_{\pm 0.017}$	&	\text{N/A}	&	$0.763_{\pm 0.028}$	\\
	Deer	&	$0.792_{\pm 0.031}$	&	$0.960_{\pm 0.029}$	&	$0.836_{\pm 0.037}$	&	$0.130_{\pm 0.804}$	&	$0.808_{\pm 0.011}$	\\
	Moose	&	$0.759_{\pm 0.017}$	&	$0.959_{\pm 0.008}$	&	$0.844_{\pm 0.019}$	&	$0.691_{\pm 0.136}$	&	$0.761_{\pm 0.019}$	\\
	Hamster	&	$0.520_{\pm 0.043}$	&	$0.829_{\pm 0.079}$	&	$0.535_{\pm 0.035}$	&	$0.276_{\pm 0.138}$	&	$0.553_{\pm 0.024}$	\\
	Elephant	&	$0.672_{\pm 0.038}$	&	$0.883_{\pm 0.031}$	&	$0.772_{\pm 0.051}$	&	\text{N/A}	&	$0.671_{\pm 0.040}$	\\
	Horse	&	$0.757_{\pm 0.015}$	&	$0.935_{\pm 0.022}$	&	$0.785_{\pm 0.008}$	&	\text{N/A}	&	$0.768_{\pm 0.012}$	\\
	Zebra	&	$0.725_{\pm 0.024}$	&	$0.872_{\pm 0.024}$	&	$0.752_{\pm 0.038}$	&	\text{N/A}	&	$0.733_{\pm 0.020}$	\\
	Bobcat	&	$0.727_{\pm 0.054}$	&	$0.973_{\pm 0.021}$	&	$0.793_{\pm 0.088}$	&	\text{N/A}	&	$0.727_{\pm 0.054}$	\\
	Cat	&	$0.667_{\pm 0.040}$	&	$0.943_{\pm 0.006}$	&	$0.755_{\pm 0.040}$	&	\text{N/A}	&	$0.672_{\pm 0.041}$	\\
	Cheetah	&	$0.718_{\pm 0.020}$	&	$0.956_{\pm 0.010}$	&	$0.751_{\pm 0.083}$	&	\text{N/A}	&	$0.742_{\pm 0.031}$	\\
	Jaguar	&	$0.748_{\pm 0.016}$	&	$0.971_{\pm 0.021}$	&	$0.843_{\pm 0.021}$	&	$0.649_{\pm 0.230}$	&	$0.764_{\pm 0.013}$	\\
	King Cheetah	&	$0.798_{\pm 0.079}$	&	$0.944_{\pm 0.079}$	&	$0.832_{\pm 0.138}$	&	\text{N/A}	&	$0.798_{\pm 0.079}$	\\
	Leopard	&	$0.705_{\pm 0.004}$	&	$0.938_{\pm 0.030}$	&	$0.782_{\pm 0.062}$	&	\text{N/A}	&	$0.705_{\pm 0.004}$	\\
	Lion	&	$0.685_{\pm 0.026}$	&	$0.915_{\pm 0.019}$	&	$0.714_{\pm 0.046}$	&	\text{N/A}	&	$0.685_{\pm 0.026}$	\\
	Panther	&	$0.748_{\pm 0.041}$	&	$0.983_{\pm 0.024}$	&	$0.798_{\pm 0.044}$	&	$0.217_{\pm 0.863}$	&	$0.748_{\pm 0.040}$	\\
	Persian Cat	&	$0.559_{\pm 0.046}$	&	$0.858_{\pm 0.015}$	&	$0.535_{\pm 0.070}$	&	\text{N/A}	&	$0.559_{\pm 0.046}$	\\
	Siamese Cat	&	$0.677_{\pm 0.038}$	&	$0.965_{\pm 0.025}$	&	$0.713_{\pm 0.063}$	&	\text{N/A}	&	$0.677_{\pm 0.038}$	\\
	Snow Leopard	&	$0.826_{\pm 0.058}$	&	$1.000_{\pm 0.000}$	&	$0.866_{\pm 0.077}$	&	$0.167_{\pm 0.850}$	&	$0.831_{\pm 0.048}$	\\
	Tiger	&	$0.764_{\pm 0.021}$	&	$0.962_{\pm 0.034}$	&	$0.820_{\pm 0.008}$	&	\text{N/A}	&	$0.764_{\pm 0.021}$	\\
	Giraffe	&	$0.783_{\pm 0.015}$	&	$0.949_{\pm 0.015}$	&	$0.815_{\pm 0.027}$	&	\text{N/A}	&	$0.790_{\pm 0.012}$	\\
	Hippo	&	$0.383_{\pm 0.062}$	&	$0.659_{\pm 0.070}$	&	$0.383_{\pm 0.058}$	&	$0.615_{\pm 0.245}$	&	$0.384_{\pm 0.057}$	\\
	Chimpanzee	&	$0.625_{\pm 0.031}$	&	$0.944_{\pm 0.026}$	&	$0.709_{\pm 0.051}$	&	$0.547_{\pm 0.114}$	&	$0.629_{\pm 0.038}$	\\
	Gorilla	&	$0.736_{\pm 0.029}$	&	$0.942_{\pm 0.036}$	&	$0.836_{\pm 0.058}$	&	\text{N/A}	&	$0.736_{\pm 0.029}$	\\
	Rabbit	&	$0.712_{\pm 0.032}$	&	$0.979_{\pm 0.015}$	&	$0.791_{\pm 0.031}$	&	\text{N/A}	&	$0.712_{\pm 0.032}$	\\
	Skunk	&	$0.514_{\pm 0.032}$	&	$0.832_{\pm 0.013}$	&	$0.566_{\pm 0.089}$	&	\text{N/A}	&	$0.518_{\pm 0.032}$	\\
	Mouse	&	$0.608_{\pm 0.032}$	&	$0.928_{\pm 0.030}$	&	$0.657_{\pm 0.090}$	&	$0.124_{\pm 0.795}$	&	$0.607_{\pm 0.032}$	\\
	Rat	&	$0.588_{\pm 0.079}$	&	$0.935_{\pm 0.058}$	&	$0.623_{\pm 0.128}$	&	$0.133_{\pm 0.806}$	&	$0.586_{\pm 0.083}$	\\
	Otter	&	$0.537_{\pm 0.021}$	&	$0.889_{\pm 0.037}$	&	$0.516_{\pm 0.048}$	&	\text{N/A}	&	$0.537_{\pm 0.021}$	\\
	Weasel	&	$0.654_{\pm 0.027}$	&	$0.917_{\pm 0.012}$	&	$0.715_{\pm 0.019}$	&	\text{N/A}	&	$0.656_{\pm 0.028}$	\\
	Raccoon	&	$0.596_{\pm 0.014}$	&	$0.935_{\pm 0.017}$	&	$0.588_{\pm 0.063}$	&	$0.452_{\pm 0.178}$	&	$0.612_{\pm 0.010}$	\\
	Rhino	&	$0.773_{\pm 0.043}$	&	$0.945_{\pm 0.033}$	&	$0.854_{\pm 0.076}$	&	\text{N/A}	&	$0.773_{\pm 0.043}$	\\
	Marmot	&	$0.770_{\pm 0.037}$	&	$1.000_{\pm 0.000}$	&	$0.836_{\pm 0.064}$	&	\text{N/A}	&	$0.770_{\pm 0.037}$	\\
	Squirrel	&	$0.840_{\pm 0.019}$	&	$1.000_{\pm 0.000}$	&	$0.951_{\pm 0.037}$	&	\text{N/A}	&	$0.840_{\pm 0.019}$	\\
	Pig	&	$0.616_{\pm 0.060}$	&	$0.902_{\pm 0.062}$	&	$0.712_{\pm 0.095}$	&	\text{N/A}	&	$0.615_{\pm 0.059}$	\\
	Black Bear	&	$0.682_{\pm 0.066}$	&	$0.967_{\pm 0.047}$	&	$0.807_{\pm 0.099}$	&	\text{N/A}	&	$0.686_{\pm 0.061}$	\\
	Brown Bear	&	$0.589_{\pm 0.022}$	&	$0.973_{\pm 0.005}$	&	$0.621_{\pm 0.050}$	&	$0.101_{\pm 0.779}$	&	$0.589_{\pm 0.022}$	\\
	Panda	&	$0.525_{\pm 0.033}$	&	$0.921_{\pm 0.024}$	&	$0.498_{\pm 0.081}$	&	\text{N/A}	&	$0.524_{\pm 0.033}$	\\
	Polar Bear	&	$0.599_{\pm 0.022}$	&	$0.912_{\pm 0.037}$	&	$0.639_{\pm 0.015}$	&	\text{N/A}	&	$0.599_{\pm 0.022}$	\\
    \hline
    \end{tabular}%
  \label{tab:per_species_test_scratch_60}%
\end{table}%

In this Appendix, we provide the full evaluation results of HRNet-W32~\cite{wang2020deep}, HRNet-w48~\cite{wang2020deep}, SimpleBaseline~\cite{xiao2018simple} with ResNet50~\cite{he2016deep} and ResNet101~\cite{he2016deep} and Hourglass~\cite{newell2016stacked} in terms of different evaluation metrics on the validation set of the SL Track, as shown in Table~\ref{tab:fullResults}. It shows that the advanced network structure like HRNet~\cite{wang2020deep} outperforms the other backbones on most categories in terms of all evaluation metrics. It also demonstrates that pretraining on ImageNet can accelerate the convergence speed and achieve a better performance than training from scratch.

Besides, we also present the per species results of HRNet-W32~\cite{wang2020deep} on the test set the SL Track at the settings of ``Pretraining on ImageNet'' and ``Training from scratch'' in Table~\ref{tab:per_species_test_pretrained_54} and  Table~\ref{tab:per_species_test_scratch_54}, respectively. We also provide another test result which contains 56 animals for reference. \ie,  Dividing Dog into Chihuahua, Collie, Dog, dalmatian and German Shepherd, dividing Cat into cat and Perisian Cat and Siamese Cat. As we have mentioned in \ref{subsubsec:Intra-family},  Dog and Cat varies widely in appearance owning to human's cultivation for family pets.

Comparing Table~\ref{tab:per_species_test_pretrained_54} with the last rows in Table~\ref{tab:intraCow}, Table~\ref{tab:intraDog} and Table~\ref{tab:intraCat} in the paper, where all the scores represent test performance on seen species, we can find that using the same amount of training data but from more diverse species can help the model learning better feature representation and achieve better performance, \eg, the mAP for Sheep is 0.663 v.s. 0.761, which is obtained by the model trained only using the Bovidae family). This observation further confirms the value of our AP-10K dataset.


\subsection{Motivation}

\noindent \textbf{1. For what purpose was the dataset created? Was there a specific task in mind? Was there a specific gap that needed to be filled? Please provide a description.}

\textbf{A1:} AP-10K is created to facilitate research in the area of animal pose estimation. It is important to study several challenging questions in the context of more training data from diverse species are available, such as 1) how about the performance of different representative human pose models on the animal pose estimation task? 2) will the representation ability of a deep model benefit from training on a large-scale dataset with diverse species? 3) how about the impact of pretraining, \eg, on the ImageNet dataset \cite{deng2009imagenet} or human pose estimation dataset \cite{lin2014microsoft}, in the context of the large-scale of dataset with diverse species? and 4) how about the intra- and inter-family generalization ability of a model trained using data from specific species or family? However, previous datasets for animal pose estimation contain limited number of animal species. Therefore, it is impossible to study these questions using existing datasets as they contains at most 5 species, which is far from enough to get sound conclusion. By contrast, AP-10K has 23 family and 54 species and thus can help researchers to study these questions.

\noindent \textbf{2. Who created this dataset (e.g., which team, research group) and on behalf of which entity (e.g., company, institution, organization)?}

\textbf{A2:} AP-10K is created by the authors as well as some volunteer graduate students from Xidian University, including Jiarui Fan, Ziming Bai, Jinglong Zhao, Yan Liu, Yi Liu, Jiaming Li, Hanbo Sun, Chong Guo, Junwei Duan, and Xinyu Wang.

\noindent \textbf{3. Who funded the creation of the dataset? If there is an associated grant, please provide the name of the grantor and the grant name and number.}

\textbf{A3:} The creation of the dataset is founded by the Innovation Capability Support Program of Shaanxi under the grant of Program No.2021TD-05 and the National Natural Science Foundation of China under the grant of No.62133012, No.61936006.

\subsection{Composition}

\noindent \textbf{1. What do the instances that comprise the dataset represent (e.g., documents, photos, people, countries)? Are there multiple types of instances(e.g., movies, users, and ratings; people and interactions between them; nodes and edges)? Please provide a description.}

\textbf{A1:} AP-10K is comprised of images covering 54 animal species, which are categorized following the taxonomic rank, \ie, family and species. For each animal instance, its 17 keypoint annotations are provided, including Left Eye, Right Eye, Nose, Neck, Root of Tail, Left Shoulder, Left Elbow, Left Front Paw, Right Shoulder, Right Elbow, Right Front Paw, Left Hip, Left Knee, Left Back Paw, Right Hip, Right Knee, Right Back Paw.   

\noindent \textbf{2. How many instances are there in total (of each type, if appropriate)?}

\textbf{A2:} The AP-10K dataset contains 10,015 images and 13,028 instances with keypoint annotations. Besides, AP-10K contains extra 49,643 images, where each image is only annotated with family and species labels, without keypoint annotations.

\noindent \textbf{3. Does the dataset contain all possible instances or is it a sample (not necessarily random) of instances from a larger set? If the dataset is a sample, then what is the larger set? Is the sample representative of the larger set (e.g., geographic coverage)? If so, please describe how this representativeness was validated/verified. If it is not representative of the larger set, please describe why not (e.g., to cover a more diverse range of instances, because instances were withheld or unavailable).}

\textbf{A3:} AP-10K is a real-world sample of animals, including information about their poses. It is the largest dataset in the area of animal pose estimation, \eg, 10$\times$ more species compared with other previous datasets. Due to the diversity of real-world animal species, it is impossible to cover all instances of wild animals on a single dataset. The AP-10K dataset contains 23 typical families and 54 species, which follows the taxonomic rank, providing a more diverse set of instances than ever before and will facilitate further studies of animal pose estimation. Biological proximity exploited in AP-10K may be used to help the model extend to more species in the wild.

\noindent \textbf{4. What data does each instance consist of? “Raw” data (e.g., unprocessed text or images)or features? In either case, please provide a description.}

\textbf{A4:} Each instance consists of one animal with its location (bounding box), family and species labels, keypoint annotations, and the unprocessed image data.

\noindent \textbf{5. Is there a label or target associated with each instance? If so, please provide a description.}

\textbf{A5:} Yes. Each target is associated with labels following the COCO-style, which contain the instance id, image id, category information (family and species), the box area, whether is crowed or not, the number of keyponts, detailed keypoints information (location and category), as well as the background category.

\noindent \textbf{6. Is any information missing from individual instances? If so, please provide a description, explaining why this information is missing (e.g., because it was unavailable). This does not include intentionally removed information, but might include, e.g., redacted text.}

\textbf{A6:} Yes. Some instances may not have the complete keypoint annotation due to occlusion, blur, or small scale, similar to those instances in the COCO human pose dataset. Therefore, the location and visibility of these keypoints are marked zero following the COCO-style.

\noindent \textbf{7. Are relationships between individual instances made explicit (e.g., users’ movie ratings, social network links)? If so, please describe how these relationships are made explicit.}

\textbf{A7:} Yes. The instances' information are stored in the COCO-style, where the relationship between instances, \eg, whether two instances belong to the same category or are on the same image, can be queried using the \href{https://github.com/jin-s13/xtcocoapi}{COCO APIs}.

\noindent \textbf{8. Are there recommended data splits (e.g., training, development/validation, testing)? If so, please provide a description of these splits, explaining the rationale behind them.}

\textbf{A8:} Yes. We randomly split the dataset into disjoint train, validation, and test sets following the ratio of 7:1:2. The splits are done within each species to keep the origin distribution.

\noindent \textbf{9. Are there any errors, sources of noise, or redundancies in the dataset? If so, please provide a description.}

\textbf{A9:} Although we have carefully double-check the annotations, there may be some inaccurate keypoint annotations, \eg, small drifts in the annotation of keypoint locations. To figure out the error rate of AP-10K annotations, we re-examine the annotations, and the results are available in Table~\ref{tab:error_rate_table}.

\begin{table}[htbp]
  \centering
  \caption{Error rate in AP-10K.}
    \begin{tabular}{cccc}
     \hline
    \multicolumn{1}{l}{\textbf{Number of Images}} & \multicolumn{1}{l}{\textbf{Total keypoints}} & \multicolumn{1}{l}{\textbf{Mislabeled keypoints}} & \multicolumn{1}{l}{\textbf{Error rate}} \\
    \hline
    10015 & 130142 & 15    & 0.012\% \\
    \hline
    \end{tabular}%
  \label{tab:error_rate_table}%
\end{table}%

\noindent \textbf{10. Is the dataset self-contained, or does it link to or otherwise rely on external resources (e.g., websites, tweets, other datasets)? If it links to or relies on external resources, a) are there guarantees that they will exist, and remain constant, over time; b) are there official archival versions of the complete dataset (i.e., including the external resources as they existed at the time the dataset was created); c) are there any restrictions (e.g., licenses, fees) associated with any of the external resources that might apply to a future user? Please provide descriptions of all external resources and any restrictions associated with them, as well as links or other access points, as appropriate.}

\textbf{A10:} The AP-10K is comprised of the publicly available datasets, including African Wildlife~\cite{africanWildLife}, {Animal-Pose Dataset}~\cite{Cao_2019_ICCV}, {Animal Image Dataset(DOG, CAT and PANDA)}~\cite{animalDCP}, {Endangered Animals}~\cite{endangeredanimals}, {IUCN Animals Dataset}~\cite{IUCN}, {Animals with Attributes 2}~\cite{xian2018zero}, {Animals-5}~\cite{animal5}, {Animals-10}~\cite{animals10}, and {Wild Cats}~\cite{wcats}. These datasets are publicly avaiable and can be downloaded from their websites, \eg, the Animals-10 dataset follows the GPL v2 license while Endangered Animals follows the CC-BY-4.0 license. We appreciate the significant contribution of the authors to the research community. Table~\ref{tab:source_of_AP-10K} shows the number of images contained in the each dataset that contribute to the AP-10K dataset.

\begin{table}[htbp]
  \centering
  \caption{The source dataset adopted in AP-10K.}
    \begin{tabular}{c|c}
    \hline
    \textbf{Source Dataset} & \textbf{Number of Images} \\
    \hline
    African Wildlife~\cite{africanWildLife} & 1454 \\
    Animal Image Dataset(DOG, CAT and PANDA)~\cite{animalDCP} & 2753 \\
    Animal-Pose~\cite{Cao_2019_ICCV} & 970 \\
    Animals with Attributes 2~\cite{xian2018zero} & 31701 \\
    Animals-5~\cite{animal5} & 5110 \\
    Animals-10~\cite{animals10} & 14029 \\
    Endangered Animals~\cite{endangeredanimals} & 284 \\
    IUCN Animals Dataset~\cite{IUCN} & 1689 \\
    Wild Cats~\cite{wcats} & 1558 \\
    Internet & 110 \\
    \hline
    \end{tabular}%
  \label{tab:source_of_AP-10K}%
\end{table}%

\noindent \textbf{11. Does the dataset contain data that might be considered confidential (e.g., data that is protected by legal privilege or by doctorpatient confidentiality, data that includes the content of individuals non-public communications)? If so, please provide a description.}

\textbf{A11:} No.

\noindent \textbf{12. Does the dataset contain data that, if viewed directly, might be offensive, insulting, threatening, or might otherwise cause anxiety? If so, please describe why.}

\textbf{A12:} No.

\subsection{Collection Process}

\noindent \textbf{1. How was the data associated with each instance acquired? Was the data directly observable (e.g., raw text, movie ratings), reported by subjects (e.g., survey responses), or indirectly inferred/derived from other data (e.g., part-of-speech tags, model-based guesses for age or language)? If data was reported by subjects or indirectly inferred/derived from other data, was the data validated/verified? If so, please describe how.}

\textbf{A1:} The data associated with each instance are directly observable, as they are stored as the common COCO format and can be easily viewed via the \href{https://github.com/jin-s13/xtcocoapi}{COCO APIs}.

\noindent \textbf{2. What mechanisms or procedures were used to collect the data (e.g., hardware apparatus or sensor, manual human curation, software program, software API)? How were these mechanisms or procedures validated?}

\textbf{A2:} The images in AP-10K come from dataset publicly available datasets described above, which can be directly downloaded from their websites. 

\noindent \textbf{3. If the dataset is a sample from a larger set, what was the sampling strategy (e.g., deterministic, probabilistic with specific sampling probabilities)?}

\textbf{A3:} No.

\noindent \textbf{4. Who was involved in the data collection process (e.g., students, crowdworkers, contractors) and how were they compensated (e.g., how much were crowdworkers paid)?}

\textbf{A4:} The first author of this paper.

\noindent \textbf{5. Over what timeframe was the data collected? Does this timeframe match the creation timeframe of the data associated with the instances (e.g., recent crawl of old news articles)? If not, please describe the timeframe in which the data associated with the instances was created.}

\textbf{A5}: It took about 1 day to collect the data and about 3 months to complete organization and annotation, as each participant labelled the bonding boxes and keypoints about one hour per workday.

\subsection{Preprocessing/cleaning/labeling}

\noindent \textbf{1. Was any preprocessing/cleaning/labeling of the data done (e.g., discretization or bucketing, tokenization, part-of-speech tagging, SIFT feature extraction, removal of instances, processing of missing values)? If so, please provide a description. If not, you may skip the remainder of the questions in this section.}

\textbf{A1:} We remove replicated images by using aHash~\cite{ahash_algorithm} to detect similar images and manually checking. Then, images with heavy occlusion and logos are removed manually. These images are categorized into family and species, which are double checked by the annotators to ensure the image quality of AP-10K. Next, the annotators after training start to annotate the images.

During training, annotators first learned about the physiognomy, body structure and distribution of keypoints of the animals. Then, five images of each species were presented to annotators to annotate keypoints, which were used to assess their annotation quality. Annotators with good annotation quality were further trained on how to deal with the partial absence of the body due to occlusion and were involved in the subsequent annotation process. Annotators were asked to annotate all visible keypoints. For the occluded keypoints, they were asked to annotate keypoints whose location they could estimate based on body plan, pose, and the symmetry property of the body, where the length of occluded limbs or the location of occluded keypoints could be inferred from the visible limbs or keypoints. Other keypoints were left unlabeled. As shown in Figure 1, the right limb of the polar bear on the right could not be estimated accurately, so the annotator did not annotate it, while the right limb of the panda was annotated because it could be easily estimated from their pose and the symmetry property of the body.

To guarantee the annotation quality, we have adopted a sequential labeling strategy. Three rounds of cross-check and correction are conducted with both manual check and automatic check (according to specific rules, \eg, keypoints belonging to an instance are in the same bounding box) to reduce the possibility of mislabeling. 

\noindent \textbf{2. Was the “raw” data saved in addition to the preprocessed/cleaned/labeled data (e.g., to support unanticipated future uses)? If so, please provide a link or other access point to the “raw” data.}

\textbf{A2:} No.

\noindent \textbf{3. Is the software used to preprocess/clean/label the instances available? If so, please provide a link or other access point.}

\textbf{A3:} We use the open source labelling \href{https://github.com/wkentaro/labelme}{tool}.

\subsection{Uses}

\noindent \textbf{1. Has the dataset been used for any tasks already? If so, please provide a description.}

\textbf{A1:} No.

\noindent \textbf{2. Is there a repository that links to any or all papers or systems that use the dataset? If so, please provide a link or other access point.}

\textbf{A2:} N/A.

\noindent \textbf{3. What (other) tasks could the dataset be used for?}

\textbf{A3:} AP-10K can be used for the research of animal pose estimation. Besides, it can also be used for specific machine learning topics such as few-shot learning, domain generalization, self-supervised learning. Please see the Discussion part in the paper. 

\noindent \textbf{4. Is there anything about the composition of the dataset or the way it was collected and preprocessed/cleaned/labeled that might impact future uses? For example, is there anything that a future user might need to know to avoid uses that could result in unfair treatment of individuals or groups (e.g., stereotyping, quality of service issues) or other undesirable harms (e.g., financial harms, legal risks) If so, please provide a description. Is there anything a future user could do to mitigate these undesirable harms?}

\textbf{A4:} No.

\noindent \textbf{5. Are there tasks for which the dataset should not be used? If so, please provide a description.}

\textbf{A5:} No.

\subsection{Distribution}

\noindent \textbf{1. Will the dataset be distributed to third parties outside of the entity (e.g., company, institution, organization) on behalf of which the dataset was created? If so, please provide a description.}

\textbf{A1:} Yes. The dataset will be made publicly available to the research community.

\noindent \textbf{2. How will the dataset will be distributed (e.g., tarball on website, API, GitHub)? Does the dataset have a digital object identifier (DOI)?}

\textbf{A2:} It will be publicly available on the project website at \href{https://github.com/AlexTheBad/AP10k}{GitHub}.

\noindent \textbf{3. When will the dataset be distributed?}

\textbf{A3:} The dataset will be distributed once the paper is accepted after peer-review.

\noindent \textbf{4. Will the dataset be distributed under a copyright or other intellectual property (IP) license, and/or under applicable terms of use (ToU)? If so, please describe this license and/or ToU, and provide a link or other access point to, or otherwise reproduce, any relevant licensing terms or ToU, as well as any fees associated with these restrictions.}

\textbf{A4:} It will be distributed under the MIT licence.

\noindent \textbf{5. Have any third parties imposed IP-based or other restrictions on the data associated with the instances? If so, please describe these restrictions, and provide a link or other access point to, or otherwise reproduce, any relevant licensing terms, as well as any fees associated with these restrictions.}

\textbf{A5:} No.

\noindent \textbf{6. Do any export controls or other regulatory restrictions apply to the dataset or to individual instances? If so, please describe these restrictions, and provide a link or other access point to, or otherwise reproduce, any supporting documentation.}

\textbf{A6:} No.

\subsection{Maintenance}

\noindent \textbf{1. Who will be supporting/hosting/maintaining the dataset?}

\textbf{A1:} The authors.

\noindent \textbf{2. How can the owner/curator/manager of the dataset be contacted (e.g., email address)?}

\textbf{A2:} They can be contacted via email available on the project website.

\noindent \textbf{3. Is there an erratum? If so, please provide a link or other access point.}

\textbf{A3:} No. 

\noindent \textbf{4. Will the dataset be updated (e.g., to correct labeling errors, add new instances, delete instances)? If so, please describe how often, by whom, and how updates will be communicated to users (e.g., mailing list, GitHub)?}

\textbf{A4:} No. We have carefully double checked the annotations to reduce the labeling errors. There may be a very few of labeling errors, which can be treated as noise.

\noindent \textbf{5. Will older versions of the dataset continue to be supported/hosted/maintained? If so, please describe how. If not, please describe how its obsolescence will be communicated to users.}

\textbf{A5:} N/A.

\noindent \textbf{6. If others want to extend/augment/build on/contribute to the dataset, is there a mechanism for them to do so? If so, please provide a description. Will these contributions be validated/verified? If so, please describe how. If not, why not? Is there a process for communicating/distributing these contributions to other users? If so, please provide a description.}

\textbf{A6:} N/A. 

\begin{figure}
    \centering
    \includegraphics[width=1\linewidth]{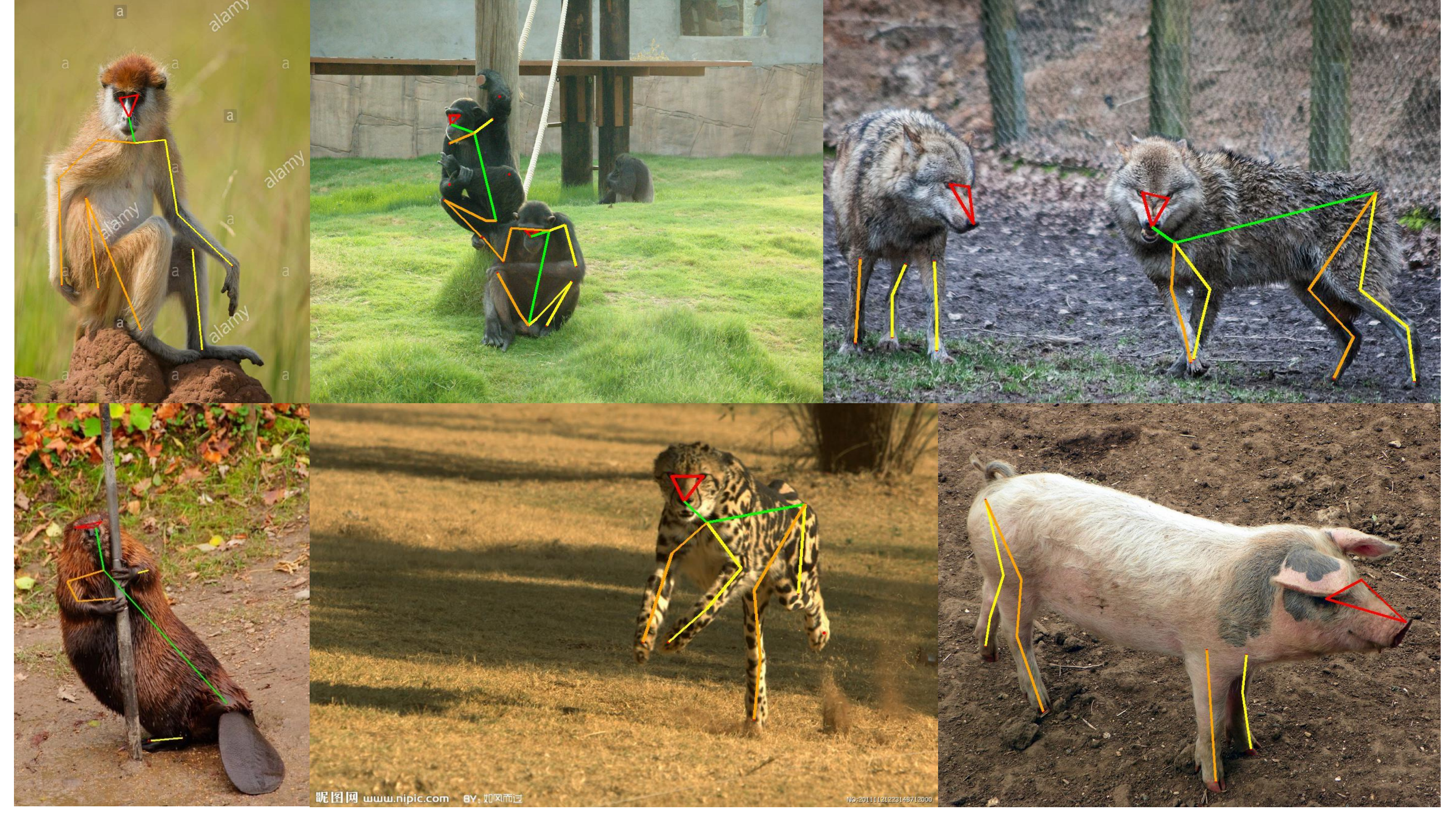}
    \caption{representative examples of failures of the pose estimation in our AP-10K.
    }
    \label{fig:failure_of_pose}
\end{figure}

\begin{figure}
    \centering
    \includegraphics[width=1\linewidth]{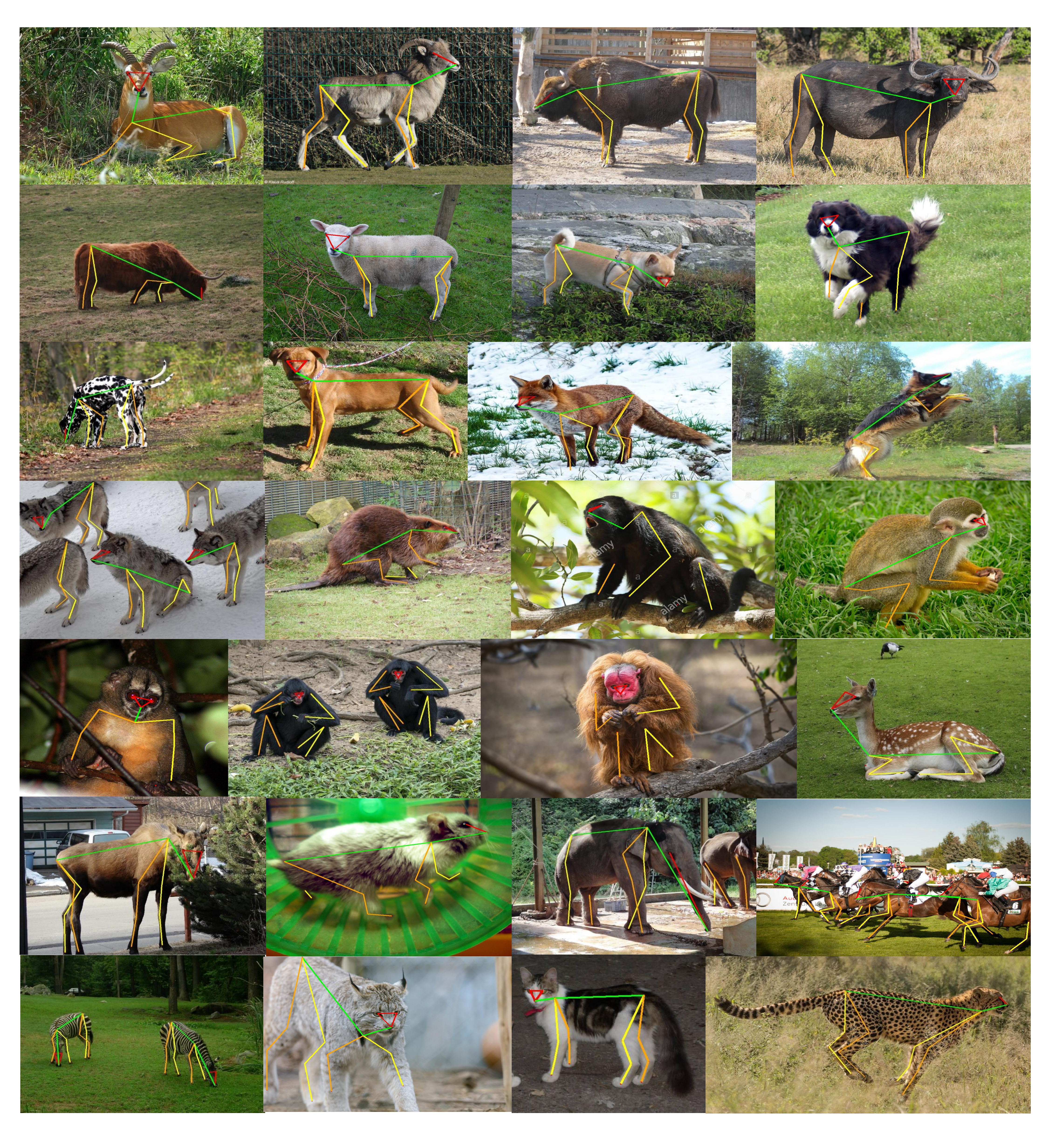}
    \caption{Some visual examples of animal species and annotations in our AP-10K.
    }
    \label{fig:up_28species}
\end{figure}

\begin{figure}
    \centering
    \includegraphics[width=1\linewidth]{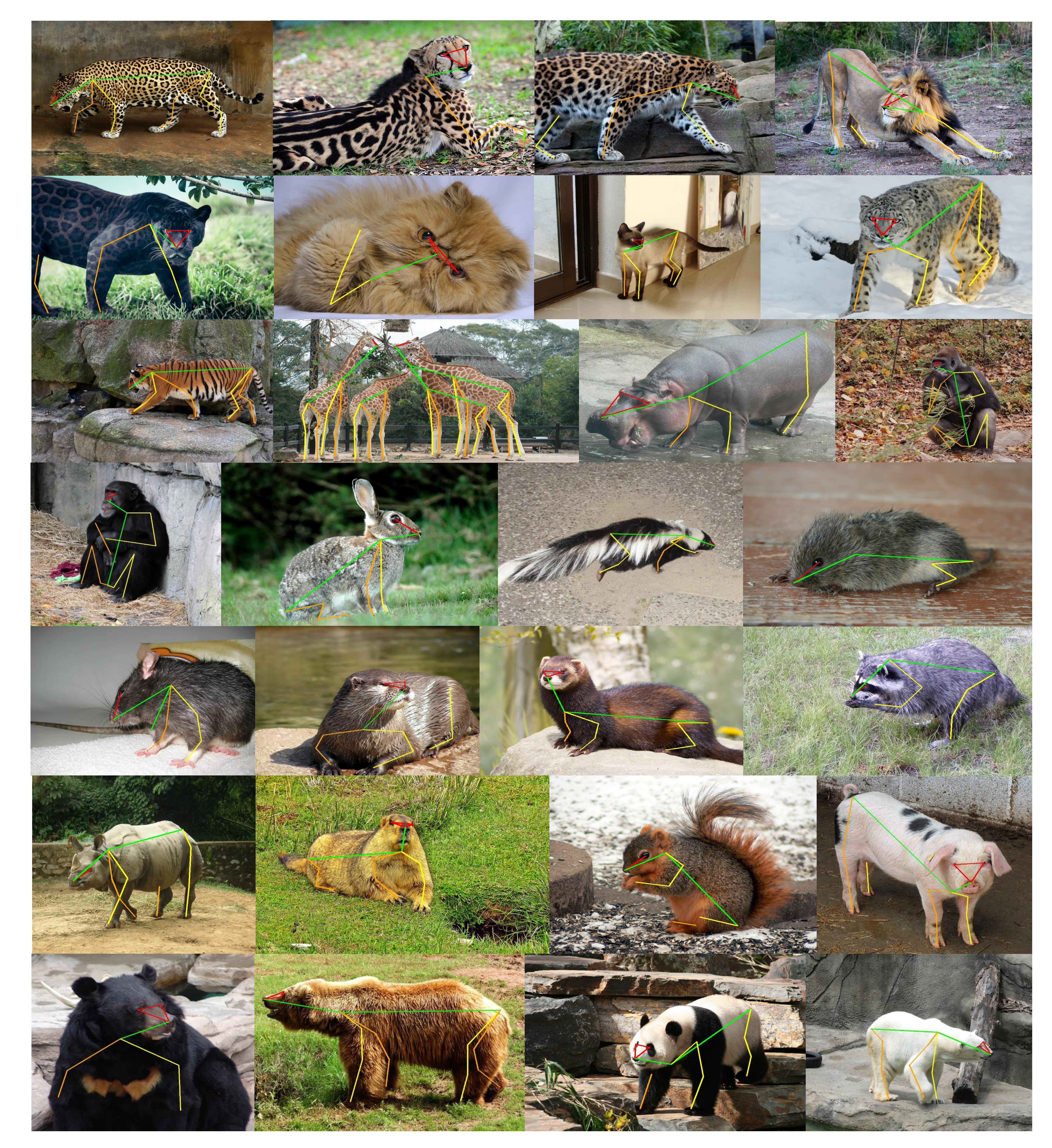}
    \caption{Some visual examples of animal species and annotations in our AP-10K.
    }
    \label{fig:down_28species}
\end{figure}

\begin{figure}
    \centering
    \includegraphics[width=1\linewidth]{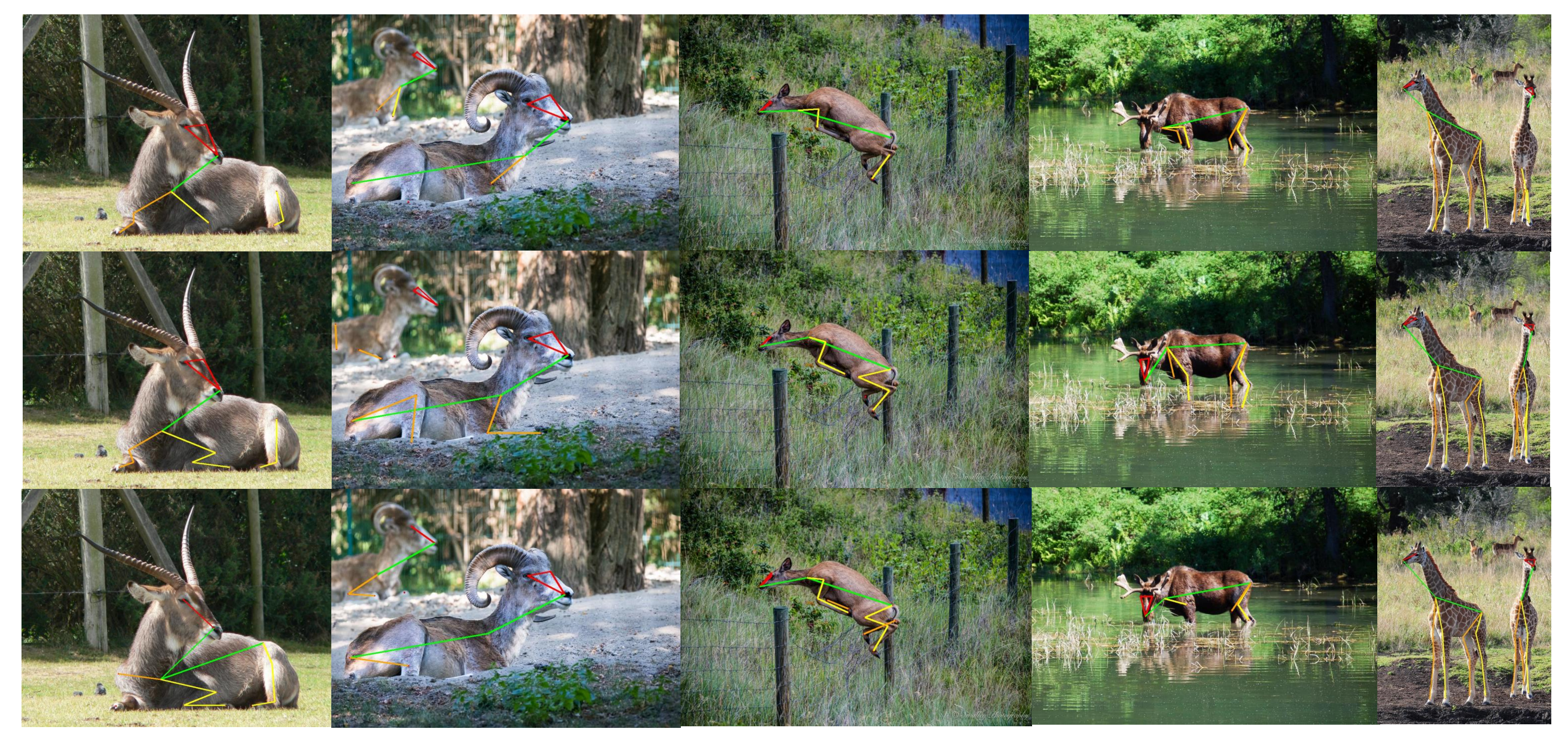}
    \caption{More qualitative results of HRNet-w32 trained on the Animal Pose dataset~\cite{Cao_2019_ICCV} (the first row) and our AP-10K dataet (the second row). The ground truth poses are shown in the last row. These animals are Antelope, Argali Sheep, Deer, Moose, and Giraffe, respectively.
    }
    \label{fig:appendix_pose_1}
\end{figure}

\begin{figure}
    \centering
    \includegraphics[width=1\linewidth]{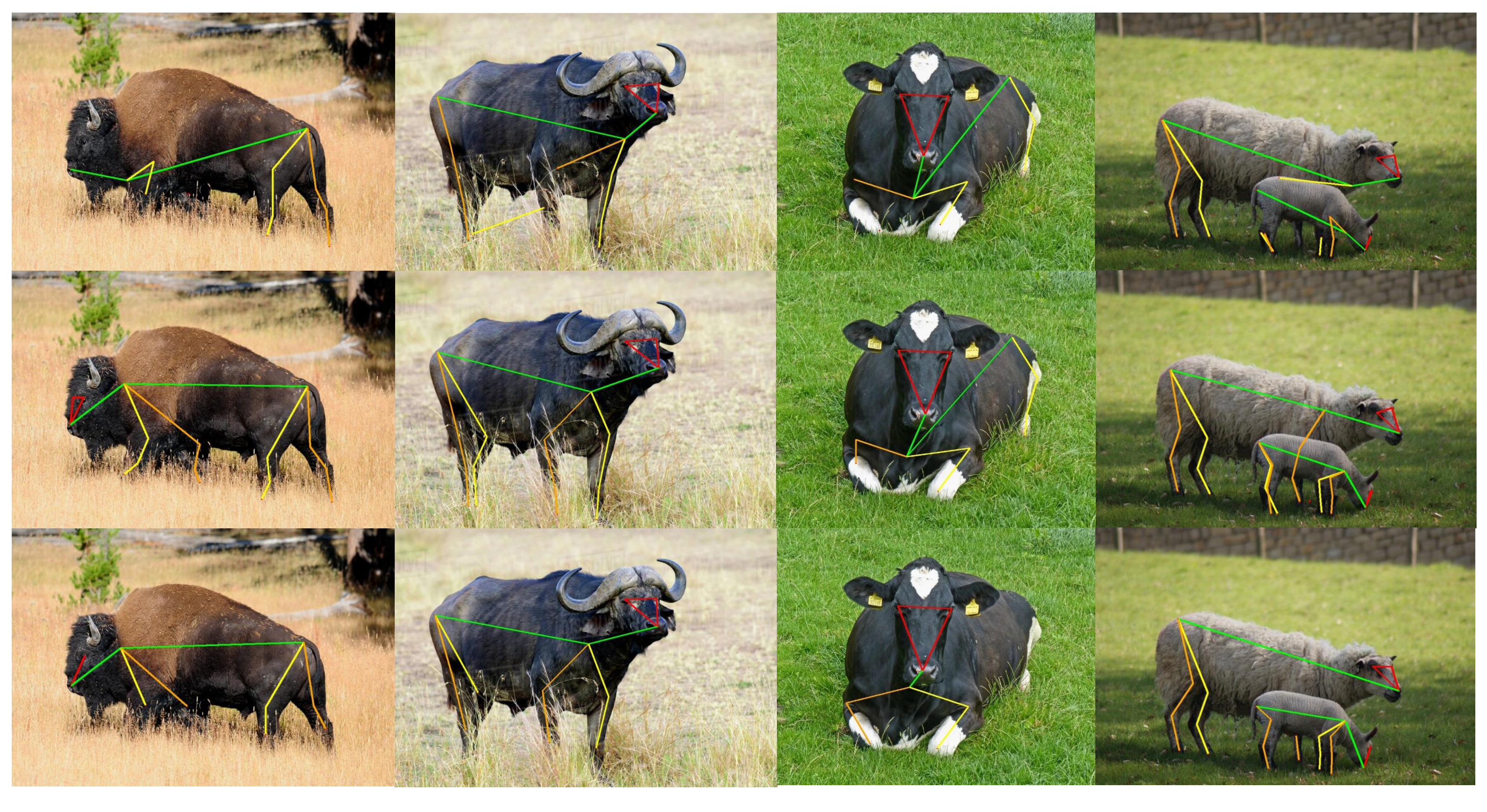}
    \caption{More qualitative results of HRNet-w32 trained on the Animal Pose dataset~\cite{Cao_2019_ICCV} (the first row) and our AP-10K dataet (the second row). The ground truth poses are shown in the last row. These animals are Bison, Buffalo, Cow, and Sheep, respectively.
    }
    \label{fig:appendix_pose_2}
\end{figure}

\begin{figure}
    \centering
    \includegraphics[width=1\linewidth]{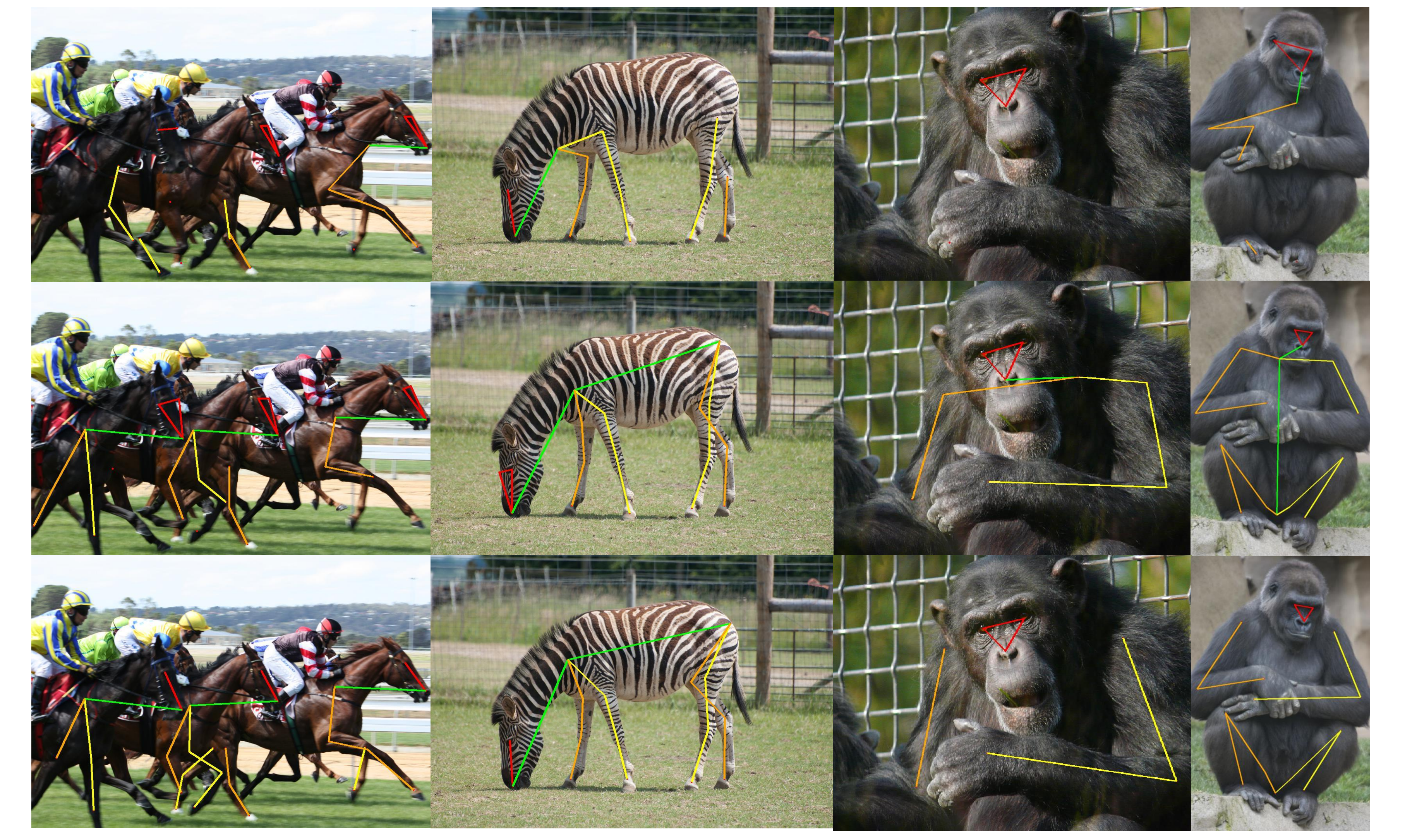}
     \caption{More qualitative results of HRNet-w32 trained on the Animal Pose dataset~\cite{Cao_2019_ICCV} (the first row) and our AP-10K dataet (the second row). The ground truth poses are shown in the last row. These animals are Horse, Zebra, Chimpanzee, and Gorilla, respectively.
    }
    \label{fig:appendix_pose_3}
\end{figure}

\begin{figure}
    \centering
    \includegraphics[width=1\linewidth]{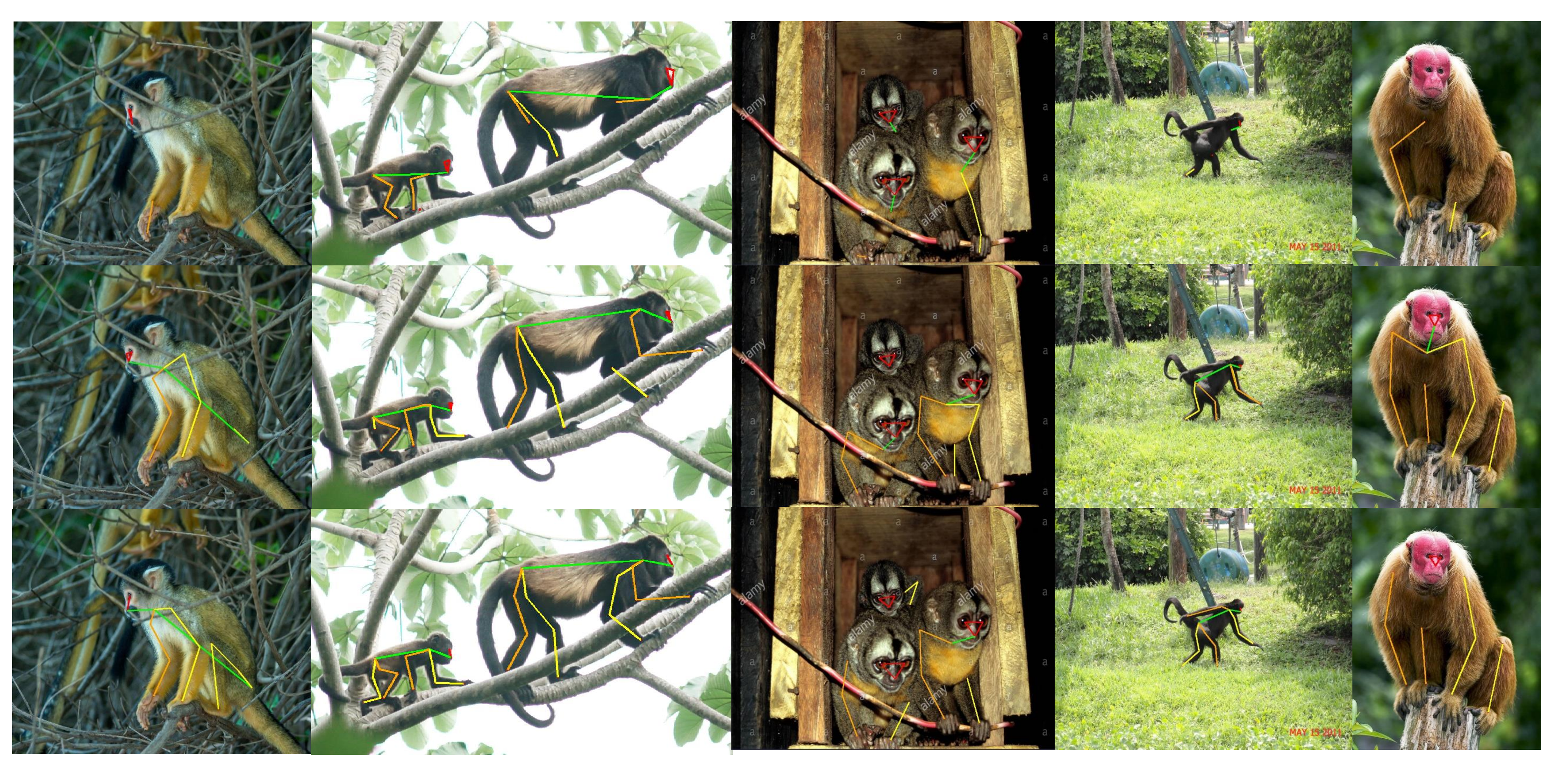}
     \caption{More qualitative results of HRNet-w32 trained on the Animal Pose dataset~\cite{Cao_2019_ICCV} (the first row) and our AP-10K dataet (the second row). The ground truth poses are shown in the last row. These animals are Monkey, Alouatta, Noisy Night Monkey, Spider Monkey, and Uakari, respectively.
    }
    \label{fig:appendix_pose_4}
\end{figure}

\begin{figure}
    \centering
    \includegraphics[width=1\linewidth]{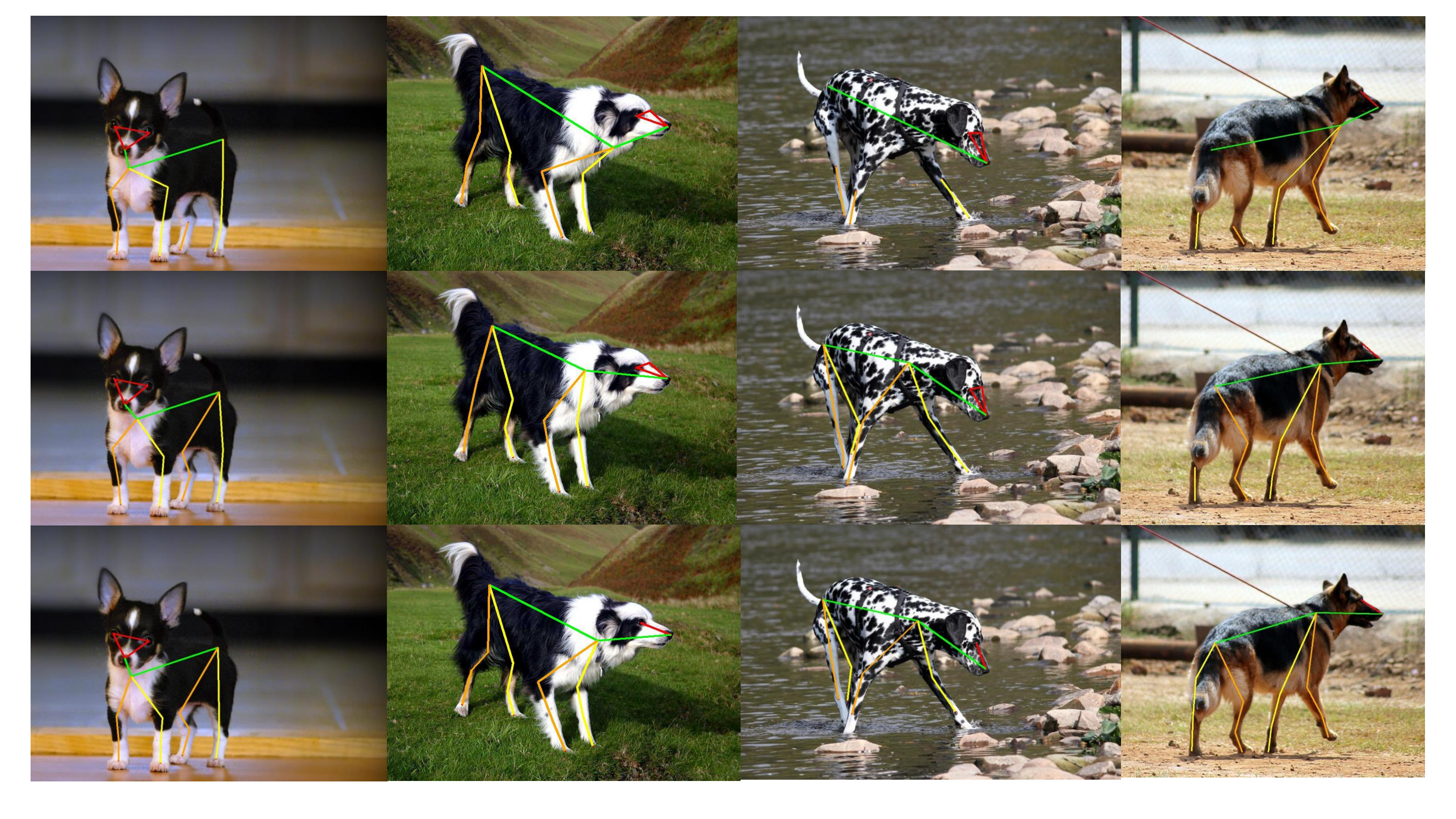}
     \caption{More qualitative results of HRNet-w32 trained on the Animal Pose dataset~\cite{Cao_2019_ICCV} (the first row) and our AP-10K dataet (the second row). The ground truth poses are shown in the last row. These animals are Chihuauha, Collie, Dalmatian, and German Shepherd, respectively.
    }
    \label{fig:appendix_pose_5}
\end{figure}

\begin{figure}
    \centering
    \includegraphics[width=1\linewidth]{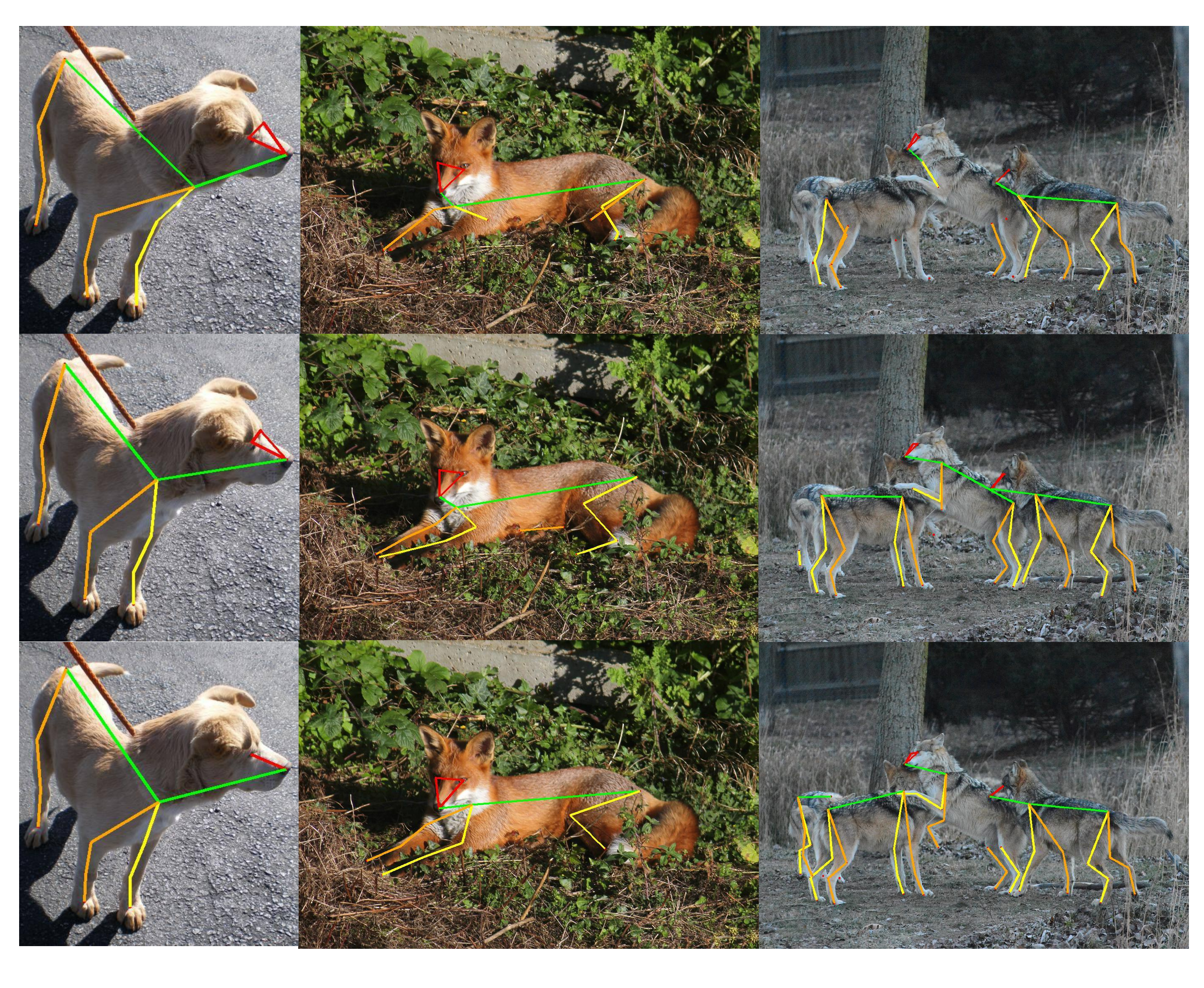}
    \caption{More qualitative results of HRNet-w32 trained on the Animal Pose dataset~\cite{Cao_2019_ICCV} (the first row) and our AP-10K dataet (the second row). The ground truth poses are shown in the last row. These animals are Dog, Fox, and Wolf, respectively.
    }
    \label{fig:appendix_pose_6}
\end{figure}

\begin{figure}
    \centering
    \includegraphics[width=1\linewidth]{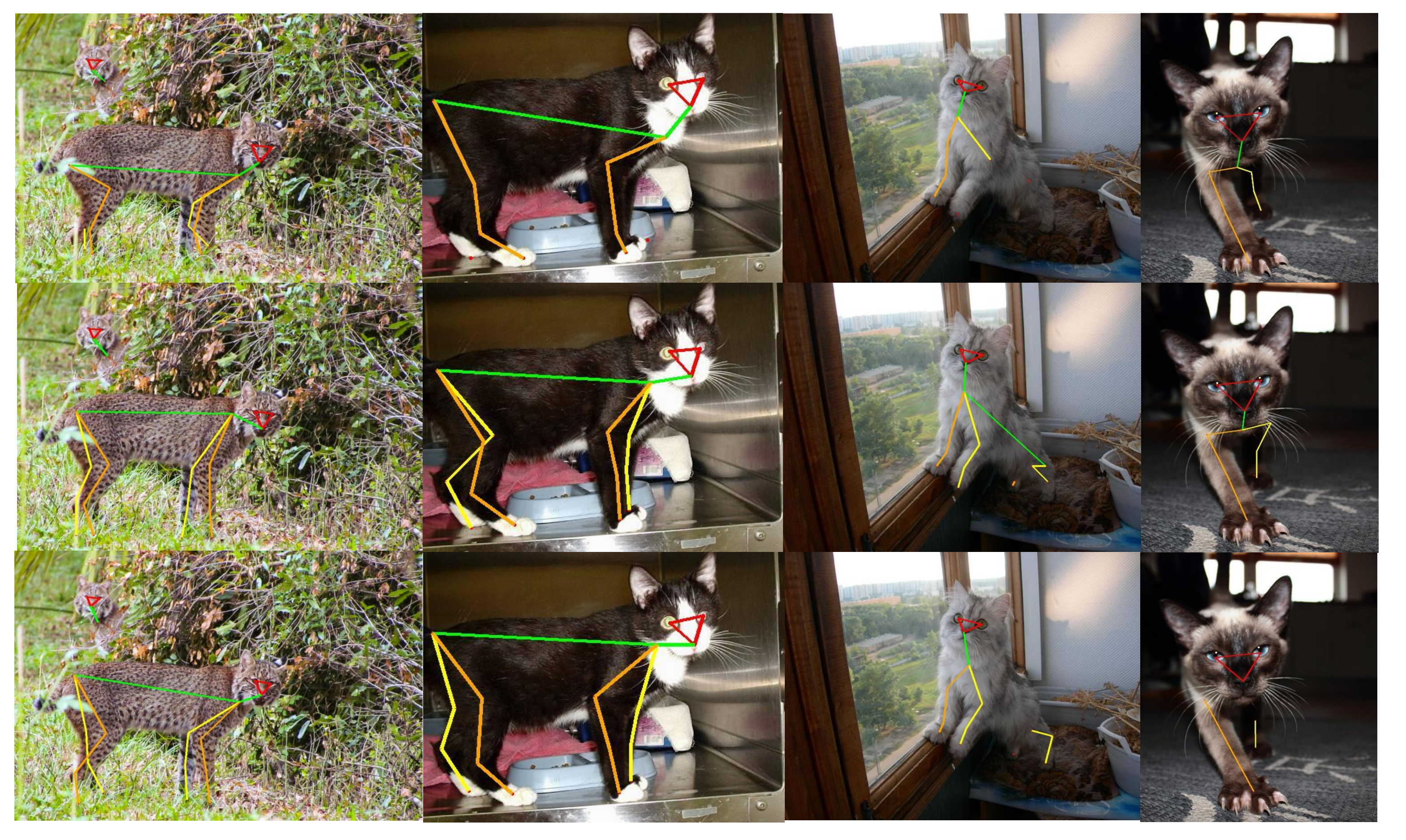}
    \caption{More qualitative results of HRNet-w32 trained on the Animal Pose dataset~\cite{Cao_2019_ICCV} (the first row) and our AP-10K dataet (the second row). The ground truth poses are shown in the last row. These animals are Bobcat, Cat, Persian Cat, and Siamese Cat, respectively.
    }
    \label{fig:appendix_pose_7}
\end{figure}

\begin{figure}
    \centering
    \includegraphics[width=1\linewidth]{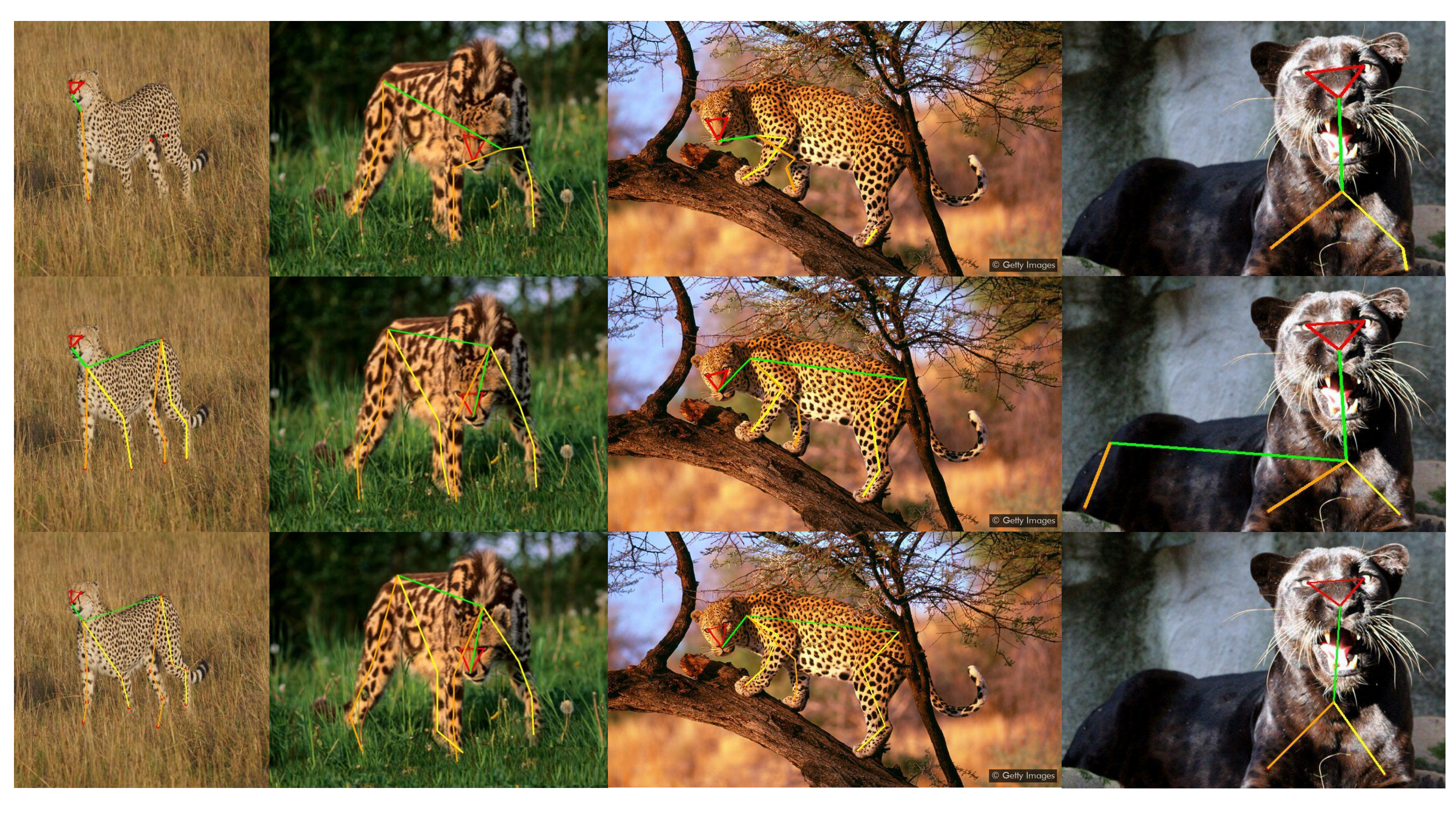}
     \caption{More qualitative results of HRNet-w32 trained on the Animal Pose dataset~\cite{Cao_2019_ICCV} (the first row) and our AP-10K dataet (the second row). The ground truth poses are shown in the last row. These animals are Cheetah, King Cheetah, Leopard, and Panther, respectively.
    }
    \label{fig:appendix_pose_8}
\end{figure}

\begin{figure}
    \centering
    \includegraphics[width=1\linewidth]{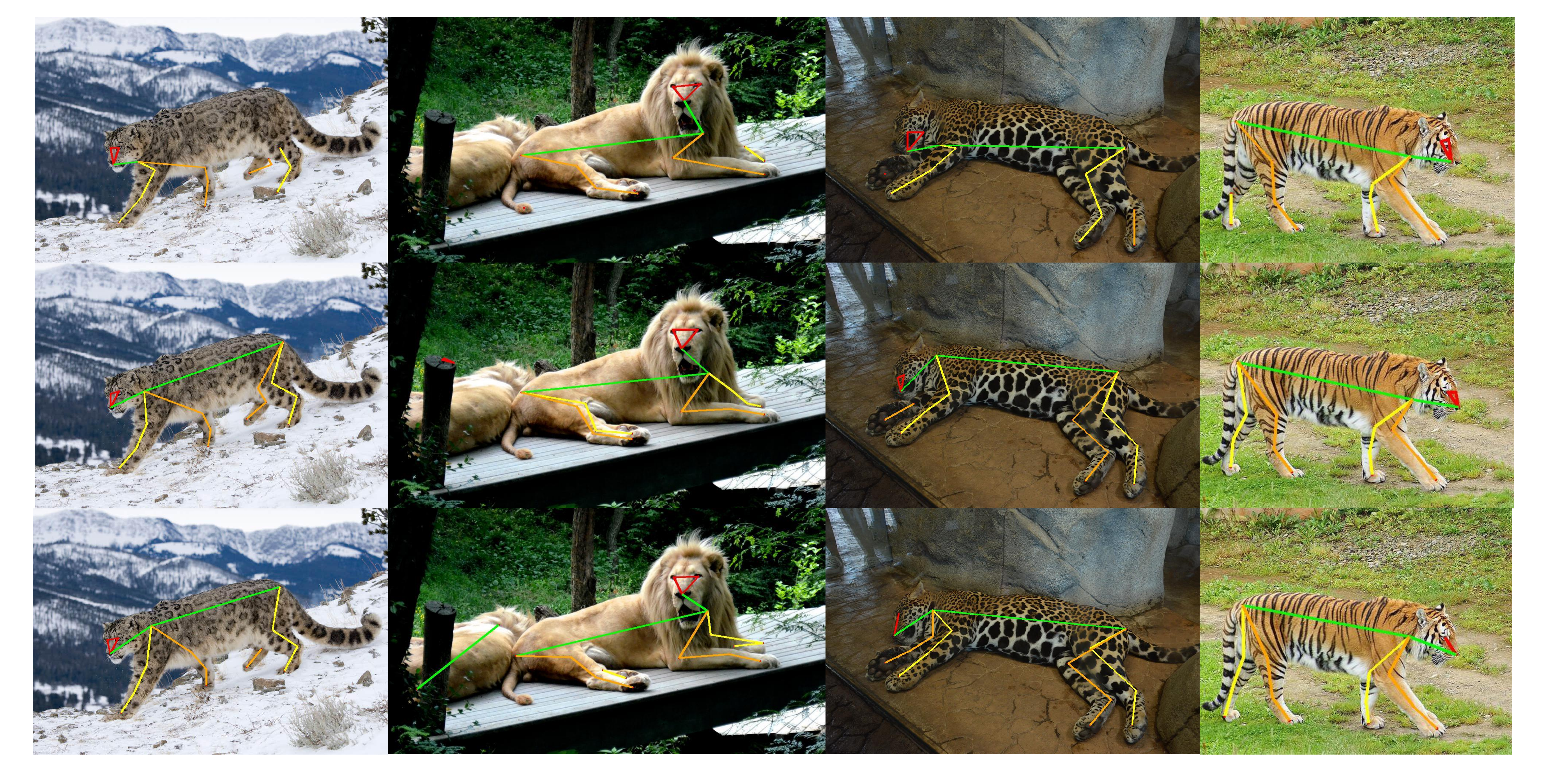}
   \caption{More qualitative results of HRNet-w32 trained on the Animal Pose dataset~\cite{Cao_2019_ICCV} (the first row) and our AP-10K dataet (the second row). The ground truth poses are shown in the last row. These animals are Snow Leopard, Lion, Jaguar, and Tiger, respectively.
    }
    \label{fig:appendix_pose_9}
\end{figure}

\begin{figure}
    \centering
    \includegraphics[width=1\linewidth]{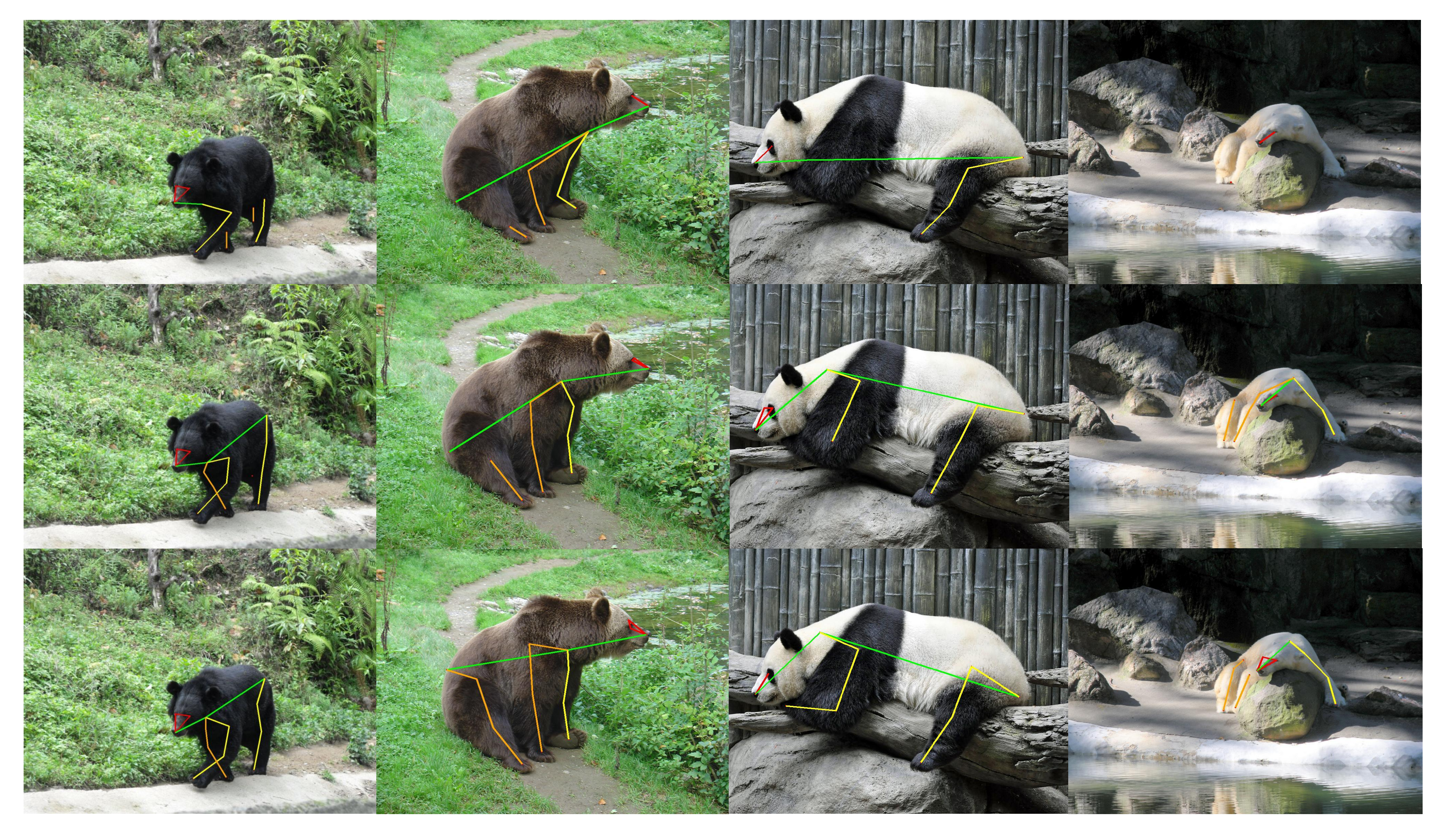}
    \caption{More qualitative results of HRNet-w32 trained on the Animal Pose dataset~\cite{Cao_2019_ICCV} (the first row) and our AP-10K dataet (the second row). The ground truth poses are shown in the last row. These animals are Black Bear, Brown Bear, Panda, and Polar Bear, respectively.
    }
    \label{fig:appendix_pose_10}
\end{figure}

\begin{figure}
    \centering
    \includegraphics[width=1\linewidth]{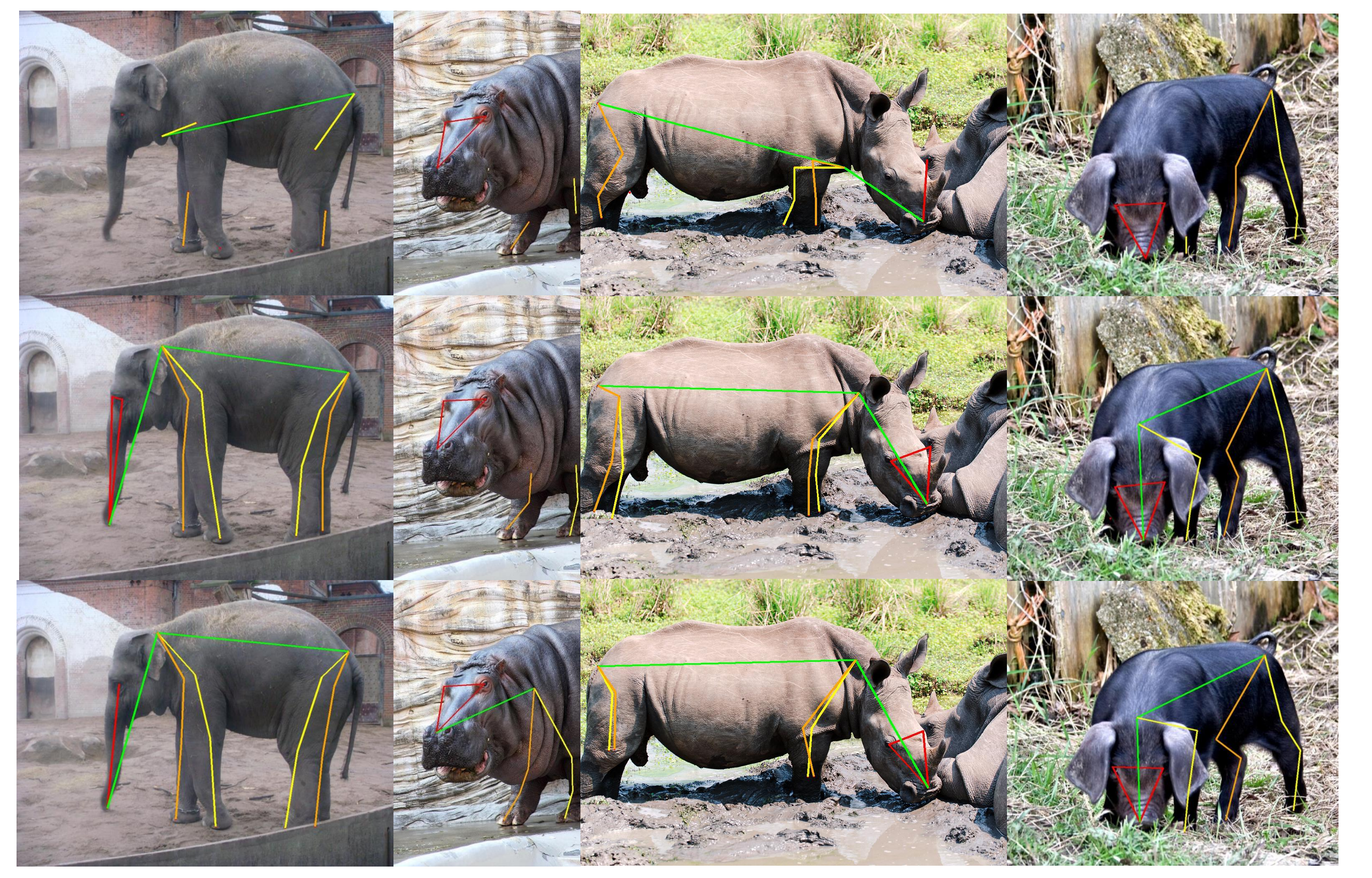}
    \caption{More qualitative results of HRNet-w32 trained on the Animal Pose dataset~\cite{Cao_2019_ICCV} (the first row) and our AP-10K dataet (the second row). The ground truth poses are shown in the last row. Elephant, Hippo, Rhino, and Pig, respectively.
    }
    \label{fig:appendix_pose_11}
\end{figure}

\begin{figure}
    \centering
    \includegraphics[width=1\linewidth]{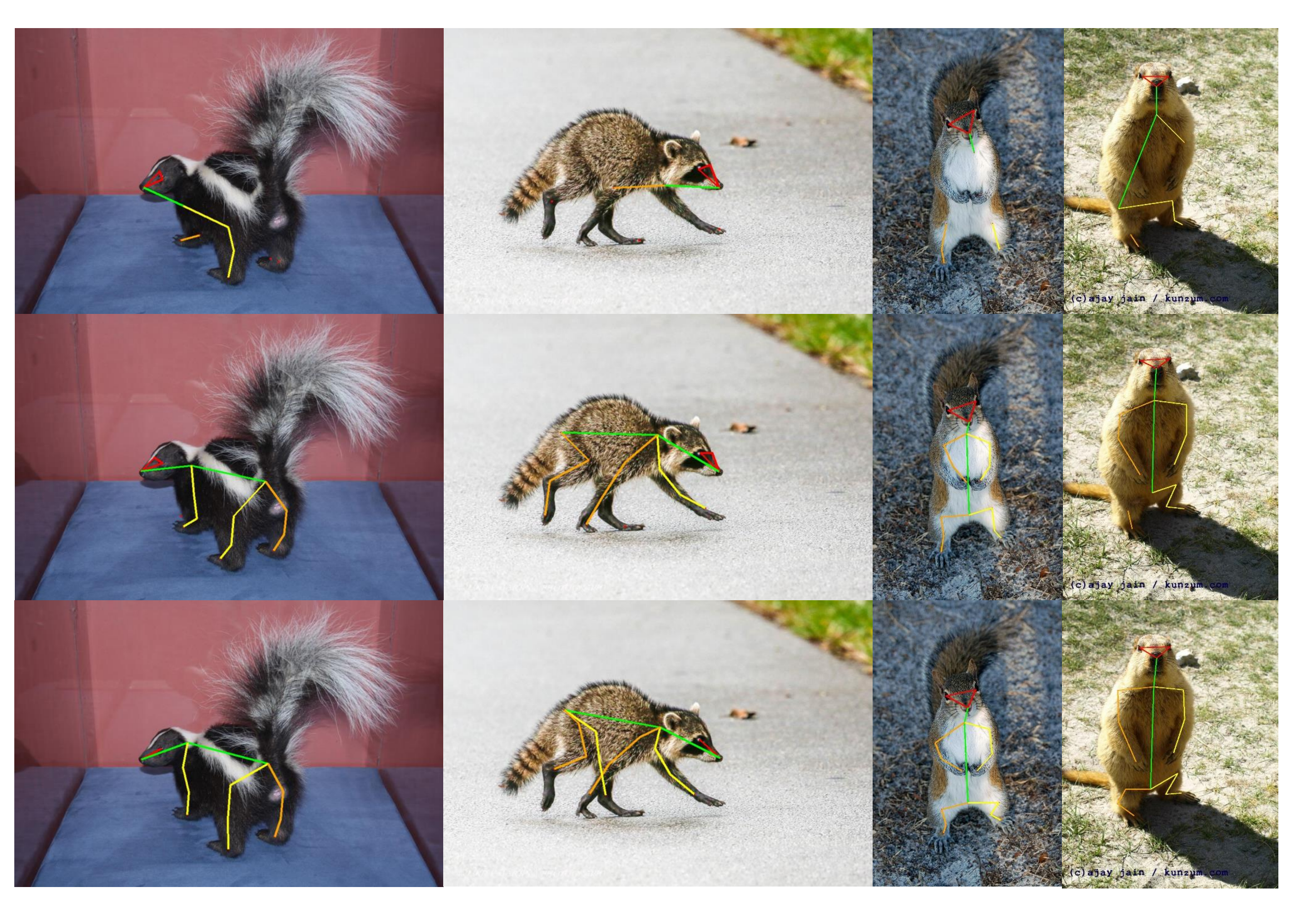}
    \caption{More qualitative results of HRNet-w32 trained on the Animal Pose dataset~\cite{Cao_2019_ICCV} (the first row) and our AP-10K dataet (the second row). The ground truth poses are shown in the last row. These animals are Skunk, Raccoon, Squirrel, and Marmot, respectively.
    }
    \label{fig:appendix_pose_12}
\end{figure}

\begin{figure}
    \centering
    \includegraphics[width=1\linewidth]{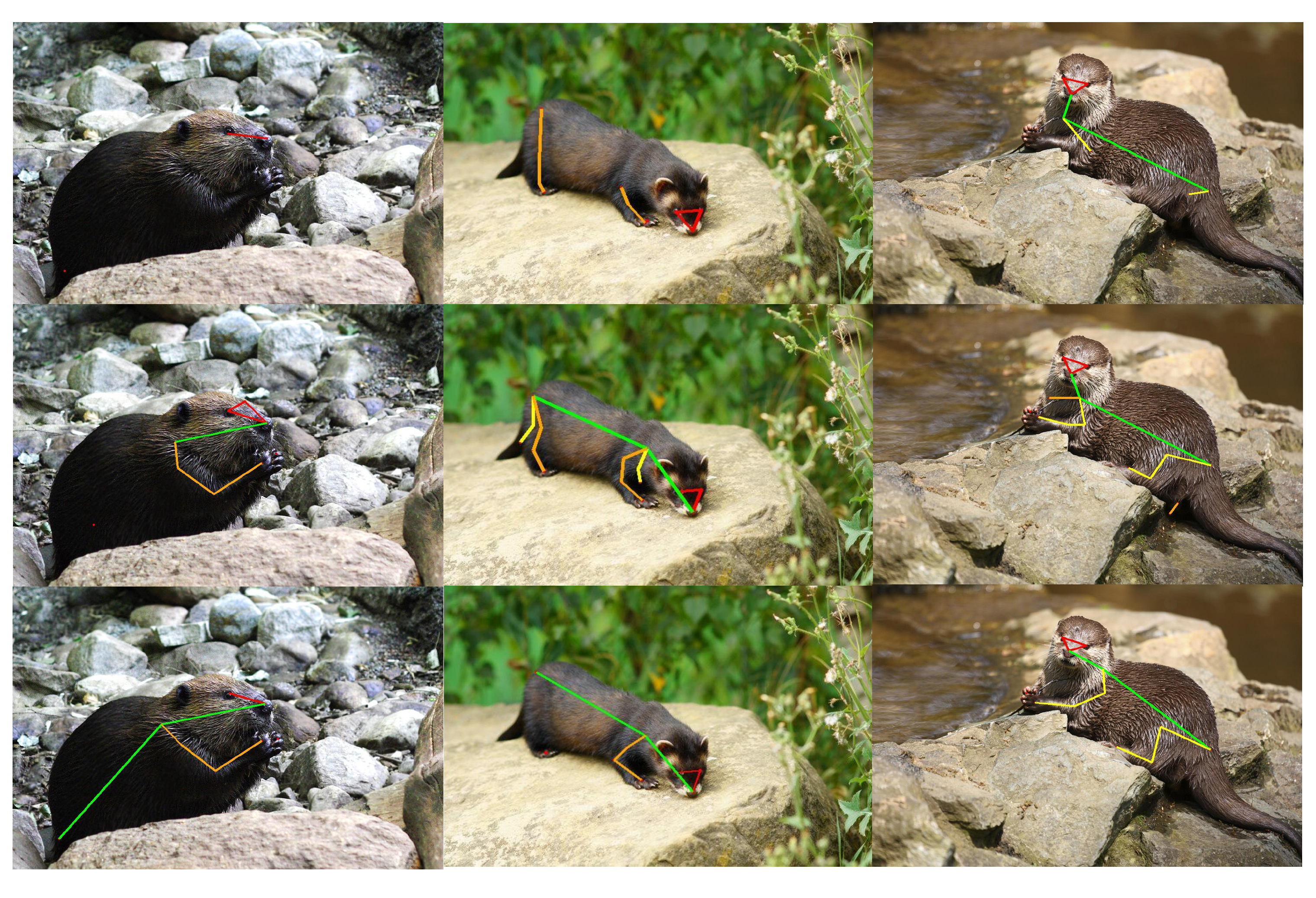}
     \caption{More qualitative results of HRNet-w32 trained on the Animal Pose dataset~\cite{Cao_2019_ICCV} (the first row) and our AP-10K dataet (the second row). The ground truth poses are shown in the last row. These animals are Beaver, Otter, and Weasel, respectively.
    }
    \label{fig:appendix_pose_13}
\end{figure}
\begin{figure}
    \centering
    \includegraphics[width=1\linewidth]{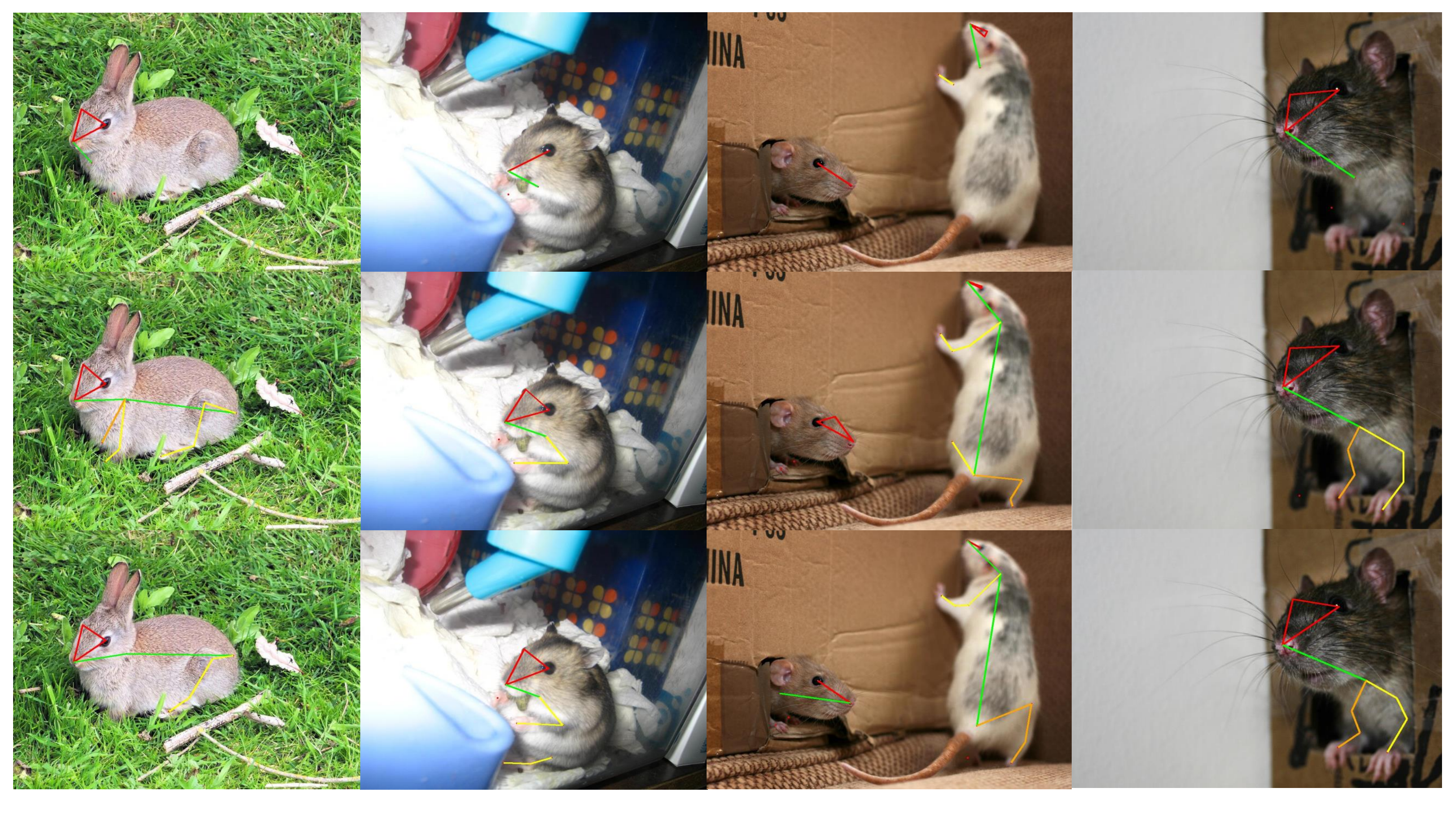}
    \caption{More qualitative results of HRNet-w32 trained on the Animal Pose dataset~\cite{Cao_2019_ICCV} (the first row) and our AP-10K dataet (the second row). The ground truth poses are shown in the last row.These animals are Rabbit, Hamster, Mouse, and Rat, respectively.
    }
    \label{fig:appendix_pose_14}
\end{figure}

\end{document}